\documentclass[runningheads,a4paper]{llncs}

\usepackage{amssymb}
\usepackage{graphicx}

\usepackage{booktabs}
\usepackage{multirow}
\usepackage{subfig}
\usepackage{caption}

\usepackage{rotating}

\usepackage{url}

\urldef{\mailsa}\path|joseph.antony@insight-centre.org|
\newcommand{\keywords}[1]{\par\addvspace\baselineskip
\noindent\keywordname\enspace\ignorespaces#1}

\emergencystretch=\maxdimen
\begin{document}

\mainmatter  % start of an individual contribution

\title{Feature Learning to Automatically Assess Radiographic Knee Osteoarthritis Severity}

\titlerunning{Feature Learning to Assess Knee OA Severity}

\author{Joseph Antony, Kevin McGuinness, Kieran Moran, Noel E O' Connor}
\authorrunning{Joseph Antony, Kevin McGuinness, Kieran Moran, Noel E O'Connor}

\institute{Insight Centre for Data Analytics, Dublin City University, Ireland\\
\mailsa}

\toctitle{Abstract}
\tocauthor{ }
\maketitle

%--------------------------------------------------------------------------%

\begin{abstract}

Feature learning refers to techniques that learn to transform raw data input into an effective representation for further higher-level processing in many computer vision tasks. This chapter presents the investigations and the results of feature learning using convolutional neural networks to automatically assess knee osteoarthritis (OA) severity and the associated clinical and diagnostic features of knee OA from radiographs (X-ray images). Also, this chapter demonstrates that feature learning in a supervised manner is more effective than using conventional handcrafted features for automatic detection of knee joints and fine-grained knee OA image classification. In the general machine learning approach to automatically assess knee OA severity, the first step is to localize the region of interest that is to detect and extract the knee joint regions from the radiographs, and the next step is to classify the localized knee joints based on a radiographic classification scheme such as Kellgren and Lawrence grades. First, the existing approaches for detecting (or localizing) the knee joint regions based on handcrafted features are reviewed and outlined in this chapter. Next, three new approaches are introduced: 1) to automatically detect the knee joint region using a fully convolutional network, 2) to automatically assess the radiographic knee OA using CNNs trained from scratch for classification and regression of knee joint images to predict KL grades in ordinal and continuous scales, and 3) to quantify the knee OA severity optimizing a weighted ratio of two loss functions: categorical cross entropy and mean-squared error %and CNNs are jointly trained to quantify the clinical features of the knee OA: joint space narrowing and osteophytes along with the KL grades 
using multi-objective convolutional learning. The results from these methods show progressive improvement in the overall quantification of the knee OA severity. Two public datasets: the OAI and the MOST are used to evaluate the approaches with promising results that outperform existing approaches. In summary, this work primarily contributes to the field of automated methods for localization (automatic detection) and quantification (image classification) of radiographic knee OA.


\end{abstract}
%--------------------------------------------------------------------------%

\tableofcontents

%--------------------------------------------------------------------------%

\section{Introduction}
Traditionally, many handcrafted features have been successfully used in computer vision tasks, and they often simplify machine learning tasks. Nevertheless, they have a few limitations. These features are often low-level as prior knowledge is hand-encoded and features in one domain do not always generalize to other domains.  In recent years, learning feature representations in a supervised manner also known as supervised feature learning is preferred over handcrafted features as they have outperformed the state-of-the-art in many computer vision tasks and have been highly successful. This chapter focuses on feature learning to automatically assess radiographic knee OA severity using convolutional neural networks (CNNs). 

Clinically to assess knee OA severity, highly experienced clinicians or radiologists assess the knee joints in X-ray images \cite{shamir2009knee,thomson2015automated} and assign an ordinal grade based on a radiographic grading scheme. The most commonly used gradings, like the Kellgren and Lawrence (KL) grading scheme and Ahlback system, use distinctive grades (0 to 4). However, clinical features of knee OA are continuous in nature, and attributing distinctive grades is the subjective opinion of the graders. There are also uncertainties and variations in the subjective gradings. There is a need for automated methods to overcome the limitations arising from this subjectivity, and to improve the reliability in the measurements and classifications \cite{thomson2015automated}.

 The automatic assessment of knee OA severity has been previously approached in the literature as an image classification problem \cite{shamir2009knee,shamir2009early,shamir2008wndchrm}, with the KL grading scale as the ground truth. WNDCHARM\footnote{Weighted Neighbor Distance using Compound Hierarchy of Algorithms Representing Morphology}, a multi-purpose biomedical image classifier was used to classify knee OA images \cite{shamir2008wndchrm,orlov2008wnd}. High binary classification accuracies (80\% to 91\%) have been reported using the WNDCHARM classifier for classifying the extreme stages: grade 0 (normal) vs grade 4 (severe), grade 0 vs grade 3 (moderate). However, the classification accuracies of the images belonging to successive grades are low (55\% to 65\%) and the multi-class classification accuracy is low (35\%). The overall classification accuracies of knee OA needs improvement for real-world computer aided diagnosis \cite{shamir2009knee,shamir2009early,oka2008fully}. 
 
Radiographic features detected and learned through a computer-aided analysis can be useful to quantify knee OA severity and to predict the future development of knee OA \cite{shamir2009early}. Instead of manually designing features, the author proposes that learning feature representations using deep learning architectures can be a more effective approach for the classification of knee OA images. Traditionally, hand-crafted features based on pixel statistics, object and edge statistics, texture, histograms, and transforms, are typically used for multi-purpose medical image classification \cite{shamir2008wndchrm,orlov2008wnd,park2013practical}. However, these features are not efficient for fine-grained classification such as classifying successive grades of knee OA images. Manually designed or hand-engineered features often simplify machine learning tasks. Nevertheless, they have a few disadvantages. The process of engineering features requires domain related expert knowledge and is often very time consuming \cite{lee2010}. These features are often low-level as prior knowledge is hand-encoded and features in one domain do not always generalize to other domains \cite{le2013}. The next logical step is to automatically learn effective features for the desired task.

Over the last decade, learning feature representations or feature learning has been preferred to hand-crafted features in many computer vision tasks, particularly for fine-grained classification, because rich appearance and shape features are essential for describing subtle differences between categories \cite{yang2013feature}. Feature learning refers to techniques that learn to transform raw data input or pixels of an image to an effective representation for further higher-level processing such as object detection, automatic detection, segmentation, and classification. Feature learning approaches provide a natural way to capture cues by using a large number of code words (sparse coding) or neurons (deep networks), while traditional computer vision features, designed for basic-level category recognition, may eliminate many useful cues during feature extraction \cite{yang2013feature}. Deep learning architectures are multi-layered and they are used to learn feature representations in the hidden layer(s). These representations are subsequently used for classification or regression at the output layer. Feature learning is an integral part of deep learning \cite{lee2010,donoghe2013}.

Even though many deep learning architectures have been proposed and have existed for decades, in recent times CNNs have become highly successful in the field of computer vision \cite{litjens2017survey,shen2017deep}. AlexNet\cite{krizhevsky2012imagenet} won the ILSVRC\footnote {ImageNet Large Scale Visual Recognition Challenge} in 2012 by a large margin. CNNs have since then become more popular, widely-used and highly-successful in computer vision tasks such as object detection, image recognition, automatic detection and segmentation, content based image retrieval, and video classification \cite{litjens2017survey}. Apart from computer vision tasks, CNNs are finding applications in natural language processing, hyper-spectral image processing, and medical image analysis \cite{litjens2017survey,tajbakhsh2016conv}.

CNNs have also recently become successful in many medical applications such as knee cartilage segmentation \cite{prasoon2013deep} and brain tumour segmentation \cite{havaei2017brain} in MRI scans, multi-modality iso-intense infant brain image segmentation \cite{zhang2015deep}, pancreas segmentation in CT images \cite{roth2015deep}, and neuronal membrane segmentation in electron microscopy images \cite{ciresan2012deep}. Inspired by these success stories, the author proposes CNNs for classification of knee OA images and to improve the quantification of knee OA severity and knee OA diagnostic features. The author believes that this can lead to building a real-world knee OA diagnostic system that outperforms the existing approaches.

The remainder of this chapter is organized as follows. Section 1.1 introduces knee osteoarthritis (OA), the diagnostic features and the clinical evaluation of knee OA. Section 1.2 lists the contributions of this research. Section 2 provides an overview of the background, a comprehensive summary of the related work, and a critical analysis of the state-of-the-art in computer aided diagnosis of knee OA. Section 3 introduces the public datasets used in this study. Section 4 presents the baseline methods using hand-crafted features and the proposed approaches in this chapter for automatic detection of knee joints in the radiographs. Section 5 presents the baseline methods and the proposed approaches in this chapter to quantify knee OA severity using CNNs. %Section 6 presents the automatic quantification of knee OA diagnostic features: joint space narrowing and osteophytes using multi-objective convolutional learning. 
Section 6 concludes this chapter by analyzing the current work and summarizing the research methodology, and providing future directions of research based on the proposed methods.

\subsection{Knee Osteoarthritis}
Knee Osteoarthritis (OA) is a debilitating joint disorder that mainly degrades the knee articular cartilage and in its severe stages it causes excruciating pain and often leads to total joint arthroplasty. In general, knee OA is characterized by joint pain, cartilage wear, and bony growths.  Knee OA has a high-incidence among the elderly, obese, and those with a sedentary lifestyle. Early diagnosis is crucial for clinical treatments and pathology \cite{shamir2009early,oka2008fully}. 

\subsubsection{Diagnostic Features of Knee OA}
\begin{figure}[t]
\centering
\includegraphics[width=0.75\textwidth]{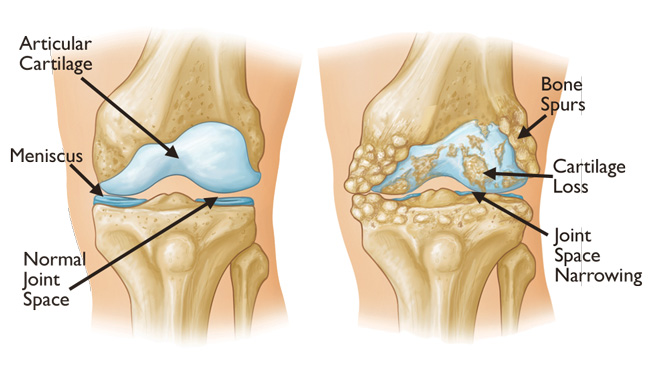}
\caption{A healthy knee and a knee joint affected with OA.} 
\label{fig:KneeOA}
%\small{Source: {\url{https://orthoinfo.aaos.org/en/diseases--conditions/}}}
\end{figure}

~\\
Clinically, the major pathological features for knee OA include joint space narrowing, osteophytes formation, and sclerosis \cite{oka2008fully,marijnissen2008}. Figure \ref{fig:KneeOA} shows the anatomy of a healthy knee and a knee affected with osteoarthritis, and the characteristic features of knee OA. The causes for knee OA include mechanical abnormalities such as degradation of articular cartilage, menisci, ligaments, synovial tissue, and sub-chondral bone.

The major clinical features, joint space narrowing and osteophyte formation, are easily visualized using radiographs \cite{oka2008fully,park2013practical,felson1997defining}. Despite the introduction of several imaging methods such as magnetic resonance imaging (MRI), computed tomography (CT), and ultrasound for augmented OA diagnosis, radiographs have traditionally been preferred \cite{braun2012,felson1997defining}, and remain as the main accessible tool and ``gold standard" for preliminary knee OA diagnosis \cite{shamir2009knee,shamir2009early}. Inspired by the previous successful approaches in the literature for early identification \cite{shamir2009early} and automatic assessment of knee OA severity  \cite{shamir2008wndchrm,oka2008fully,braun2012}, the focus is on radiographs in this work. More importantly, there are public datasets available that contain radiographs with associated ground truth. Public datasets for knee OA study, such as the OAI\footnote{Osteoarthritis Initiative} and the MOST\footnote{Multicenter Osteoarthritis Study} datasets, provide radiographs with KL scores, and the OARSI\footnote{Osteoarthritis Research Society International} readings for distinct knee OA features such as JSN, osteophytes, and sclerosis. 

% \subsubsection{Clinical Significance of Knee OA Studies}~\\
% Due to the increasing prevalence of knee OA, diminishing health-related quality of life, and total joint arthroplasty as a serious consequence, there is a growing need for effective clinical and scientific tools for early detection of knee OA reliably \cite{shamir2009knee,shamir2009early,oka2008fully}. Early identification of knee OA and assessment of the severity are crucial for pathology, clinical decision making, and to study disease progression \cite{braun2012}. 
% As per a recent study \cite{gelber2015}, more than 250 million people across the globe are affected by knee OA alone. A study \cite{culliford2015future} on future projections of total hip and knee arthroplasty in the UK estimates the total primary hip and knee replacement counts in 2035 at 439,097 and 1,219,362 respectively. The National Institutes of Health (NIH) has sponsored a research project called the Osteoarthritis Initiative (OAI) to develop a public domain resource to facilitate knee OA research, to identify and validate knee OA biomarkers that will help to better understand how to prevent and treat knee OA.

\subsubsection{Radiographic Classification of Knee OA}~\\
Knee OA develops gradually over years and progresses in stages. In general, the severity of knee OA is divided into five stages. The first stage (stage 0) corresponds to normal healthy knee and the final stage (stage 4) corresponds to the most severe condition (see Figure 2). The most commonly used systems for grading knee OA are the International Knee Documentation Committee (IKDC) system, the Ahlback system, and the Kellgren \& Lawrence (KL) grading system. The other widely used non-radiographic knee OA assessment system is WOMAC\footnote{Western Ontario and McMaster Universities Osteoarthritis Index}, which measures pain, stiffness, and functional limitation. The public datasets, the OAI and the MOST used in this work, are provided with the KL grades and they are used as the ground truth to classify the knee OA X-ray images.

\textbf{Kellgren and Lawrence Scores.} 

\begin{figure}[t]
  \centering
  \includegraphics[width=0.8\textwidth]{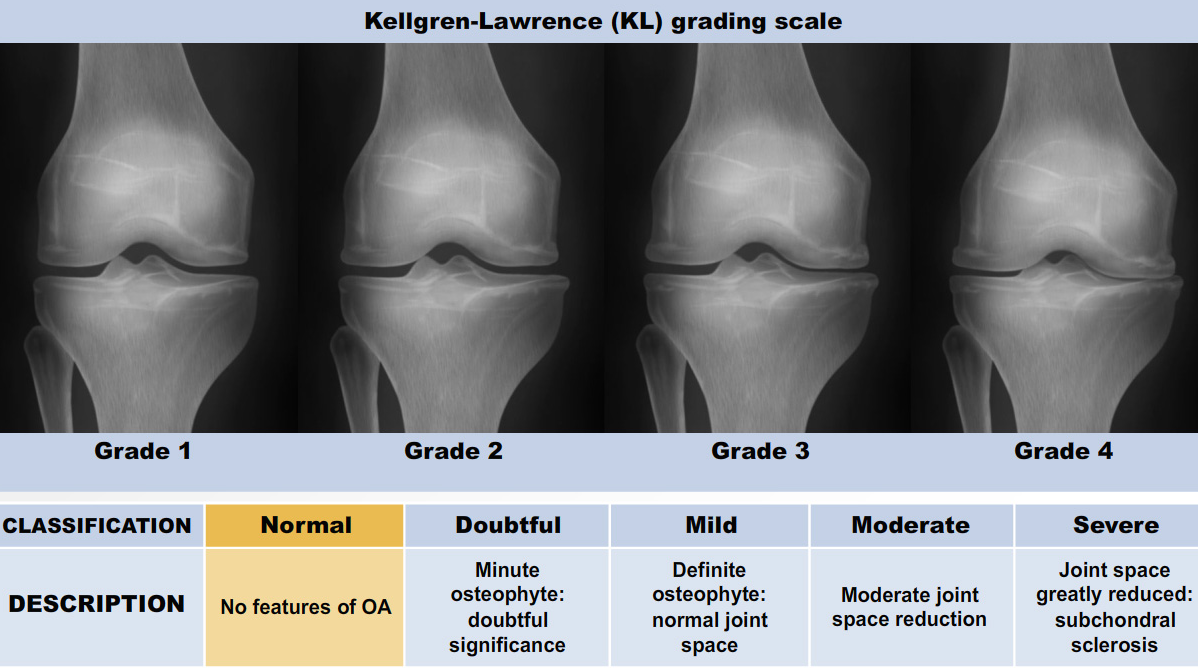}
  \caption{The Kellgren and Lawrence grading system to assess the severity of knee OA.}
  \label{fig:KL}
  %\small{Source: \url{http://www.adamondemand.com/clinical-management-of-osteoarthritis/}}
\end{figure}

The KL grading scale was approved by the World Health Organisation as the reference standard for cross-sectional and longitudinal epidemiologic studies \cite{park2013practical,felson1997defining,emrani2008joint,hart2003kellgren}. The KL grading system is still considered the gold standard for initial assessment of knee osteoarthritis severity in radiographs \cite{shamir2009knee,orlov2008wnd,oka2008fully,park2013practical}. Figure \ref{fig:KL} shows the KL grading system. The KL grading system categorizes knee OA  severity into five grades (grade 0 to 4). The KL grading scheme for quantifying knee OA severity from X-ray images is defined as follows \cite{shamir2009knee,orlov2008wnd}:

\begin{itemize}
  \item[]Grade 0 : absence of radiographic features (cartilage loss or osteophytes) of OA.
  \item[]Grade 1 : doubtful joint space narrowing (JSN), osteophytes sprouting, bone marrow oedema (BME), and sub-chondral cyst.
  \item[]Grade 2 : visible osteophytes formation and reduction in joint space width on the antero-posterior weight-bearing radiograph with BME and sub-chondral cyst.
  \item[]Grade 3 : multiple osteophytes, definite JSN, sclerosis, possible bone deformity.
  \item[]Grade 4 : large osteophytes, marked JSN, severe sclerosis, and definite bone deformity.
\end{itemize}

\subsection{Contributions}
The research contributions of this work are as follows.
\begin{itemize}
\item Proposing a novel and highly accurate technique to automatically detect and localise the knee joints from the X-ray images using a fully convolutional network (FCN).
\item Developing a classifier based on a CNN to assess knee OA severity that is highly accurate in comparison to previous methods. 
\item Proposing a novel approach to train a CNN with a weighted ratio of two loss functions: categorical cross entropy and mean squared error with the natural benefit of predicting knee OA severity in ordinal (0, 1 to 4) and continuous (0--4) scales.
\item Developing an ordinal regression approach using CNNs to automatically quantify knee OA severity in a continuous scale. 
%\item Developing CNN classifiers to assess the distinct knee OA diagnostic features: lateral and medial JSN, femoral and tibial osteophytes in lateral and medial compartments.
%\item Further improving the quantification of knee OA severity by jointly training CNN to predict the knee OA clinical features along with the KL grades.
\item Developing an automatic knee OA diagnostic system i.e. an end-to-end pipeline incorporating the FCN for automatically localising the knee joints and the CNN for automatically quantifying the localised knee joints. %The overall assessment of knee OA by this system agrees with the gold standard. 
\end{itemize}

%--------------------------------------------------------------------------%

\section{Related Work and Background}
The automatic assessment of knee OA severity from radiographs has been approached as an image classification problem \cite{shamir2009knee,shamir2009early,shamir2008wndchrm}. According to the literature and in the machine learning approach to automatically assess knee OA severity, the first step is to localise the region of interest (ROI) that is to detect and extract the knee joint regions from the radiographs, and the next step is to classify the localised knee joints. First, the different approaches for detecting (or localising) the knee joint regions in the radiographs are outlined. Next, the approaches in the literature to assess knee OA severity are investigated and the focus is on the automated methods. %Following this, the key pathological features of knee OA are introduced and the state-of-the-art methods for quantifying radiographic knee OA clinical features are reviewed. 
This section concludes with a discussion outlining the limitations in the state-of-the-art methods on automatic detection of knee joints and automatic assessment of knee OA severity, and how these limitations can be addressed. 

\subsection{Detecting Knee Joints in Radiographs}
%Despite the introduction of several imaging modalities such as MRI, CT, and ultrasound for augmented OA diagnosis, radiography (X-ray) has been traditionally preferred, and remains the main accessible tool and ``gold standard" for preliminary knee OA diagnosis \cite{shamir2009knee,shamir2008wndchrm,oka2008fully}. The main pathological features of knee OA such as loss of joint cartilage, reduction in joint space width, and osteophytes (bone spurs), can be easily visualized and examined in plain radiographs \cite{felson1997defining,braun2012}. Advanced imaging modalities such as MRI and CT may be required when the clinical investigations from radiographs are inconclusive and do not give clear reasons for joint pain \cite{braun2012}. Expert radiologists specifically examine the knee joint regions in radiographs for joint space narrowing and osteophytes, for knee OA diagnosis \cite{shamir2009early}. 

There are several approaches in the literature for detecting and segmenting knee joints and specific parts of the knee such as cartilage, menisci, and bones structures from 3D MRI and CT scan images \cite{shaikh2014image,sun2006discussions}. Nevertheless, the existing approaches are less accurate for automatically detecting the knee joints in radiographs \cite{sun2006discussions,gornale2016detection}. According to the literature, detecting knee joints remains a challenging task \cite{sun2006discussions,tiulpin2017novel}. In this chapter, automated methods for detecting knee joints in radiographs are investigated. %For this reason, the author limits the scope of review to the approaches based on radiographs. 
The advantages of automatic methods are discussed and the need to investigate such methods are emphasized. Previous approaches in the literature that investigate the knee joints in radiographs can be categorized into manual, semi-automatic, and fully automatic, based on the level of manual intervention required \cite{shaikh2014image,sun2006discussions}. %The following sections review each of these approaches in detail.

\subsubsection{Manual Methods}~\\
Expert radiologists or trained physicians visually examine the knee joint regions and trace the structures using simple image processing and computer vision-based tools in radiographs, and may even use CAD-based measurements for assessing knee OA severity \cite{sun2006discussions}. The expert knowledge-based manual segmentations are useful to build an atlas or template of anatomical structures, which are used to develop advanced interactive and automatic segmentation methods \cite{sun2006discussions}. The knee joints labelled manually are reliable and are often used as ground truth for evaluating automatic methods \cite{stammberger1999interobserver,cohen1999knee}. Nevertheless, such manual methods are subjective, highly experience-based, and they are laborious and time-consuming when a large number of subjects are to be examined. 

There are previous studies in the literature that use manually-defined ROIs (knee joints) in radiographs for assessing knee OA severity. Hirvasniemi et al.\cite{hirvasniemi2014quantification} quantified the differences in bone density using texture analysis and local binary patterns (LBP) in plain radiographs to assess knee osteoarthritis. Woloszynski et al.\cite{woloszynski2010signature} developed a signature dissimilarity measure for the classification of trabecular bone texture in knee radiographs. In both these methods, the ROIs are manually marked and extracted for texture analysis. 

\subsubsection{Semi-automatic Methods}~\\
Semi-automatic or interactive methods are developed to minimize manual interventions by automating essential steps in the detection and segmentation process \cite{shaikh2014image,zhao2013overview}. These methods often include manual initialization with low-level image processing, followed by manual evaluations and corrections of the results \cite{pirnog2005articular}. The main advantage of the semi-automatic methods are flexibility in manual intervention that allow incorporating expert knowledge, plus the use of advanced computer vision-based tools to automate the essential steps. An expert may improve the detection and segmentation performance through tuning the essential parameters for instance seed region and threshold values in region growing, initial shape of active models, and delineating the required contour \cite{sun2006discussions} to define the region of interest. However, these methods may not be reproducible due to inter-observer or inter-user variations and there is a possibility of oversight or human error in the manual evaluations. 

There are some knee OA studies in the literature which use semi-automatic methods to detect the knee joints in radiographs. Knee OA computer aided diagnosis (KOACAD)\cite{oka2008fully} is an interactive method to measure the joint space narrowing, osteophytes formation and joint angulation in radiographs. In KOACAD, a Roberts filter is used to obtain the rough contour of tibia and femur bone structures and a vertical neighbourhood difference filter is used to identify points with high absolute values of difference of scales. The centre of all the points is calculated and a rectangular region around the centre, of size $480\times200$ pixels, is selected as the knee joint region. This system has purportedly provided accurate assessment of structural severity of knee OA after detecting the knee joint regions. However, human intervention is required for plotting various lines for the measurement, and automatic detection is not feasible with this system. 

Knee images digital analysis (KIDA) is a tool to analyse knee radiographs interactively, proposed by Marijnissen et al. \cite{marijnissen2008}. KIDA quantifies the individual radiographic features of knee OA like medial and lateral joint space width (JSW) measurements, subchondral bone densities and osteophytes. However, this interactive tool can only be used by experts for quantitative measurements and requires expert intervention for objective quantitative evaluation.  

Duryea et al. \cite{duryea2000trainable} proposed a trainable rule-based algorithm (software) to measure the joint space width between the edges of the femoral condyle and the tibial plateau on knee radiographs. Contours marking the edges of the femur and tibia are automatically generated. This interactive method can be used to monitor joint space narrowing and the progression of knee osteoarthritis.

\subsubsection{Automatic Methods}~\\
Automatic segmentation methods have become an essential part of computer aided diagnosis and clinical decision support systems \cite{tiulpin2017novel}. These methods are fast and accurate, and they are highly beneficial in clinical trials and pathology \cite{sun2006discussions}. According to the literature, there have been multiple attempts to automatically localise knee joints in radiographs. Nevertheless, this task still remains a challenge. 

Podsiadlo et al. \cite{podsiadlo2008automated} proposed an automated system for the prediction and early diagnosis of knee OA. In this approach, active shape models and morphological operations are used to delineate the cortical bone plates and locate the ROIs in radiographs. This approach is developed for selection of tibial trabecular bone regions in the knee joints as ROIs. Nevertheless, this approach can be extended to localise the entire knee joint. A set of 40 X-ray images are used for training and 132 X-ray images are used for testing in this method. The automatic detections from this method are compared to the gold standard, which contains manually annotated ROIs from the expert radiologists and the similarity indices (SI) are calculated. This method achieved SI of 0.83 for the medial and 0.81 for the lateral regions of the knee joints.

Shamir et al. \cite{shamir2009knee} proposed template matching for automatic knee joint detection in radiographs. Template matching uses predefined joint centre images as templates and calculates Euclidean distances over every patch in an X-ray image using a sliding window. The image patch with the shortest distance is recorded as the detected knee joint centre. After detecting the centre, an image segment of 700$\times$500 pixels around the centre is extracted as the knee joint region. The X-ray images from the BLSA dataset are used in this method. In total 55 X-ray images from each grade are used for the experiments, such that 20 images from each grade for training and 35 images from each grade for testing. Shamir et al. reported that template matching was successful in finding the knee joint centres in all the X-ray images in their dataset.

Anifah et al. \cite{anifah2011automatic} investigated template matching and contrast-limited adaptive histogram equalisation for detecting knee joints and quantifying joint space area. In total 98 X-ray images are used in this method. The detection accuracy achieved by this method varies from 83.3\% to 100\% for the left knees and 60.4\% to 100\% for the right knees. Template matching is a simple and relatively fast method. However, this method is ad hoc, entirely based on the set of templates used and is unlikely to generalise well for larger datasets. 

Recently, Tuilpin et al. \cite{tiulpin2017novel} investigated a SVM-based method to automatically localise knee joints in plain radiographs. This method uses knee anatomy-based region proposals, and the best candidate region from the proposals are selected using histogram of oriented Gradients (HOG) as feature descriptors and a SVM. This method generalises well in comparison to the previous methods and shows reasonable improvement in automatic detections with mean intersection over union (IOU) of 0.84, 0.79 and 0.78 on the public datasets MOST, Jyvaskyla, and OKOA.

\subsection{Assessing Radiographic Knee OA Severity}
The key pathological features of knee OA include joint space narrowing, osteophytes (bone spurs) formation, and sclerosis (bone hardening) \cite{oka2008fully,marijnissen2008}. All these features are implicitly integrated in composite scoring systems, like Kellgren \& Lawrence (KL) grading system, to quantify knee OA severity \cite{oka2008fully,marijnissen2008}, and the OARSI readings provide the gradings of distinct knee OA features. There are two common approaches for assessing knee OA severity in plain radiographs: 1) quantifying the distinct pathological features of knee OA, and 2) automatic classification based on composite scoring systems such as KL grades. 

\subsubsection{Quantitative Analysis}~\\
The most conventional system to assess radiographic knee OA severity has been KL gradings \cite{shamir2009early,orlov2008wnd,oka2008fully}. Nevertheless, some researchers \cite{oka2008fully,marijnissen2008} argue that categorical systems like KL gradings are limited by incorrect assumptions that the progression of distinct OA features like JSN and osteophytes formation is linear and constant, their relationships are proportional, and such grading systems are less sensitive to small changes in distinct features. Therefore, quantification of individual features of knee OA is required to overcome the problems with KL gradings and to improve the overall radiographic assessment of knee OA \cite{oka2008fully,marijnissen2008}. The OARSI has published a radiographic atlas of individual features to assess and to quantitatively evaluate the knee OA features \cite{oka2008fully}. 

Interactive methods (KOACAD \cite{oka2008fully} and KIDA \cite{marijnissen2008}) measure individual knee OA radiographic features such as joint space width (JSW), osteophyte area, sub-chondral bone density, joint angle, and tibial eminence height as continuous variables. These measurements were compared to KL gradings and significant differences were found between healthy knees and knees with OA. In this context, a trainable rule-based algorithm has also been proposed \cite{duryea2000trainable} to measure the minimum joint space width (mJSW) between the edges of the femoral condyle and the tibial plateau, and thus to monitor the progression of knee OA. Podsiadlo et al. \cite{podsiadlo2008automated} have used a slightly different approach for quantitative knee OA analysis. In this method, the trabecular bone regions of the tibia are automatically located as the ROI after delineating the cortical bone plates using active shape models, followed by fractal analysis of bone textures for the diagnosis of knee OA. In a similar approach, Lee et al. \cite{lee2006automatic} use active shape models to detect the tibia and femur joint boundaries, and calculate anatomical geometric parameters to diagnose knee OA. 

Even though these methods are simple to implement, objective, and purpotedly accurate in evaluating radiographic knee OA, a great deal of manual intervention is required. Hence, these methods become very time-consuming and laborious when large numbers of subjects are to be investigated. Furthermore, the measurements from these methods are prone to inter- and intra-observer variability. 

\subsubsection{Automatic Classification}~\\
After the introduction of radiography-based semi quantitative scoring systems like KL gradings, the assessment of radiographic knee OA severity has been approached as an image classification problem \cite{thomson2015automated,shamir2009early,subramoniam2013local,subramoniam2015non,deokar2015effective}. According to the literature, the most common approach to classify knee OA images includes two steps: 1) extracting image features from the knee joints, and 2) applying a classification algorithm on the extracted features. A brief review of such approaches follows. 

Subramoniam et al. \cite{subramoniam2013local,subramoniam2015non} investigated two methods using: 1) the histograms of local binary pattern extracted from knee images and a k-Nearest neighbour classifier \cite{subramoniam2013local} and 2) Haralick features extracted from the ROI of knee images and a SVM \cite{subramoniam2015non}. Thomson et al. \cite{thomson2015automated} proposed an automated method that uses features derived from tibia and femur bone shapes, and image textures extracted from the tibia with a simple weighted sum of the outputs of two random forest classifiers. Deokar et al. \cite{deokar2015effective} investigated an artificial neural network based approach for knee OA images classification using grey level co-occurrence matrix (GLCM) textures, shape, and statistical features. Even though these methods claim high accuracy, the datasets are not publicly available and these datasets contain only a few hundred radiographs. The classification accuracies of all these methods for public datasets like the OAI and the MOST need to be studied to derive conclusive results. 

In this context, there are two approaches that use large public datasets like the OAI: 1) WNDCHRM, and 2) an artificial neural network-based scoring system. Shamir et al. proposed WNDCHRM, a multi purpose medical image classifier to automatically assess knee OA severity in radiographs\cite{shamir2009knee,shamir2009early}.  A set of features based on polynomial decompositions, high contrast, pixel statistics, and textures are used in WNDCHRM. Besides extracting features from raw image pixels, features extracted from image transforms like Chebyshev, Chevbyshev-Fourier, Radon, and Gabor wavelets are included to expand the feature space \cite{shamir2009knee,shamir2008wndchrm,orlov2008wnd}. From the entire feature space, highly informative features are selected by assigning feature weights based on a Fisher discriminant score for all the extracted features \cite{shamir2009knee,shamir2009early,orlov2008wnd}. WNDCHRM uses a variant of the k-Nearest Neighbour classifier. 

Yoo et al. \cite{yoo2016simple} have built a self-assessment scoring system and an artificial neural network (ANN) model for radiographic and symptomatic knee OA risk prediction. In a recent approach, Tiulpin et al. \cite{tiulpin2018automatic} presented a new computer-aided diagnostic approach based on deep Siamese CNNs, which were originally designed to learn a similarity metric between pairs of images. However, rather than comparing image pairs, the authors extend this idea to similarity in knee x-ray images (with 2 symmetric knee joints). Splitting the images at the central position and feeding both knee joints into a separate CNN branch allows the network to learn identical weights for both branches. They outperform the previous approaches by achieving an average multi-class testing accuracy score of 66.7 \% on the entire OAI dataset. %, despite also needing  a localization step to focus the network branches on the knee joint areas \cite{gorriz2018assessing}.

\subsection{Discussion}
According to the literature, the automatic quantification of knee OA severity involves two steps: 1) automatically detecting the ROI, and 2) classifying the detected knee joints. Many previous studies investigated automatic methods for both localisation and classification of knee joint images, but still these tasks remain a challenge.

The common approaches in the literature for automatic detection of knee joints in radiographs include template matching \cite{shamir2009knee,anifah2011automatic}, active shape models and morphological operations \cite{podsiadlo2008automated}, and a classifier-based sliding window method\cite{tiulpin2017novel}. Template matching and active shape models based approaches do not generalise well and are slow for large datasets. Classifier-based methods that use hand-crafted features are subjective and the classification accuracy is influenced by the choice of extracted features. Therefore, there is still a need for an automated method for detecting knee joints in radiographs which gives high accuracy and precision. A deep learning based method for this is investigated in this chapter.

There are several approaches in the literature for knee OA image classification that have extracted and tested many image features, such as Haralick textures \cite{subramoniam2015non}, Gabor textures \cite{thomson2015automated}, GLCM textures \cite{deokar2015effective}, local binary patterns \cite{subramoniam2013local}, shape, and statistical features of knee joints \cite{deokar2015effective}. There is even an approach that uses a large set of features based on pixel statistics, object and edge statistics, texture, histograms, and transforms \cite{park2013practical,orlov2008wnd,shamir2008wndchrm}. Different classifiers have been tested for knee OA images classification such as k-Nearest Neighbour \cite{subramoniam2013local,shamir2009knee}, SVM \cite{subramoniam2015non}, and random forest classifiers \cite{thomson2015automated}. However, all these approaches have achieved low multi-class classification accuracy, and in particular classifying successive grade knee OA images still remains a challenging task.  There is a need for a highly accurate real world automated system that can be used as a support system by clinicians and medical practitioners for knee OA diagnosis.

In recent years, many methods using manually designed or hand-crafted features have been outperformed by approaches that learn feature representations using deep neural networks. In particular, convolutional neural networks (CNN) have become highly successful in many computer vision tasks like object detection, face recognition, content based image retrieval, pose estimation, and shape recognition, and even in medical applications such as knee cartilage segmentation in MRI scans\cite{prasoon2013deep}, brain tumour segmentation in magnetic resonance imaging (MRI) scans\cite{havaei2017brain}, multi-modality iso-intense infant brain image segmentation\cite{zhang2015deep}, pancreas segmentation in CT images \cite{roth2015deep}, and neuronal membrane segmentation in electron microscopy images \cite{ciresan2012deep}. CNNs for automatically quantifying knee OA severity is investigated in this work. The next section introduces the public knee OA datasets.

%--------------------------------------------------------------------------%

\section{Public Knee OA Datasets}
\begin{figure}[t]
  \centering
  \includegraphics[width=0.6\textwidth,height=0.6\textwidth]{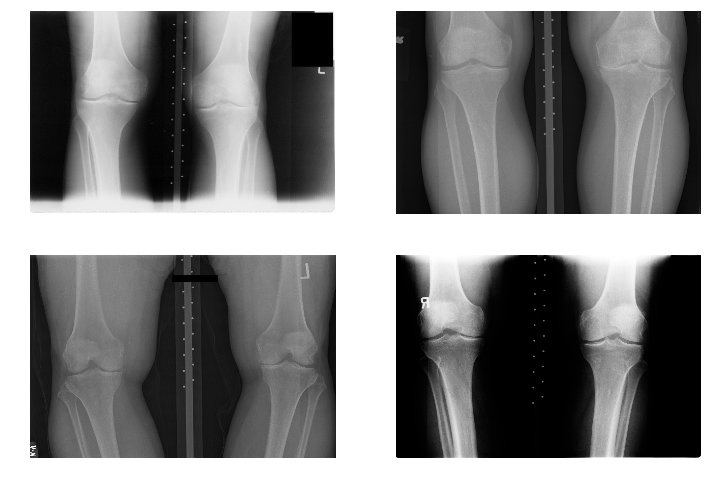}
  \caption{Samples of bilateral PA fixed flexion knee OA radiographs.}
  \label{fig:sam}
\end{figure}

The data used for the experiments and analysis in this study are bilateral PA fixed flexion knee X-ray images. Figure~\ref{fig:sam} shows some samples of knee X-ray images from the dataset. Due to variations in X-ray imaging protocols, there are some visible artefacts in the X-ray images (Figure~\ref{fig:sam}).

The datasets are from the Osteoarthritis Initiative (OAI) and Multicenter Osteoarthritis Study (MOST) in the University of California, San Francisco. These are standard public datasets used in knee osteoarthritis studies.

\subsection{OAI Dataset}
\begin{figure}[ht]
\centering
\includegraphics[width=0.9\textwidth, height=0.4\textwidth]{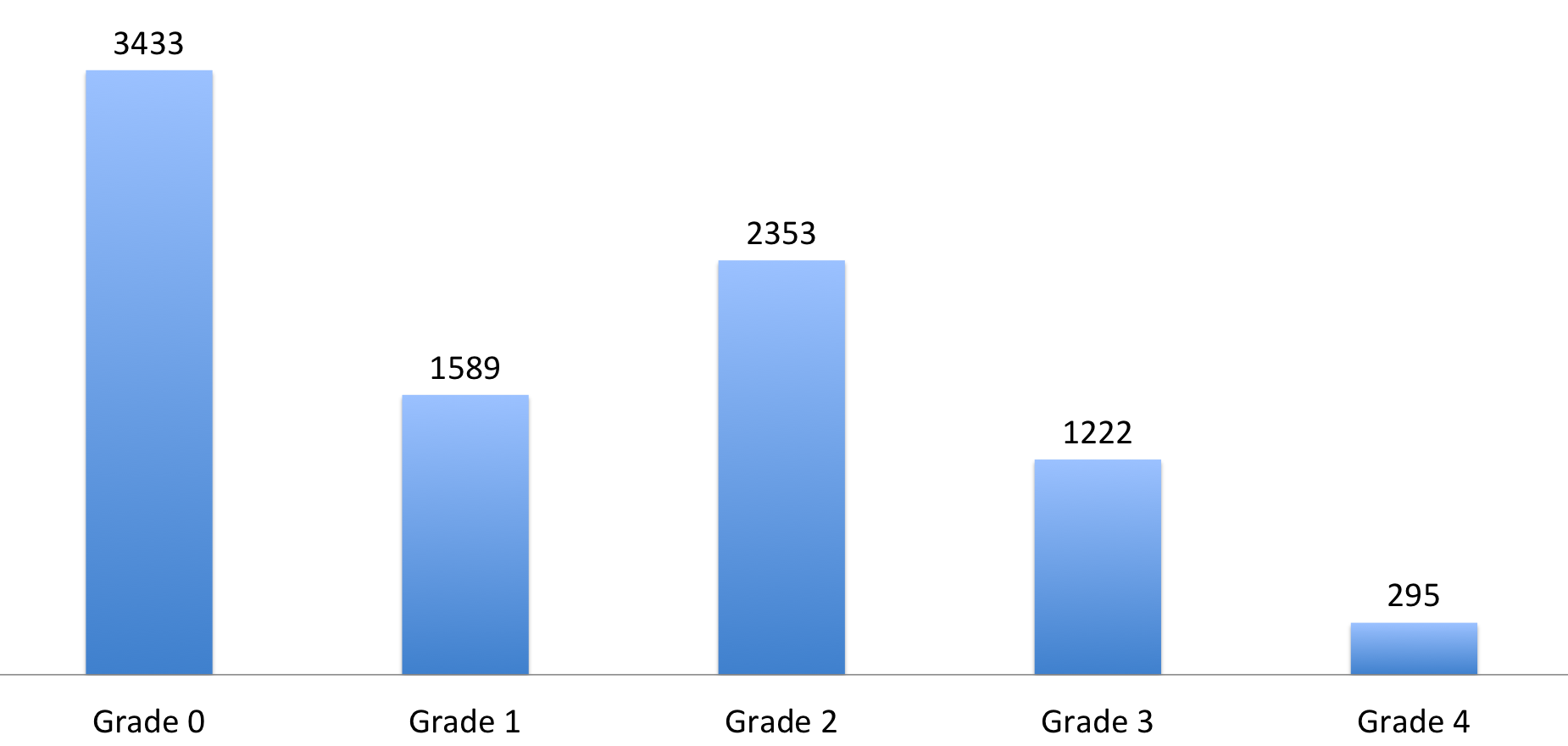}
\caption{The OAI baseline data set distribution based on KL grades.} 
\label{fig:OAI_Dist}
\end{figure}

The baseline cohort of the OAI dataset contains MRI and X-ray images of 4,746 participants. In total 4,446 X-ray images are selected from the entire cohort based on the availability of KL grades for both knees as per the assessments by Boston University X-ray reading centre (BU). In total there are 8,892 knee images. Figure \ref{fig:OAI_Dist} shows the distribution as per the KL grades. 

\subsection{MOST Dataset}
\begin{figure}[ht]
\centering
\includegraphics[width=0.9\textwidth,height=0.4\textwidth]{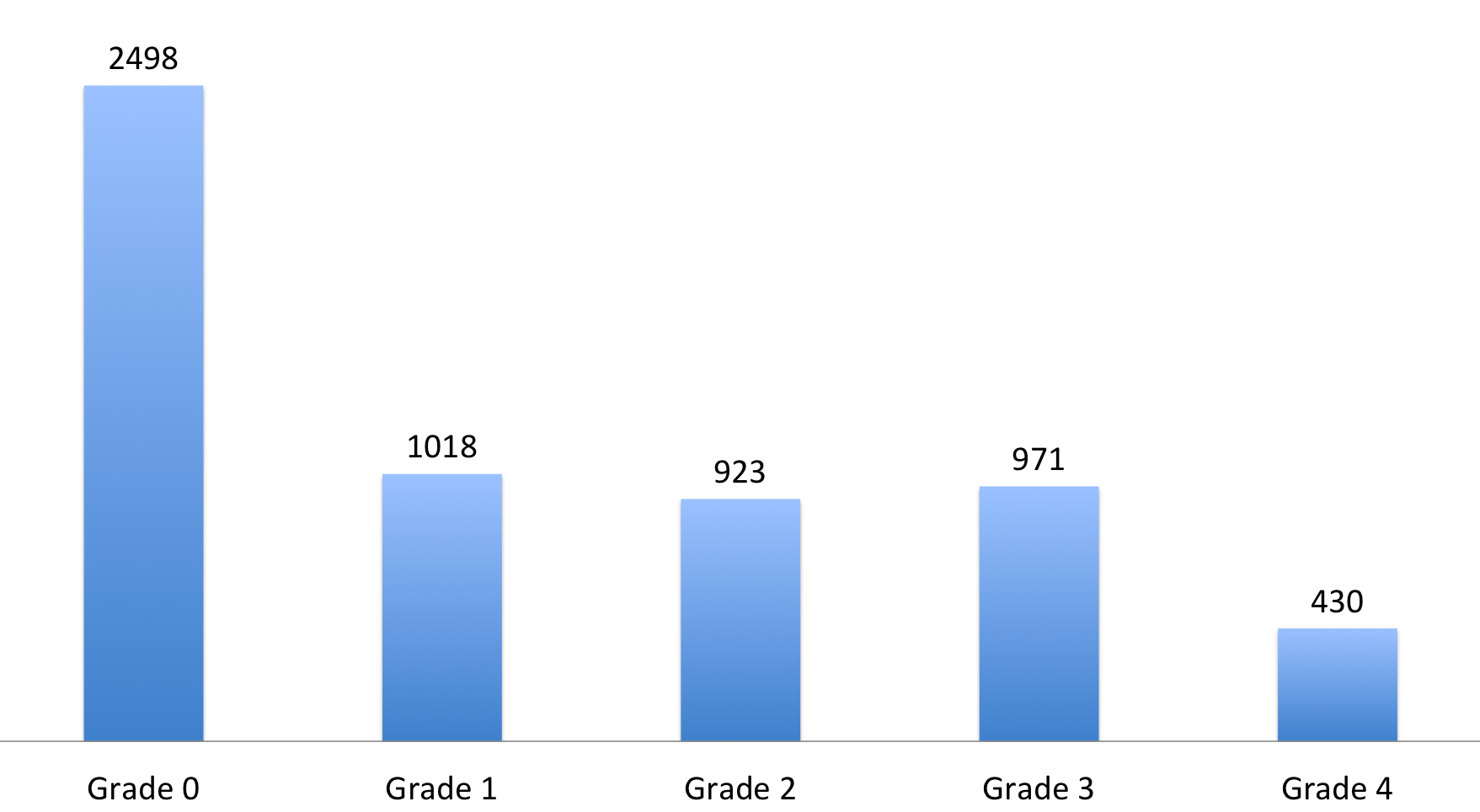}
\caption{The MOST data set distribution based on KL grades.} 
\label{fig:MOST_Dist}
\end{figure}

The MOST dataset includes lateral knee radiograph assessments of 3,026 participants. In total 2,920 radiographs are selected in this study based on the availability of KL grades for both knees as per baseline to 84-month longitudinal knee radiograph assessments. There are 5,840 knee images in this dataset. Figure \ref{fig:MOST_Dist} shows the distribution as per KL grades.

%--------------------------------------------------------------------------%

\section{Automatic Detection of Knee Joints}
\begin{figure}[t]
  \centering
  \includegraphics[width=0.6\textwidth, height=0.45\textwidth]{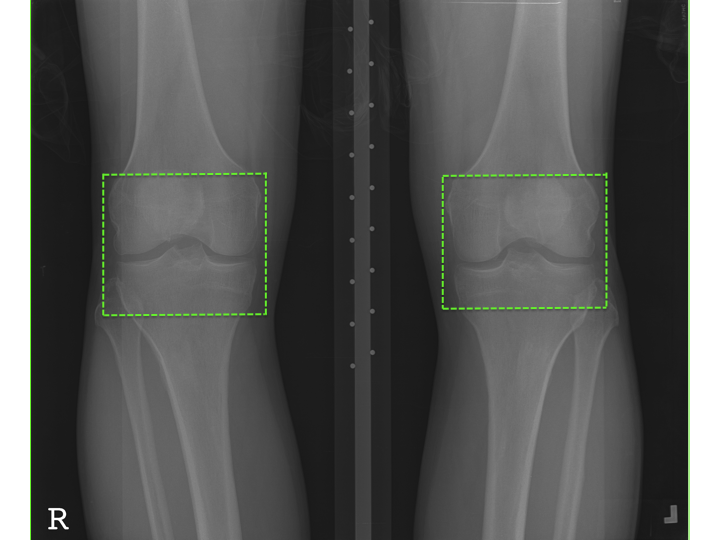}
  \caption{A knee OA X-ray image with the region of interest: the knee joints.}
  \label{fig:ROI}
\end{figure}

Classification of knee OA images and the assessment of severity conditions can be achieved by examining the characteristic features of knee OA: variations in the joint space width and the osteophytes (bone spurs) formations in the knee joints \cite{oka2008fully}. Radiologists and medical practitioners examine only the knee joint regions in the X-ray images to assess knee OA. Hence, the region of interest (ROI) for classifying knee OA images is only the knee joint regions (left and right knees). Figure \ref{fig:ROI} shows the ROI in a X-ray image. The author believes that it is better to focus on the ROI instead of the entire X-ray image for accurate classification and this is also computationally economical. For these reasons, automatically detecting and extracting the knee joint regions from the X-ray images becomes an essential pre-processing step, before classification.  

\subsection{Baseline Methods}
First, template matching for automatic detection of the knee joints \cite{shamir2009early} is implemented as a baseline. Next, the author proposes an SVM-based automatic detector for this. The implementation details and outcomes of these methods are discussed in this section. 

\subsubsection{Template Matching}~\\
In digital image processing, template matching is a technique for finding portions of an image that are similar to a standard template image. Shamir et al. \cite{shamir2009early} proposed this approach for automatically detecting the centre of the knee joints. As a baseline, the template matching approach is adapted. The steps involved in this method are as follows:

\begin{itemize}
\item First, the radiographs are downscaled to 10\% of the original size and subjected to histogram equalisation for intensity normalisation. This step is followed as proposed by Shamir et al. \cite{shamir2009early}.

\item An image patch (20$\times$20 pixels) containing the centre of the knee joint is taken as a template. 5 image patches are taken from each grade, so that in total 25 patches are pre-selected as templates. Figure \ref{fig:templates} shows the pre-selected knee joint centres of size 20$\times$20 pixels extracted from the knee joint images as templates. 

\begin{figure}[t]
  \centering
  \includegraphics[width=0.6\textwidth]{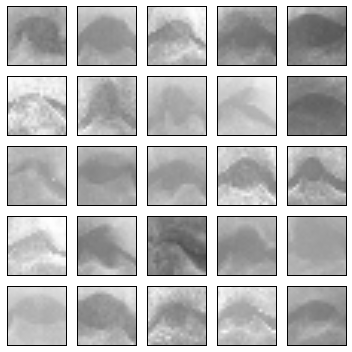}
  \caption{Pre-selected knee joint centres (20$\times$20 pixels) extracted from knee joint images for template matching.}
  \label{fig:templates}
\end{figure}

\item Each image is scanned by an overlapping (20$\times$20) sliding window. For each location at an interval of 10 pixels, distances (Euclidean) between an image patch (20$\times$20 pixels) and 25 pre-selected templates (patches with knee joint centre) are computed using;
$$dist_{i,w} = \sqrt{\sum_{y=1}^{20} \sum_{x=1}^{20} (I_{x,y} - W_{x,y})^2}, $$ where $I_{x,y}$ is the intensity of pixel $(x,y)$ in the knee joint image $I$, $W_{x,y}$ is the intensity of pixel $(x,y)$ in the sliding window, and $dist_{i,w}$ is the Euclidean distance between the knee joint image $(I)$ and the sliding window $W$.

\item In total, 25 different distances are calculated at each location of the sliding window for the 25 templates, and the shortest among the 25 distances is recorded. 

\item The window with the smallest Euclidean distance is selected as the centre of the knee joint after scanning the image with a sliding window, and a fixed size region (700$\times$500 pixels) around this centre is extracted as the knee joint region from the X-ray image. 

\item The input X-ray images are horizontally split in half to isolate left and right knees separately and the sliding window is run on both halves. 
\end{itemize}

\subsubsection*{Experiments and Results.}
For the experiments on template matching, the baseline data sample of 200 progression and incidence cohort subjects under the knee OA study is used. This dataset contains in total 191 X-ray images (382 knee joints) and it is a subset of the large OAI dataset. 
%After these experiments, we acquired the entire OAI dataset containing records of 4,796 subjects. 

In this implementation, five different sets of templates (each set with 25 templates) are used to show the influence of templates on knee joint detections. The templates are selected from a separate training set. Visual inspection is used to evaluate the results of template matching by plotting a bounding box (20$\times$20 pixels) on the image patch that recorded the shortest Euclidean distance after template matching. Table \ref{Tab:TMResults} shows the total number of true positives (the detected knee joint centres), the total number of false positives and the precision. 

\begin{table}[t]
\caption{Detection of knee joint centres using template matching method.}
\label{Tab:TMResults}
\centering
\begin{tabular}{ c c c c }
\toprule
Templates & True Positives & False Positives & Precision \\
\midrule
\midrule
Set 1 & 87 & 295 & 22.8 \% \\

Set 2 & 78 & 304 & 20.4 \% \\

Set 3 & 99 & 283 & 25.9 \%\\

Set 4 & \textbf{116} & \textbf{266} & \textbf{30.3 \%} \\

Set 5 & 55 & 327 & 14.4 \% \\
\bottomrule
\end{tabular}
\end{table}

  It is clearly evident from the results (Table \ref{Tab:TMResults}) that template matching is not precise in detecting the knee joints and that the detections are heavily dependent on the choice of templates. The number of templates was increased to 50, but still there was no further improvement in the results. The reason for low-performance of the template matching is that the computations are mainly based on the intensity level difference between an image patch and a template. There are also possibilities for image patches not around the knee joint, having the shortest Euclidean distance to a template in the set and thus, being detected as matches. In the next section, a new SVM-based method is investigated to improve the detection of the knee joints. 

\subsubsection{SVM-Based Detection}~\\
Standard template matching is not scalable and produces poor detection accuracy on large datasets like the OAI. We proposed a classifier-based model to automatically detect the knee joints in the X-ray images \cite{antony2016}. The idea is to use well-known Sobel edge detection \cite{sobel1990isotropic} for detecting the knee joints. The two major steps involved in this method are 1) training a classifier and 2) developing a sliding window detector. 

\subsubsection*{Training a Classifier.}
First, image patches (20$\times$20 pixels) are generated from the input X-ray images. The image patches containing the knee joint centre (20$\times$20 pixels) are used as positive samples and randomly sampled patches excluding the knee joint centre are used as negative samples. In total, 200 positive and 600 negative samples are used. The image patches (samples) are split into training (70\%) and test (30\%) sets. Sobel horizontal image gradients are extracted as features from all these samples to train a classifier. The powerful and well-known SVM is used for classification. A linear SVM is fitted with default parameters (C=1, and linear kernel), using Sobel horizontal image gradients as the features.

Before settling on Sobel horizontal image gradients as features, the state-of-the-art features such as histogram of oriented gradients, Tamura and Haralick textures, and the Gabor features were tested. The HOG features are highly accurate and efficient in object detection and human detection \cite{dalal2005histograms}. The Tamura, Haralick, and Gabor features are highly influential and top-ranked among the features used in WNDCHRM for knee OA image classification \cite{shamir2009knee,shamir2009early,orlov2008wnd,oka2008fully}. The Sobel operator or Sobel filter uses vertical and horizontal image gradients to emphasise the edges in images \cite{sobel1990isotropic}. From these, the horizontal image gradients are used as the features for detecting the knee joints centres. Intuitively, the knee joint images primarily contain horizontal edges that are easy to detect.

\subsubsection*{Sliding Window Detector.}
\begin{figure}[t]
  \centering
  \includegraphics[width=0.6\textwidth, height=0.4\textwidth]{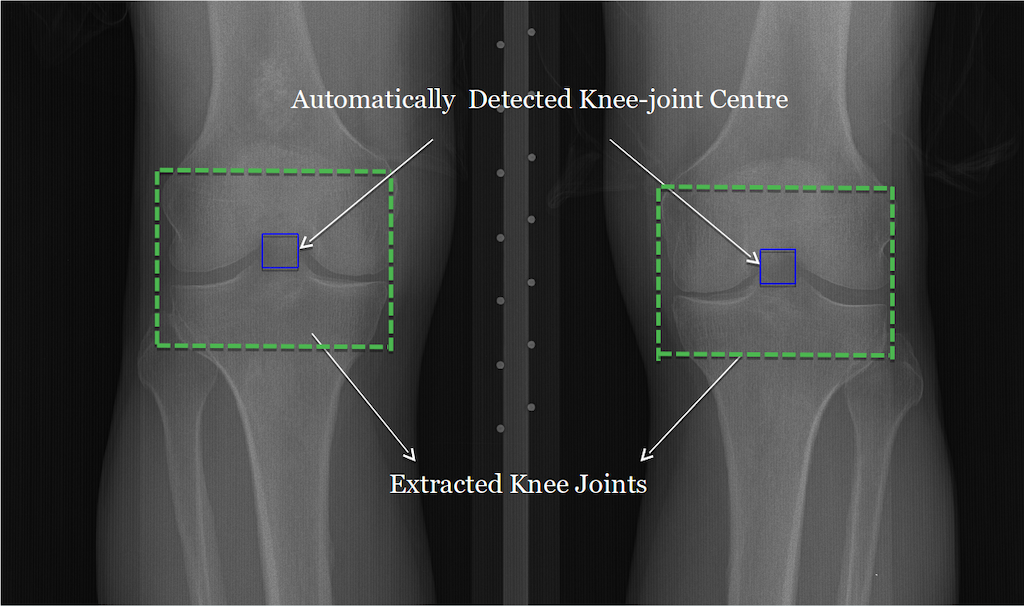}
  \caption{Detecting the knee joint centres and extracting the knee joints.}
  \label{fig:AutoDet}
\end{figure}

To detect the knee joint centre from both left and right knees, input images are split in half to isolate left and right knees separately. A sliding window (20$\times$20 pixels) is used on either half of the image, and the Sobel horizontal gradient features are extracted for every image patch. The image patch with the maximum score based on the SVM decision function is recorded as the detected knee joint centre, and the area (200$\times$300 pixels) around the knee joint centre is extracted from the input images using the corresponding recorded coordinates. Figure \ref{fig:AutoDet} shows an instance of a detected knee joint and the extracted ROI in a X-ray image.

\subsubsection*{Results and Discussion.}
In total, 200 image patches with the knee joint centres as positive samples and 600 image patches that exclude the centre of knee joint as negative samples are used. These images are split into training (70\%) and test (30\%) sets. 
Fitting a linear SVM with the training data produced a 5-fold cross validation accuracy of $\textbf{95.2\%}$ and an accuracy of $\textbf{94.2\%}$ for the test data. Table~\ref{Tab:AD} shows the precision, recall, and $F_{1}$ scores of this classification. To evaluate the automatic detection, the ground truth is generated by manually annotating the knee joint centres (20$\times$20 pixels) in 4,446 radiographs using an annotation tool that we developed, which recorded the bounding box (20$\times$20 pixels) coordinates of each annotation.

\begin{table}[t]
\caption{Classification metrics of the SVM for detection.}
\label{Tab:AD}
\centering
\begin{tabular}{ l c c c }
\toprule
Class & Precision & Recall & $F_{1}$score\\
\midrule
\midrule
 Positive & 0.93 & 0.84 & 0.88  \\
 Negative & 0.95 & 0.98 & 0.96  \\
\midrule
 Mean & 0.94 & 0.94 & 0.94\\
 \bottomrule
\end{tabular}
\end{table}

The well-known Jaccard index (JI) is used to give a matching score for each detected instance. The Jaccard index JI(A,D) is given by,
\begin{equation}
JI(A,D) = \frac{A \cap D} {A \cup D}
\end{equation}
where A, is the manually annotated and D is the automatically detected knee joint centre using the proposed method. 

\begin{table}[t]
\caption{Comparison of template matching and the proposed SVM-based method.}
\label{Tab:Jac}
\centering
\begin{tabular}{ l c  c  c }
\toprule
Method & $JI=1$ & $JI\geq0.5$ & $JI>0$\\
\midrule
\midrule
 Template Matching & 0.3 \% & 8.3 \% & 54.4 \% \\
 Proposed Method & 1.1 \% & 38.6 \% & \textbf{81.8 \%} \\
\bottomrule
\end{tabular}
\end{table}

Table~\ref{Tab:Jac} shows the resulting average detection accuracies based on thresholding of Jaccard indices. The mean JI for the template matching and the classifier methods are $\textbf{0.1}$ and $\textbf{0.36}$. From Table~\ref{Tab:Jac}, it is evident that the proposed method is more accurate than template matching. This is due to the fact that template matching relies upon the intensity level difference across an input image. Thus, it is prone to matching a patch with small Euclidean distance that does not actually correspond to the knee joint centre. Also, the templates are varied in a set, and it is observed that the detection is highly dependent on the choice of templates. Template matching is similar to a k-nearest neighbour classifier with $k=1$. 

The reason for higher accuracy in the proposed method is the use of horizontal edge detection instead of intensity level differences. The knee joints primarily contain horizontal edges and thus are easily detected by the classifier using horizontal image gradients as features. The proposed method is approximately $80\times$ faster than template matching; for detecting all the knee joints in the dataset comprising $4,446$ radiographs, the proposed method took $\sim$9 minutes and the template matching method took $\sim$798 minutes.

Despite sizeable improvements in accuracy and speed using the proposed approach, detection accuracy still falls short. Therefore, manual annotations for the incorrect detections from this method were substituted to investigate KL grade classification performance independently of knee joint detection. The next Section describes the proposed methods for automatically localising the knee joint region using fully convolutional neural networks. 

\subsection{Fully Convolutional Network Based Detection}

A typical CNN architecture consists of three main types of layers: convolutional, pooling and fully-connected or dense layers. A fully convolutional network (FCN) is similar to a CNN, but the fully-connected layers are replaced by convolutional layers \cite{long2015fully}. A FCN consists of mostly convolutional layers and if pooling layers are used, then suitable up-sampling layers are added before the last convolutional layer. The two major differences of FCNs over CNNs can be summarised as:

\begin{itemize}
\item FCNs are trained end-to-end to make pixel-wise predictions \cite{long2015fully}. Even the decision-making layers at the last stage of the network use learned convolutional filters. 

\item The input image size need not be fixed as there are no fully-connected layers in the FCN. CNNs with fully connected layers can operate only on a fixed size input. 
\end{itemize}

FCNs have achieved great success in semantic segmentations of general images \cite{long2015fully}. Recent approaches using FCNs for medical image segmentation show promising results \cite{milletari2016v,kayalibay2017cnn,christ2017automatic}. Motivated by this, the use of FCN is investigated in this chapter for automatically detecting the knee joints. Two approaches are developed for localising the knee joints: 1) training a FCN to detect the centre of knee joints and extract a fixed-size region around the detected centre, and 2) training a FCN to detect the ROI and thus extract the knee joints directly.

\subsubsection{Localisation with Reference to Knee Joint Centre}
\begin{figure}[t]
\centering
\includegraphics[width=1.0\textwidth] {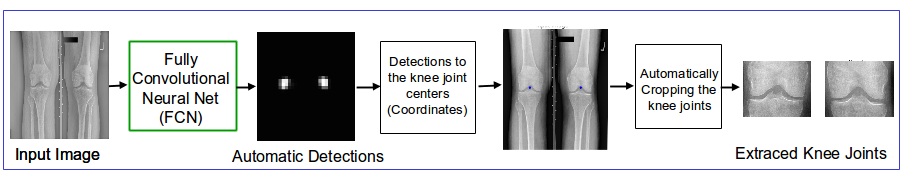}
\caption{Automatic localisation of knee joints with reference to the centre of the knee joints.} 
\label{fig:BD_Local}
\end{figure}

~\\
In the initial approach to localise the knee joints in X-ray images using a FCN, a similar strategy to template matching and the SVM based methods is followed; that is to detect the centre of knee joints and to extract the ROI with reference to the detected centres. Figure \ref{fig:BD_Local} shows the steps involved in this method: training a FCN to detect the knee joint centres (20$\times$20 pixels), computing the coordinates of the centres from the FCN output, and extracting a fixed size region as knee joints. In the next section, the experimental data and the ground truth used to train the FCNs are introduced.

\subsubsection*{Dataset and Ground Truth Generation.}
\begin{figure}[t]
  \centering
  \subfloat[]{\includegraphics[width=0.4\textwidth]{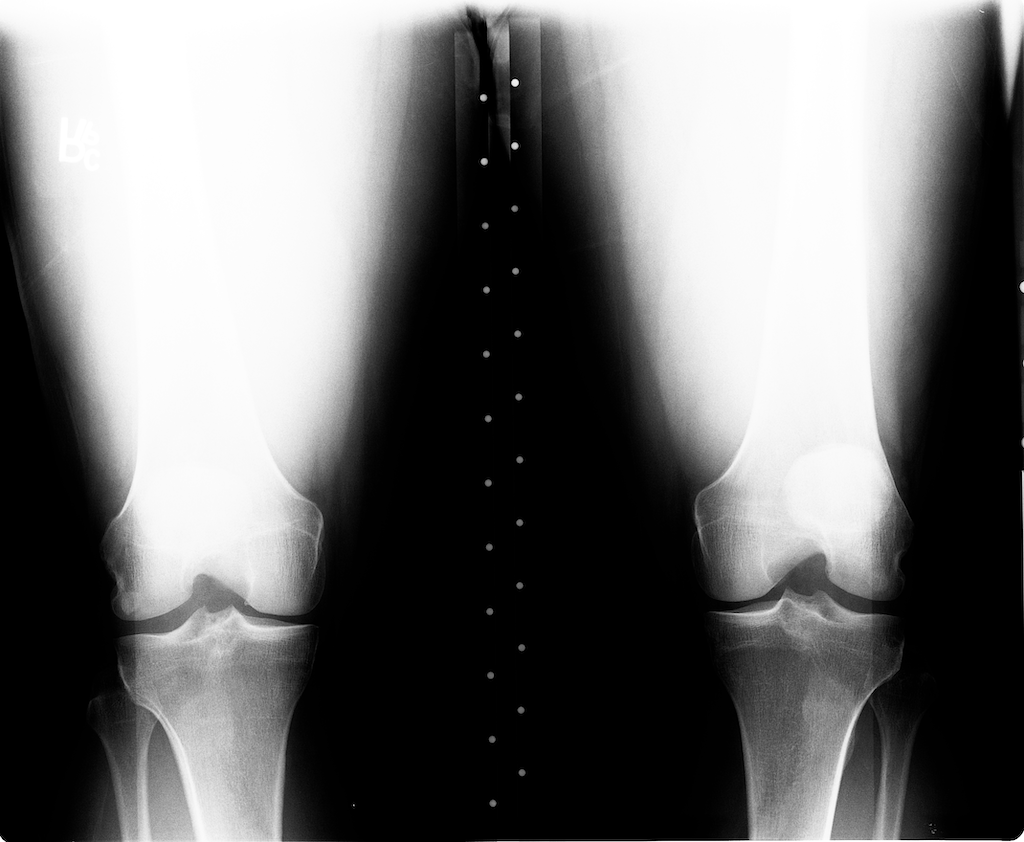}}
  \hspace{1em}
  \subfloat[]{\includegraphics[width=0.4\textwidth]{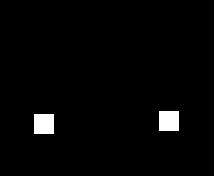}}
  \caption{(a) An input X-ray image and (b) The binary mask annotations for knee joint centres.}
  \label{fig:gt_fcn}
\end{figure}

The data used for the experiments are taken from the baseline cohort of the OAI dataset. In total 4,446 X-ray images are selected from the entire dataset based on the availability of KL grades for both knee joints. The knee joint centres in all these X-ray images are manually annotated, after downscaling to 10\% of the actual size. Binary masks of size 20$\times$20 pixels are marked around the knee joint centres using the annotations. Figure \ref{fig:gt_fcn} shows an instance of an input X-ray image and the binary mask annotations corresponding to the knee joint centres. The image patches from the masked region i.e. the knee joint centres, are taken as positive training samples and the patches from rest of the image are taken as the negative training samples to train an FCN. The dataset is split into training (3,333 images) and test (1,113 images) sets. 

\subsubsection*{Training Fully Convolutional Neural Networks}˜\\
%\subsubsection*{Initial Configuration and Training.}
To start, a FCN is configured with a lightweight architecture containing 4 convolutional layers followed by a fully convolutional layer, which is a convolutional layer with a kernel size [1$\times$1] and that uses a \textit{sigmoid activation}. FCNs use fully convolutional layers at the last stage to make pixel-wise predictions \cite{long2015fully}. Table \ref{Tab:FCN1} shows the network configuration in detail. Each convolution layer is followed by a ReLU layer.

\begin{table}[ht]
\caption{Initial FCN Configuration for detecting the knee joint centres.}
\label{Tab:FCN1}
\centering
\begin{tabular}{l c c }
\toprule
Layer & Kernel & Kernel Size\tabularnewline
\midrule
\midrule
Conv1 & 32 & 3$\times$3\tabularnewline
Conv2 & 32 & 3$\times$3\tabularnewline
Conv3 & 64 & 3$\times$3\tabularnewline
Conv4 & 64 & 3$\times$3\tabularnewline
Conv5 & 1 & 1$\times$1\tabularnewline
\bottomrule
\end{tabular}
\end{table}

\begin{figure}[t]
\centering
\includegraphics[width=0.8\textwidth] {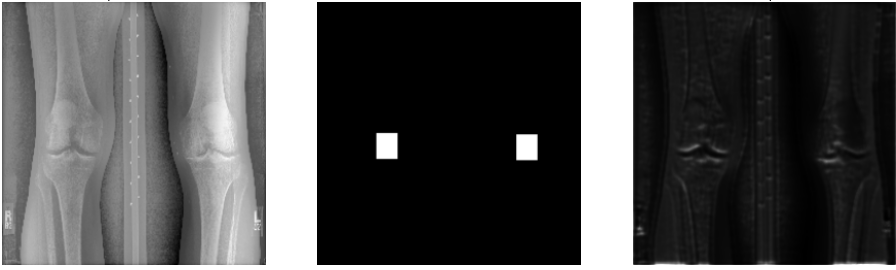}
\caption{An instance of input, ground truth and output (predictions) of FCN.} 
\label{fig:fcn_out1}
\end{figure}

The network parameters are trained from scratch with training samples of knee OA radiographs from the OAI dataset. The dataset is split into training (3,333 images) and test (1,113 images) sets. The ground truth for training the network are binary images with masks specifying the ROI: the knee joints. The network is trained to minimise the total \textit{binary cross entropy} between the predicted pixels and the ground truth. \textit{Stochastic gradient descent} (SGD) with default parameters: learning rate = $0.01$, decay = $1\mathrm{e}^{-6}$, momentum = $0.9$, and nesterov = True, is used. The network is trained for 40 epochs and the batch size is 10. Figure \ref{fig:fcn_out1} shows an instance of the test input, the ground truth and the output (pixel-wise predictions) of the FCN. From the predictions of this FCN, it is observed that the network is able to slightly detect the edges of the knee joints and these are promising initial results. In an attempt to improve the detections, the FCN configurations are experimented and for this the hyper-parameters of the network are tuned.

\subsubsection*{Receptive Field.}
When dealing with high-dimensional inputs such as images, it is impractical to connect neurons in the current level to all the neurons in the previous volume. Instead, each neuron is only connected to a local region of the input volume. The spatial extent of this connectivity is a hyper-parameter called the receptive field of the neuron \cite{cs231n}. The receptive field size, otherwise termed the effective aperture size of a CNN, shows how much a convolutional node sees of the input pixels (patch) that affects a node's output. The effective aperture size depends on kernel size and strides of the previous layers. For instance, a 3$\times$3 kernel can see a 3$\times$3 patch of the previous layer and a stride of 2 doubles what all succeeding layers can see. 

The receptive field size of neurons in the final layer of the FCNs is calculated and used to analyse the output of FCNs and the overall detection results. The receptive field size of a neuron in the final layer (Conv5) of the initial FCN configuration (Table 4.1) is 9, which is low and may be a reason for poor performance of this network. Larger convolutional kernel sizes to increase the receptive field of the network is investigated. The forthcoming Section will show that a network (Table \ref{Tab:FCN5}) with larger receptive field gives the best results for detecting the knee joint centres.  

\subsubsection*{Tuning the FCN Hyper-parameters.}
VGG-M-128 \cite{simonyan2014very}, the deep convolutional neural network developed by the Oxford visual geometry group (VGG) uses kernel size 7$\times$7 in the first convolutional layer and 5$\times$5 in the following convolutional layer. Inspired by this, kernel sizes of 5$\times$5, and 7$\times$7 for the first convolutional layer are tested retaining the other settings. The kernel size 7$\times$7 gives better results in this configuration. This is because of the larger receptive field size of the 7$\times$7 kernel in comparison to the 3$\times$3 kernel. 

Next, the experiments are conducted by varying the number of convolutional layers and also the number of filters (kernel) in a convolutional layer, before obtaining the configuration that gave the best results based on visual observations. Table \ref{Tab:FCN2} shows the configuration of the network derived from the initial configuration and the receptive field size of a neuron in the final layer (Conv4) is 11. The networks are trained with 3,333 images and tested on 1,113 images from the OAI dataset. 

\begin{table}[t]
\caption{FCN for detecting the knee joint centres.}
\label{Tab:FCN2}
\centering
\begin{tabular}{l c c }
\toprule
Layer & Kernel & Kernel Size\tabularnewline
\midrule
\midrule
Conv1 & 32 & 7$\times$7\tabularnewline
Conv2 & 64 & 3$\times$3\tabularnewline
Conv3 & 96 & 3$\times$3\tabularnewline
Conv4 (fullyConv) & 1 & 1$\times$1\tabularnewline
\bottomrule
\end{tabular}
\end{table}

\begin{figure}[t]
\centering
\includegraphics[width=0.8\textwidth] {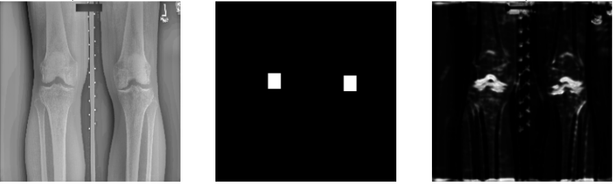}
\caption{An input image, ground truth, and outcome of the final FCN.} 
\label{fig:fcn_out2}
\end{figure}

There is an improvement in the detections using this network in comparison to the previously tested configurations. Figure \ref{fig:fcn_out2} shows an instance of the output predictions of this network. To quantitatively evaluate the automatic detections, the well-known Jaccard Index is used.

\subsubsection*{Quantitative Evaluation.}
A simple contour detection is used and the Jaccard index i.e. the overlap statistics calculated by the Intersection over Union (IoU) to evaluate the automatic detections of the FCN. The steps involved are as follows:

\begin{itemize}
\item First, the objects are detected i.e. the knee joint regions from the output image of the FCN using simple contour detection \cite{kilian2001simple}. Contours can be explained simply as a curve joining all the continuous points (along the boundary), having the same colour or intensity. The contours are a useful tool for shape analysis and simple object detection and recognition. In this method, first the images are converted to binary by applying Otsu's threshold. Next, the contours of the objects or shapes in the binary image are automatically detected and recorded \cite{kilian2001simple}.

\item Next, the detected objects in the image are sorted based on the area and from these the top two are selected. This is to eliminate noise or other faint edges picked up by the FCN. 
\item The centroids of the largest two detected regions are recorded as the knee joint centres.
\item A binary mask of 20$\times$20 pixels size is marked around each detected knee joint centre.
\item The Jaccard index is computed for each image with the masks of predicted centres and the masks predefined using manual annotation i.e. the labels used for training FCN.
\end{itemize}

In total 1,113 X-ray images (2,226 knee joints) are included in the test set. The FCN with the final configuration detects 1,851 knee joints in the test set with Jaccard index $\geq0.5$, the accuracy of detection is \textbf{83.2\%} with a mean 0.66 and standard deviation 0.18. This is an improvement in comparison to previous approaches but still falls short of perfect detections. The pooling and up-sampling layers in the FCN are varied and experimented in an attempt to improve the detection accuracy. This may help to increase the receptive field size and in turn improve the overall detections. 

\subsubsection*{FCN with Pooling and Up-sampling Layers.}
Two max pooling layers with stride 2 and up-sampling by a factor of 4 are included to the previous configuration (Table \ref{Tab:FCN2}). Table \ref{Tab:FCN3} shows the FCN architecture in detail. Each convolutional layer is followed by a ReLU activation. 
 
\begin{table}[t]
\caption{FCN with pooling and up-sampling layers.}
\label{Tab:FCN3}
\centering
\begin{tabular}{l c c c c }
\toprule 
Layer & Kernel & Kernel Size & Strides\tabularnewline
\midrule
\midrule
Conv1 & 32 & 7 $\times$7 & 1 \tabularnewline
MaxPool2 & -- & 2$\times$2 & 2 \tabularnewline
Conv3 & 64 & 3$\times$3 & 1 \tabularnewline
MaxPool4 & -- & 2$\times$2 & 2 \tabularnewline
Conv5 & 96 & 3$\times$3 & 1 \tabularnewline
UpSamp6 & -- & 4$\times$4 & 1 \tabularnewline
Conv7 (fullyConv) & 1 & 1$\times$1 & 1 \tabularnewline
\bottomrule
\end{tabular}
\end{table}

\begin{figure}[t]
\centering
\includegraphics[width=0.8\textwidth] {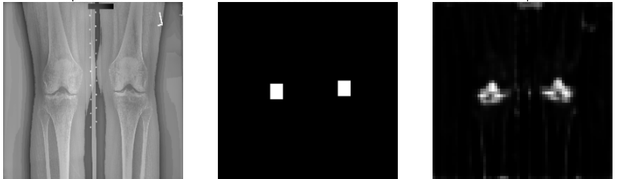}
\caption{Prediction of the FCN with max pooling and up-sampling layers.} 
\label{fig:fcn_out3}
\end{figure}

Figure \ref{fig:fcn_out3} shows the output of this network for a test image. On visual observation, the output image contains less noise and the detections are improving compared to the previous approaches, even though the output image resolution is low. This is due to the inclusion of pooling and up-sampling stages to the network and this has increased the receptive field size of the final layer (Conv7) to 34. The number of convolutional-pooling stages is increased, to see if there is improvement in the detections. Table \ref{Tab:FCN4} shows the architecture of this network in detail. 

\begin{table}[t]
\caption{FCN with 3 Convolution-Pooling stages for detecting the knee joint centres.}
\label{Tab:FCN4}
\centering
\begin{tabular}{l c c c c }
\toprule
Layer & Kernel & Kernel Size & Strides\tabularnewline
\midrule 
\midrule
Conv1 & 32 & 7$\times$7 & 1 \tabularnewline
MaxPool2 & -- & 2$\times$2 & 2 \tabularnewline
Conv3 & 32 & 3$\times$3 & 1 \tabularnewline
MaxPool4 & -- & 2$\times$2 & 2 \tabularnewline
Conv5 & 64 & 3$\times$3 & 1 \tabularnewline
MaxPool6 & -- & 2$\times$2 & 2 \tabularnewline
Conv7 & 96 & 3$\times$3 & 1 \tabularnewline
UpSamp8 & -- & 8$\times$8 & 1 \tabularnewline
Conv9 (fullyConv) & -- & 1$\times$1 & 1 \tabularnewline
\bottomrule
\end{tabular}
\end{table}

\begin{figure}[t]
\centering
\includegraphics[width=0.8\textwidth] {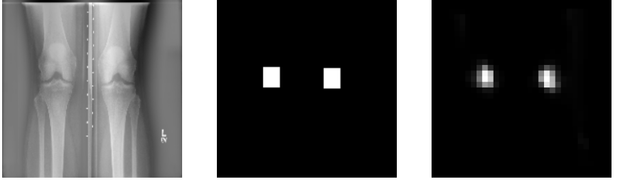}
\caption{Predictions of the FCN with 3 Convolution-Pooling stages.} 
\label{fig:fcn_out4}
\end{figure}

From the output of this FCN, it can be observed that the detections become more precise in comparison to the previous networks even though the resolution is low in comparison to the previous networks. Figure \ref{fig:fcn_out4} shows an instance of the input test image, ground truth and the FCN output. 

The outcomes of this FCN are evaluated using Jaccard index and the detection accuracy is  \textbf{96.7\%}, that is in total 2,152 out of 2,226 knee joints are detected with a Jaccard index $\geq$ 0.5. The Jaccard index mean is 0.74 and standard deviation is 0.13. The detection accuracy is high in comparison to the previous networks. Table \ref{Tab:res_fcn4} shows the detection accuracy of the FCN for the Jaccard index values at 0.25, 0.5 and 0.75.

\begin{table}[t]
\caption{Detection accuracy of FCN based on Jaccard Index.}
\label{Tab:res_fcn4}
\centering
\begin{tabular}{l c c c}
\toprule
Jaccard Index & JI $\geq$ 0.25 & JI $\geq$ 0.5 & JI $\geq$ 0.75\tabularnewline
\midrule
\midrule
Detection Accuracy & 98.5 \% & 96.7 \% & 39.6 \%\tabularnewline
\bottomrule
\end{tabular}
\end{table}

This FCN (Table \ref{Tab:FCN4}) has three convolutional-pooling stages. A configuration with 4 convolutional-pooling stages followed was tested by adding an up-sampling layer with kernel size (16$\times$16). There was no improvement in the detection accuracy for this configuration. 

\subsubsection*{Best Performing FCN for Detecting the Knee Joint Centres.} 
Before settling on the final architecture, experiments were done by varying the number of convolution stages, the number of filters and kernel sizes in each convolution layer. The best performing FCN (Table \ref{Tab:FCN5}) was selected based on a high detection accuracy on the test data. This network was trained with the OAI dataset containing 4,444 knee radiographs. The dataset was split into a training set containing 3,333 knee images and test set containing 1,113 knee images. The validation set (10\%) was taken from the training set. The effective aperture size of this FCN (Table \ref{Tab:FCN5}) for a node in the last convolutional layer (before up-sampling) is 66. The aperture size for the previous networks shown in Table \ref{Tab:FCN4} is 42 and Table \ref{Tab:FCN3} is 34. For the other tested configurations the effective aperture size is even lower (less than 30).

\begin{table}[t]
\caption{Best performing FCN for detecting the knee joint centres.}
\label{Tab:FCN5}
\centering
\begin{tabular}{ l c c c }
\toprule 
Layer & Kernel & Kernel Size & Strides\tabularnewline
\midrule
\midrule 
Conv1 & 32 & 3$\times$3 & 1 \tabularnewline
MaxPool1 & -- & 2$\times$2 & 2 \tabularnewline

Conv2\_1 & 32 & 3$\times$3 & 1 \tabularnewline
Conv2\_2 & 32 & 3$\times$3 & 1 \tabularnewline
MaxPool2 & -- & 2$\times$2 & 2 \tabularnewline

Conv3\_1 & 64 & 3$\times$3 & 1 \tabularnewline
Conv3\_2 & 64 & 3$\times$3 & 1 \tabularnewline
MaxPool3 & -- & 2$\times$2 & 2 \tabularnewline

Conv4\_1 & 96 & 3$\times$3 & 1 \tabularnewline
Conv4\_1 & 96 & 3$\times$3 & 1 \tabularnewline

UpSamp5 & -- & 8$\times$8 & 1 \tabularnewline
Conv5 (fullyConv) & -- & 1$\times$1 & 1 \tabularnewline
\bottomrule
\end{tabular}
\end{table}

Table \ref{Tab:FCN5} shows the configuration of the best performing FCN for detecting the knee joint centres. This FCN is based on a lightweight architecture and the network parameters (in total 214,177) are trained from scratch. The network consists of 4 stages of convolutions with a max-pooling layer after each convolutional stage, and the final stage of convolutions is followed by an up-sampling and a fully-convolutional layer. The network uses a uniform [3$\times$3] convolution and [2$\times$2] max pooling. Each convolution layer is followed by a ReLU activation layer. After the final convolution layer, an [8$\times$8] up-sampling is performed as the network uses 3 stages of [2$\times$2] max pooling. The up-sampling is essential for an end-to-end learning by back propagation from the pixel-wise loss and to obtain pixel-dense outputs \cite{long2015fully}; when pooling layer(s) and strides more than one are used in the network. The final layer is a fully convolutional layer with a kernel size of [1$\times$1] and uses a sigmoid activation for pixel-based classification. The input to the network is of size [256$\times25$6]. 

\begin{figure}[t]
\centering
\includegraphics[scale=0.6]{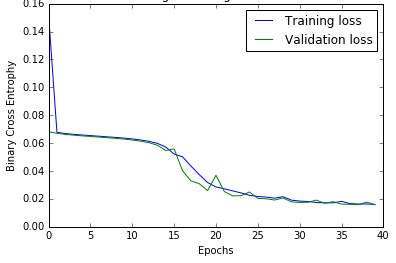}
\caption{Training and validation losses of the FCN.} 
\label{fig:LC_FCN}
\end{figure}

This network was trained to minimise the total binary cross entropy between the predicted pixels and the ground truth using \textit{stochastic gradient descent} (SGD) with default parameters: learning rate = $0.01$, decay = $1\mathrm{e}^{-6}$, momentum = $0.9$, and nesterov = True. This network was trained for 40 epochs with a batch size of 32. The validation (10\%) data was taken from the training set. Figure \ref{fig:LC_FCN} shows the learning curves when training this network and decrease in the validation and training losses.

Table \ref{Tab:res_fcn5} shows the results of the best performing FCN. This network achieved a detection accuracy of \textbf{97.1\%}, in total 2,162 knee joints out of the 2,226 test samples detected with a Jaccard index 0.5. The Jaccard index mean is 0.76 and standard deviation is 0.12. 

\begin{table}[t]
\caption{Detection accuracy of the best performing FCN.}
\label{Tab:res_fcn5}
\centering
\begin{tabular}{l c c c }
\toprule
Jaccard Index & JI $\geq$ 0.25 & JI $\geq$ 0.5 & JI $\geq$ 0.75\tabularnewline
\midrule
\midrule
Detection Accuracy & 98.9 \% & 97.1 \% & 43.3 \% \tabularnewline
\bottomrule 
%Number of Detections & 2,224 & 2,193 & 962 \tabularnewline
%\hline
\end{tabular}
\end{table}

\subsubsection*{Error Analysis.}
The results of the best performing FCN (Table \ref{Tab:res_fcn5}) show 99\% detection accuracy for a Jaccard index $\geq$ 0.1, in total 2,205 out of 2,226 knee joints are successfully detected. On observing the failed detections: 1\% (in total 21 knee joints), there are two patterns.

\begin{itemize}
\item[1.] The output of the FCN is very faint or no detections at all. Figure \ref{fig:fcnErr12} shows two instances of input X-ray images, masks defining the knee joint centres as ground truth, and output of the best performing FCN with faint detections. The input images with variations in the local contrast and local luminance due to the imaging protocol variations appear to be the main cause for this error. Histogram equalisation is used as a pre-processing step to adjust the contrast of the input images. Even though this adjusts the contrast globally in an image, there are still contrast variations in portions of the image. Local contrast enhancement algorithms \cite{bressan2007local} or adaptive histogram equalisation \cite{pizer1987adaptive} can be used to normalise the images for variations in the local contrast and local luminance.  

\begin{figure}[t]
  \centering
  \subfloat{\includegraphics[width=0.8\textwidth]{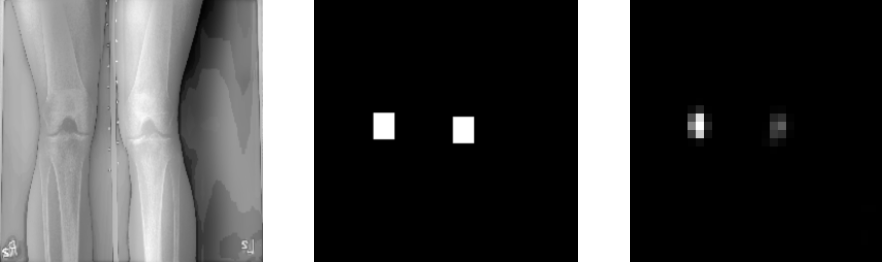}}
  \hfill
  \subfloat{\includegraphics[width=0.8\textwidth]{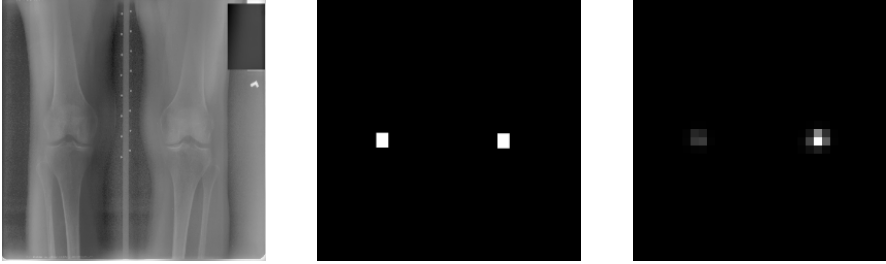}}
  \caption{Error analysis: X-ray images, ground truth, FCN output - weak detections}
  \label{fig:fcnErr12}
\end{figure}

\item[2.] The FCN output picks up noise along with the knee joints. Figure \ref{fig:fcnErr34} shows two instances of input X-ray images, masks defining the knee joint centres as ground truth, and output of the best performing FCN with noise. The reason for this error appears to be due to the variations in the imaging protocol and resolution of the X-ray images, and presence of artefacts in the input X-ray images. Intuitively, the FCN uses horizontal edge detection along with other features to detect the knee joints. The artefacts with predominant horizontal edges are picked up by the FCN along with the centre of knee joints. When simple contour detection is applied on the FCN output, instead of the knee joints the artefacts are also detected.
\end{itemize}

\begin{figure}[t]
  \centering
  \subfloat{\includegraphics[width=0.8\textwidth]{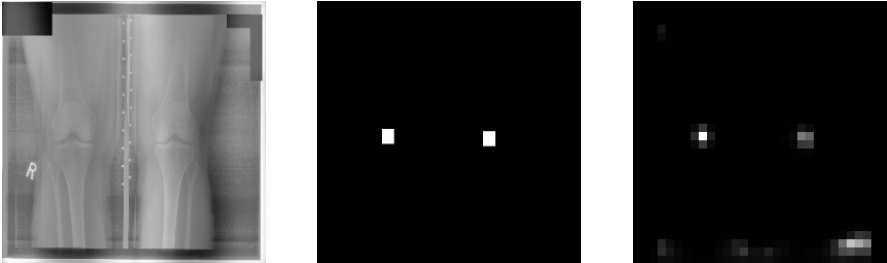}}
  \hfill
  \subfloat{\includegraphics[width=0.8\textwidth]{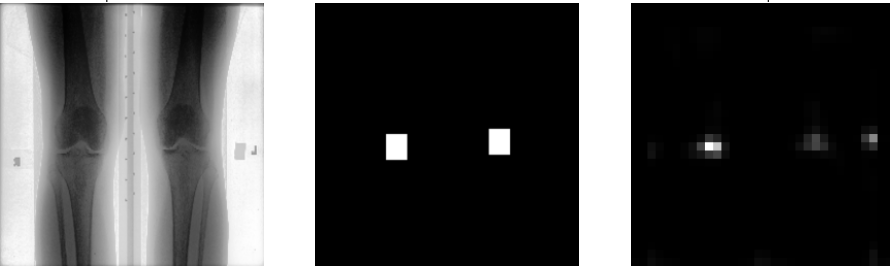}}
  \caption{Error analysis: X-ray images, ground truth, FCN output - detections with noise.}
  \label{fig:fcnErr34}
\end{figure}

\subsubsection*{Automatically Extracting the Knee Joints.}
\begin{figure}[t]
  \begin{minipage}[c]{.4\textwidth}
  \centering
  \subfloat{\includegraphics[width=0.7\textwidth, height=0.9\textwidth]{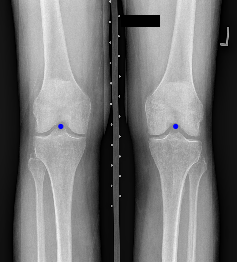}}
  \end{minipage}
  %\hfill
  \begin{minipage}[c]{.6\textwidth}
  \centering
  \subfloat{\includegraphics[width=0.7\textwidth, height=0.25\textwidth]{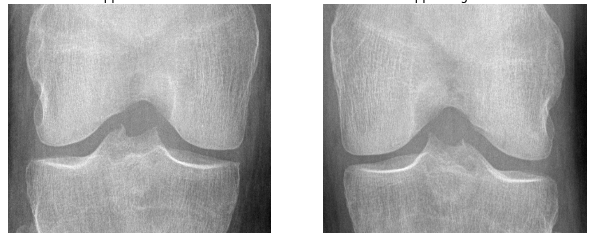}}
  \end{minipage}
  \caption{A knee X-ray image with the detected centres and the extracted left and right knees.}
  \label{fig:extKJ}
\end{figure}

After training FCNs to automatically detect the centre of the knee joints, the next step is to extract the ROI i.e. the knee joints with reference to the detected centres. The initial goal is to train an end-to-end network for localising the knee joints i.e. to directly predict the bounding box co-ordinates of the knee joints from the input X-ray images. A bounding box regression is investigated \cite{erhan2014scalable} that is a network trained on top of the FCN (Table \ref{Tab:FCN5}) output, to achieve this. First, CNNs are trained with the masks (20$\times$20) of knee joint centres as the input (256$\times$256) and the bounding box coordinates of the left knee joint ($x_1,y_1$) and right knee joint ($x_2,y_2$) as the ground truth (labels). Next, CNNs are trained with the X-ray images as input and the targets (labels) are the bounding box coordinates instead of the binary masks. However in both the experiments, the networks trained to predict the bounding boxes give low accuracy. On considering the overall knee joint centres, there is no large variations in the centre coordinates. The reason for the low accuracy is that the networks are not learning discernible features to predict the bounding box coordinates. This affects the overall performance of the localisation. Therefore, a simple approach based on contour detection is used to calculate the centres and extract the knee joints. Figure \ref{fig:extKJ} shows an X-ray image with the centres, the left and the right knee joints extracted from the X-ray image using the centroids. The steps involved in this method are as follows.

\begin{itemize}
\item First, the contour detection \cite{kilian2001simple} is used on the FCN output to calculate the spatial coordinates of the knee joint centres. In the contour detection method, first the input images (FCN output) are converted to binary by applying Otsu's threshold. Next, the contours from the binary image are automatically detected and recorded. Finally, the centroids are calculated from the detected knee joint regions. 
\item The knee OA radiographs are resized to 2560$\times$2560, that is 10 times the size of the FCN output 256$\times$256.
\item The detected knee joint centres are up-scaled to a factor of 10. 
\item Fixed size regions (640$\times$560) are extracted around the up-scaled centres as the knee joint regions.  After testing and visualising different sizes for the knee joint crop, image patch with the size (640$\times$560) is found to be mostly suitable and containing the required ROI for further quantification. Figure \ref{fig:extKJ} shows an instance of the extracted left and right knee joints. 
\end{itemize}

\subsubsection*{Localisation Results.}
The results of the FCN are compared to the previous methods: template matching and SVM-based method to automatically detect the centre of the knee joints. All these methods are evaluated based on the Jaccard index (JI). Table \ref{Tab:Res_KJC} shows the detection accuracy of the knee joint centres using FCN, SVM-based method, and template matching. The results show that the proposed method using FCN clearly outperforms the previous methods. This also demonstrates that feature learning using an FCN is a better approach for detecting the knee joints than using hand-crafted features such as Sobel gradients and the template matching method that is sensitive to intensity level variations. However, the extracted knee joints from this method have some limitations. %We discuss these limitations in the next section.

\begin{table}[t]
\caption{Comparison of methods used for localising the centre of the knee joints}
\label{Tab:Res_KJC}
\centering
\begin{tabular}{ l c c c c c }
\toprule 
Method & JI $>$ 0 & JI $\geq$ 0.5 & JI $\geq$ 0.75 & Mean & Std. Dev.\tabularnewline
\midrule 
\midrule 
Template Matching & 54.4\% & 8.3\% & 3.1\% & 0.1 & 0.2\tabularnewline
SVM-based Method & 81.8\% & 38.6\% & 10.2\% & 0.36 & 0.31\tabularnewline
Fully ConvNet & \textbf{98.9\%} & \textbf{97.1\%} & \textbf{43.3\%} & \textbf{0.76} & \textbf{0.12}\tabularnewline
\bottomrule 
\end{tabular}
\end{table}

\subsubsection*{Limitations of this Method.}
In all three approaches; FCN-based, SVM-based and template matching, the centre of the knee joints are detected and these are used as reference for automatically localising the knee joints. There are some limitations in extracting a fixed size region as the ROI with reference to the detected centres due to the variations in the resolution of the X-ray images and the variations in the size of the knee joints. 
\begin{figure}[ht]
  \centering
  \subfloat{\includegraphics[width=0.25\textwidth]{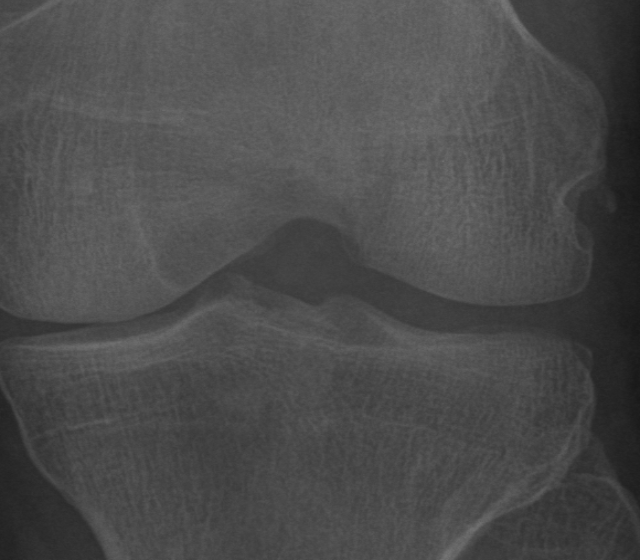}}
  \hspace{1em}
  \subfloat{\includegraphics[width=0.25\textwidth]{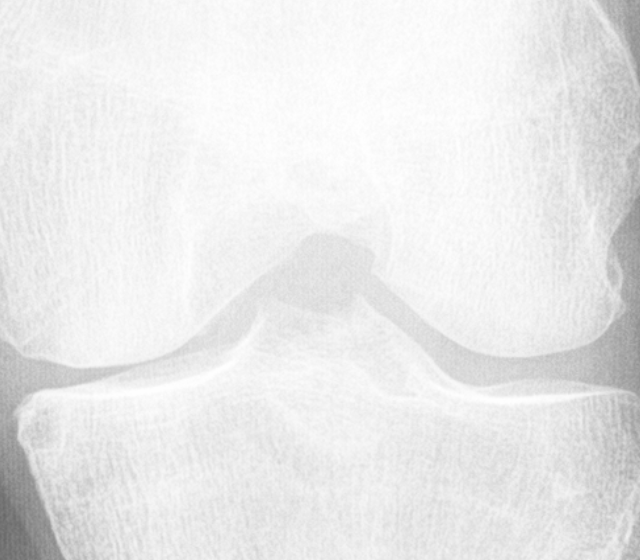}}
  \caption{Anomalies in the automatic extraction of the ROI.}
  \label{fig:KJC_am1}
\end{figure}

\begin{figure}[ht]
  \centering
  \subfloat{\includegraphics[width=0.25\textwidth]{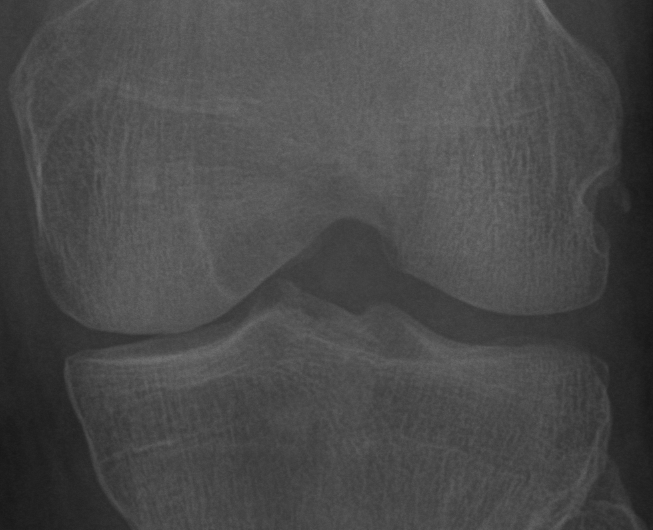}}
  \hspace{1em}
  \subfloat{\includegraphics[width=0.25\textwidth]{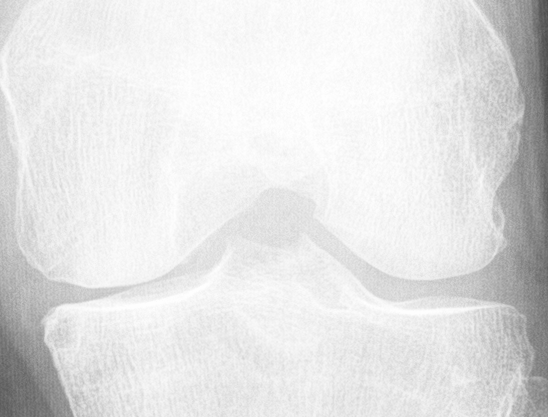}}
  \caption{The actual ROI for the knee joints in Figure \ref{fig:KJC_am1}.}
  \label{fig:KJC_man1}
\end{figure}

\begin{figure}[ht]
  \centering
  \subfloat{\includegraphics[width=0.25\textwidth]{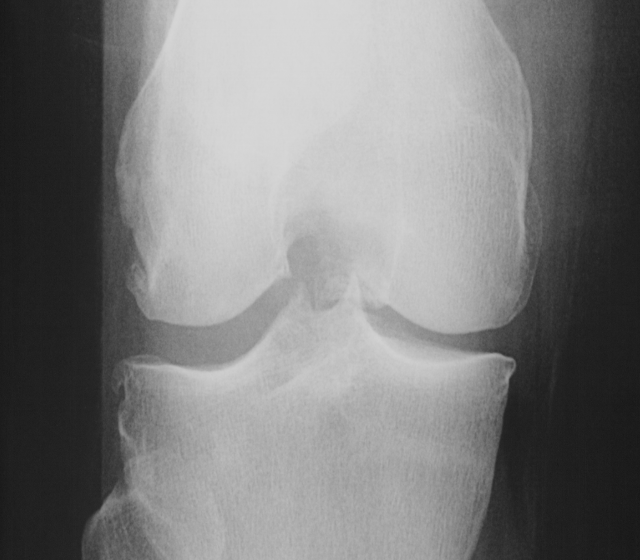}}
  \hspace{1em}
  \subfloat{\includegraphics[width=0.25\textwidth]{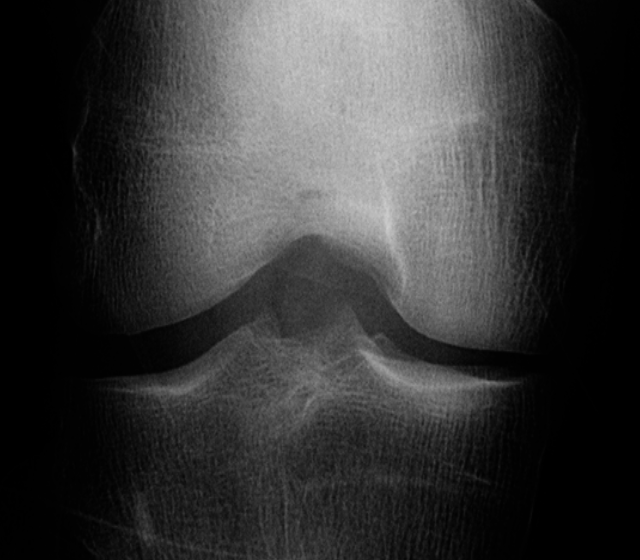}}
  \caption{Variations in the aspect ratio of the extracted knee joints.}
  \label{fig:KJC_am2}
\end{figure}

\begin{figure}[ht]
  \centering
  \subfloat{\includegraphics[width=0.25\textwidth]{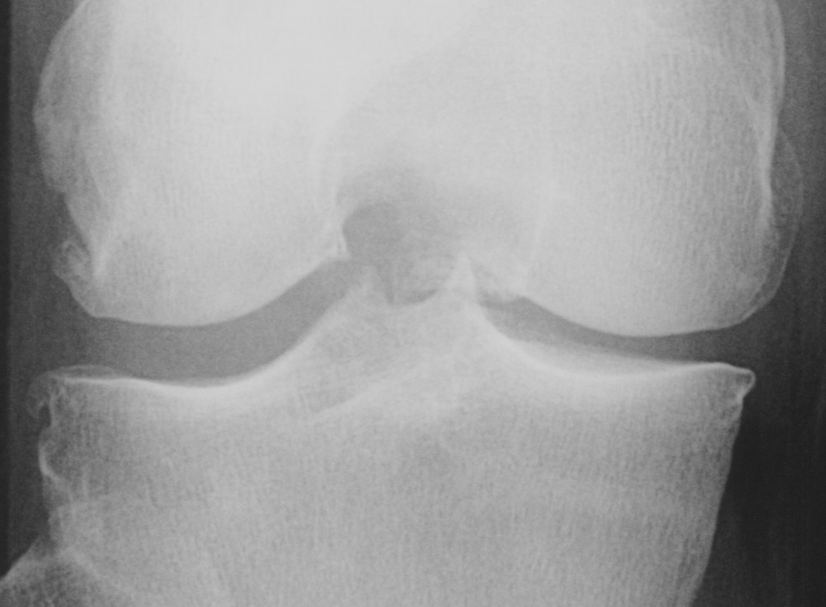}}
  \hspace{1em}
  \subfloat{\includegraphics[width=0.25\textwidth]{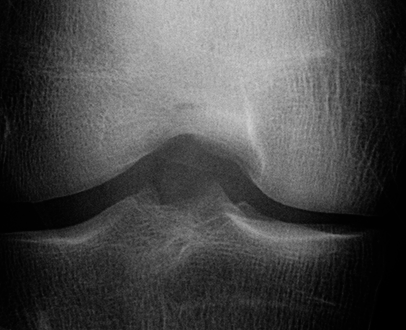}}
  \caption{The actual ROI for the extracted knee joints in Figure \ref{fig:KJC_am2}.}
  \label{fig:KJC_man2}
\end{figure}

All the images are resized to a fixed size 2,560$\times$2,560 and extract a fixed size region 640$\times$560 around the detected centres as the ROI. Due to this scaling issue, portions of the knee joints are omitted in the automatic extraction of the ROI. Figure \ref{fig:KJC_am1} shows such instances. Figure {\ref{fig:KJC_man1}} shows the corresponding actual ROIs. Due to the varying sizes of the knee joints and a fixed size region being extracted as the ROI, there are differences in the aspect ratio of the extracted and the actual ROI. Figure \ref{fig:KJC_am2} shows instances where the knee joints are small in comparison to the fixed size region extracted as the ROI. Figure \ref{fig:KJC_man2} shows the actual ROIs.

The classification of the automatically extracted knee joints is compared to the manually extracted knee joints. There is a decrease in the accuracy by a margin of 3--4\% when using the automatically extracted knee joints with reference to the detected centres. The discrepancies in the localisation of knee joints affects the overall classification of the knee OA images. To overcome these limitations, as the next approach FCNs are trained to detect the ROI itself, instead of detecting the knee joint centres. 

\subsubsection{Localising the Region of Interest}
\begin{figure}[t]
\centering
\includegraphics[width=\textwidth, height=0.5\textwidth]{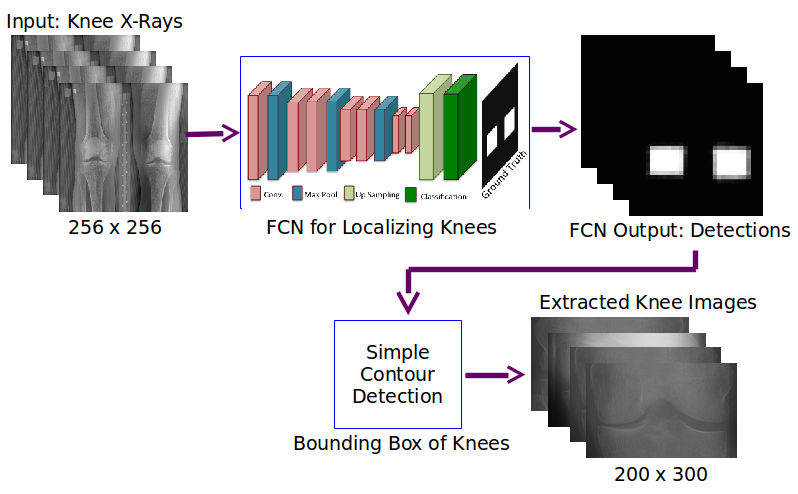}
\caption{Automatic localisation of the Region of Interest.} 
\label{fig:AD_ROI}
\end{figure}

˜\\
The previous methods to localise the knee joints in the X-ray images with reference to the automatically detected centres have certain limitations. To overcome these limitations and to improve the localisation, FCNs are trained to detect the ROI directly \cite{antony2017automatic}. Figure \ref{fig:AD_ROI} shows the steps involved in this method.

\subsubsection*{Dataset and Ground Truth.}
\begin{figure}[t]
  \centering
  \subfloat[]{\includegraphics[width=0.3\textwidth, height=0.27\textwidth]{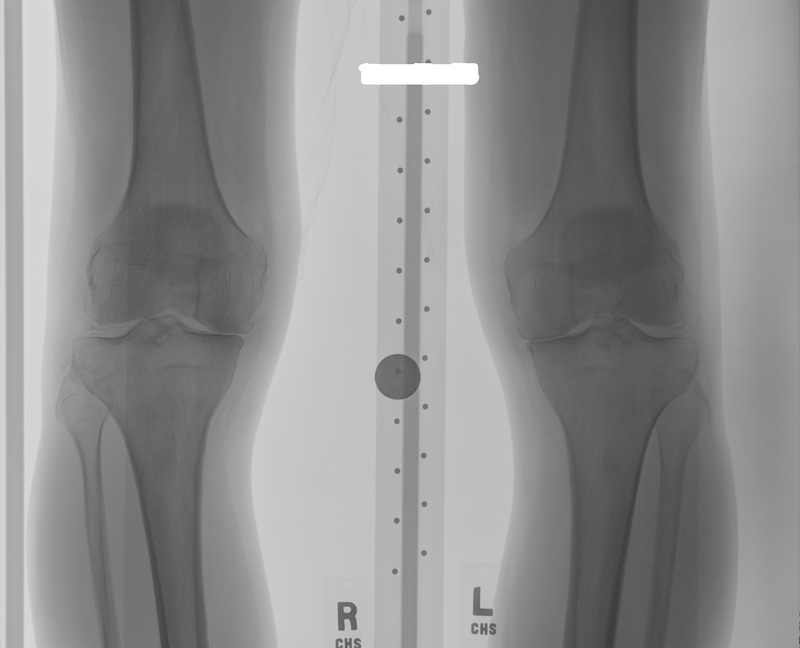}}
  \hspace{1em}
  \subfloat[]{\includegraphics[width=0.3\textwidth, height=0.27\textwidth]{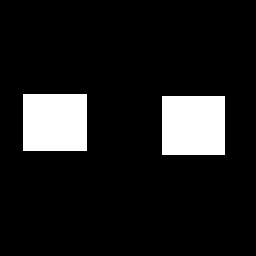}}
  \caption{(a) An input X-ray image and (b) The binary mask annotations for the region of interest.}
  \label{fig:gt_roi}
\end{figure}

For the experiments in this approach, a new dataset from the MOST is used along with the data from the previous experiments, the baseline cohort of the OAI dataset. In total 4,446 X-ray images are selected from the OAI dataset and 2,920 X-ray images from the MOST dataset based on the availability of KL grades for both knee joints. The full ROI is manually annotated in all these X-ray images, after downscaling to 10\% of the actual size. The down-sampling of the images is necessary to reduce the computational costs. Binary masks are generated based on the manual annotations. Figure \ref{fig:gt_roi} shows an instance of an input X-ray image and the binary mask annotations corresponding to the ROI. The image patches from the masked region (the knee joints) are taken as positive training samples and the patches from rest of the image are taken as the negative training samples to train a FCN.  The datasets are split into a training/validation set (70\%) and test set (30\%). The training and test samples from the OAI dataset are 3,146 images and 1,300 images, and from the MOST dataset are 2,020 images and 900 images. 

\subsubsection*{Training the FCN.}
\begin{figure}[t]
\centering
\includegraphics[scale=0.6]{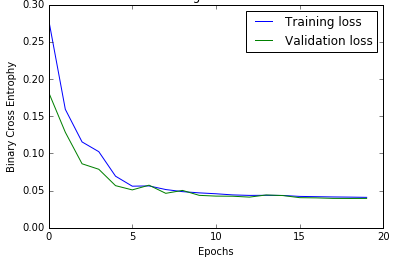}
\caption{Training and validation losses of the FCN.} 
\label{fig:LC_fcnROI}
\end{figure}

\begin{figure}[t]
\centering
\includegraphics[width=0.8\textwidth] {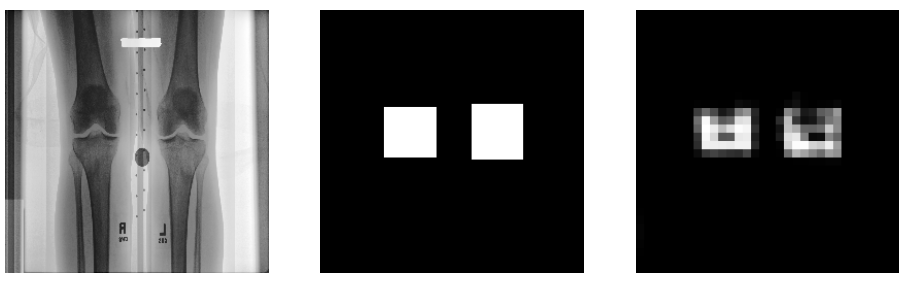}
\caption{An input X-ray image, ground truth and output prediction of the FCN.} 
\label{fig:fcn_outROI}
\end{figure}

First, a FCN is trained using the same architecture (Table \ref{Tab:FCN5}) from the previous approach to detect the ROI. Initially, the network is trained with training samples from OAI dataset and test it with OAI and MOST datasets separately. Next, the training samples are increased by including the MOST training set where the test set is a combination of both OAI and MOST test sets. This network is trained to minimise the total binary cross entropy between the predicted pixels and the ground truth using the adaptive moment estimation (Adam) optimiser with default parameters: initial learning rate $(\alpha) = 0.001$, $\beta_{1} = 0.9$, $\beta_{2} = 0.999$, $\epsilon = 1\mathrm{e}^{-8}$. Adam optimiser gives faster convergence than standard SGD. Figure \ref{fig:LC_fcnROI} shows the learning curves converging to small loss when training this network. Figure \ref{fig:fcn_outROI} shows the output of this network for a test image.

A few other network configurations are tested by varying the number of convolutional-pooling stages, convolutional layers in each stage and the number of convolutional kernels in a convolutional layer. There was no further improvement in the detection accuracy on the validation set. Therefore, this configuration was settled as the final network for localising the knee joints. 

\subsubsection*{Quantitative Evaluation.}
The Jaccard index, i.e. the intersection over Union (IoU) of the automatically detected and the annotated knee joint is used to quantitatively evaluate the automatic detections. For this evaluation, all the knee joints in both the OAI and MOST datasets are manually annotated using a fast annotation tool. Table \ref{Tab:Res_ROI} shows the number (percentage) of knee joint correctly detected based on the Jaccard index (JI) values greater than 0.25, 0.5 and 0.75 along with the mean and the standard deviation of JI. Table \ref{Tab:Res_ROI} also shows detection rates on the OAI and MOST test sets separately. %The test sets taken from the OAI contains 1,300 and the MOST contains 900 knee X-ray images.

\begin{table}[t]
\caption{Comparison of automatic detection based on the Jaccard Index (JI).}
\label{Tab:Res_ROI}
\centering
\begin{tabular}{l c c c c c}
\toprule
Test Data & JI $>$ 0 & JI $\geq$ 0.5 & JI $\geq$ 0.75 & Mean & Std. Dev.\tabularnewline
\midrule
\midrule
OAI & 100\% & 100\%& 88\% & 0.82 & 0.06\tabularnewline
MOST & 99.7\% & 98.8\% & 80.6\% & 0.80 & 0.09\tabularnewline
Combined OAI-MOST  & \textbf{100\%} & \textbf{100\%} & \textbf{92.2\%} & \textbf{0.83} & \textbf{0.06}\tabularnewline
\bottomrule
\end{tabular}
\end{table}

Considering the anatomical variations of the knee joints and the imaging protocol variations, the automatic detection with a FCN is highly accurate with 100\% detection accuracy for JI$\geq$0.5 and 92.2\% (4,056 out of 4,400) of the knee joints for J$\geq$0.75 being correctly detected. %Further evidence is provided to show that the FCN based detection is highly accurate by showing that the quantification results obtained with the automatically extracted knee joints giving results on par with manually segmented knee joints in the next chapter, Section 5.3.8.

\subsubsection*{Qualitative Evaluation.}
\begin{figure}[t]
\centering
\includegraphics[width=0.8\textwidth] {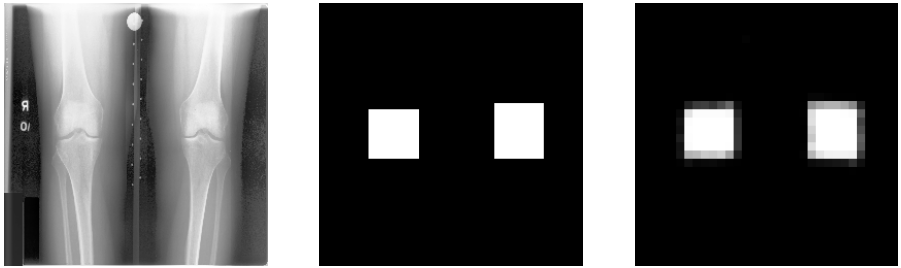}
\caption{Qualitative Evaluation: An input X-ray image, ground truth, and FCN detections: left knee with JI=0.98, right knee with JI=0.888.} 
\label{fig:QE1}
\end{figure}

\begin{figure}[t]
\centering
\includegraphics[width=0.8\textwidth] {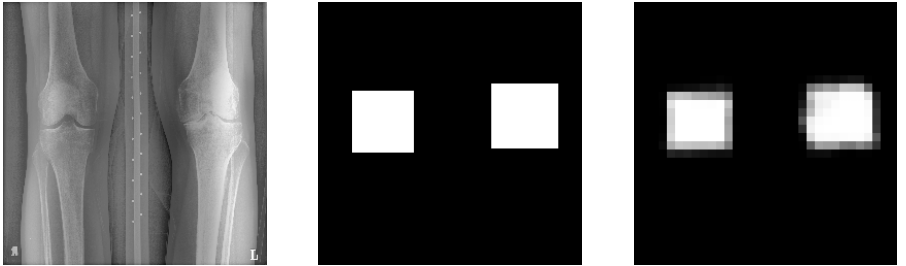}
\caption{Qualitative Evaluation: An input X-ray image, ground truth, and FCN detections: left knee with JI=0.879, right knee with JI=0.969.} 
\label{fig:QE2}
\end{figure}

\begin{figure}[t]
\centering
\includegraphics[width=0.8\textwidth] {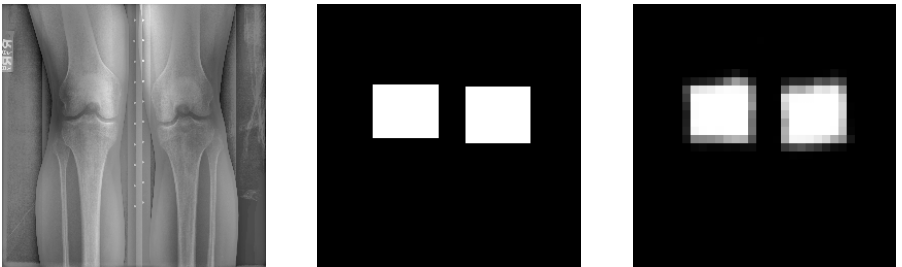}
\caption{Qualitative Evaluation: An input X-ray image, ground truth, and FCN detections: left knee with JI=0.768, right knee with JI=0.984.} 
\label{fig:QE3}
\end{figure}

Figures \ref{fig:QE1}, \ref{fig:QE2}, and \ref{fig:QE3} show a few instances of successful knee joint detections with the JI values for the left and right knee detections. Detecting the ROI directly gives high accuracy (100\%) in comparison to the previous method (Section 4.2) to detect the knee joint centres and extracting a fixed size region as the ROI. The FCN in this method learns features from a relatively larger region (the actual ROI) in comparison to the previous method where the FCN is confined to learn features from a small region (20$\times$20), the centre of the knee joints, and therefore, the detections are more accurate. 

\subsubsection*{Error Analysis.}
\begin{figure}[t]
\centering
\includegraphics[width=0.8\textwidth] {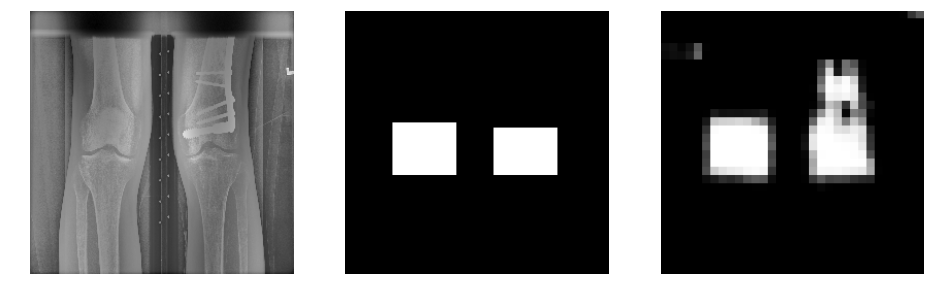}
\caption{Error Analysis: An input X-ray image, ground truth, and FCN detections: left knee with JI=0.83, right knee with JI=0.398. The implants in the right knee is the reason for this localisation error.}
\label{fig:EA1}
\end{figure}

\begin{figure}[t]
\centering
\includegraphics[width=0.8\textwidth] {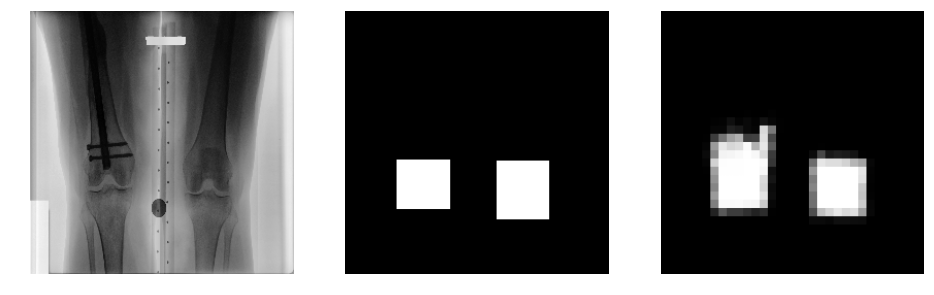}
\caption{Error Analysis: An input X-ray image, ground truth, and FCN detections: left knee with JI=0.473, right knee with JI=0.837. The implants in the left knee is the reason for this localisation error.} 
\label{fig:EA2}
\end{figure}

\begin{figure}[t]
\centering
\includegraphics[width=0.8\textwidth] {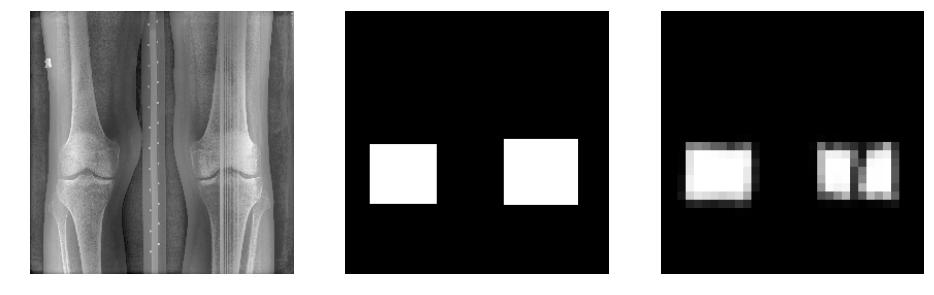}
\caption{Error Analysis: An input X-ray image, ground truth, and FCN detections: left knee with JI=0.887, right knee with JI=0.356. The noise in the right knee causes this localisation error.} 
\label{fig:EA3}
\end{figure}

\begin{figure}[t]
\centering
\includegraphics[width=0.8\textwidth] {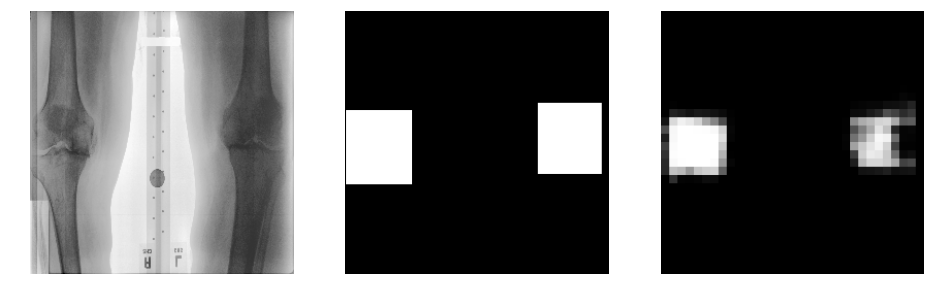}
\caption{Error Analysis: An input X-ray image, ground truth, and FCN detections: left knee with JI=0.681, right knee with JI=0.488. The localisation error in this image is due to the variation in the imaging protocol.}
\label{fig:EA4}
\end{figure}

\begin{figure}[t]
\centering
\includegraphics[width=0.8\textwidth] {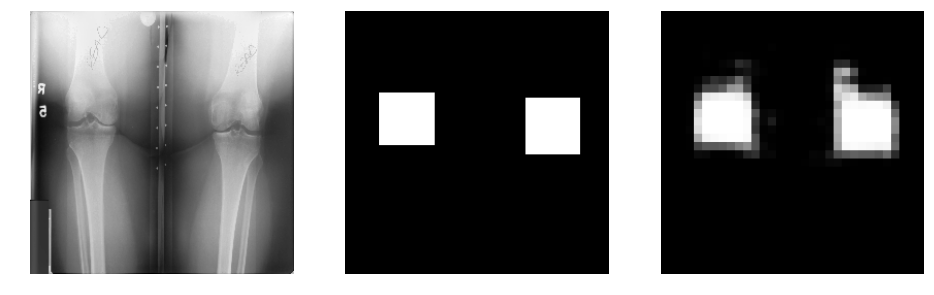}
\caption{Error Analysis: An input X-ray image, ground truth, and FCN detections: left knee with JI=0.768, right knee with JI=0.507. The variations in the local contrast and luminance affects the localisations.} 
\label{fig:EA5}
\end{figure}

This method is highly accurate with 100\% detection accuracy for a JI $\geq$ 0.5. Nevertheless, there are a few anomalies in the FCN detections due to variations in the imaging protocols, presence of artefacts and noise in the input images. Figures \ref{fig:EA1} and \ref{fig:EA2} show two instances where one knee has undergone joint-arthoplasty and the knee implants are visible in the X-ray images, and due to this the FCN detections are distorted. Figures \ref{fig:EA3}, \ref{fig:EA4} and \ref{fig:EA5} show a few instances of X-ray images with noise and presence of artefacts due to imaging protocols. This adversely affects the FCN detections. 

\subsubsection*{Extracting the Knee Joints.}
The bounding boxes of the knee joints are calculated using simple contour detection from the output predictions of the FCN. After converting the FCN output to binary image using Otsu's threshold, the contours are detected using simple image analysis by calculating the zero order moments \cite{kilian2001simple}, which gives the perimeter of the detected object. 
The contours are recorded as bounding boxes. The knee joints are extracted from knee OA radiographs using the bounding boxes. The bounding boxes are up-scaled from the output of the FCN that is of size [$256\times256$] to the original size of each knee OA radiograph, before extracting the knee joints so that the aspect ratio of the knee joints is preserved. 

\subsection{Summary and Discussion}˜\\
Automatically localising the knee joints in X-ray images is an important and an essential step before quantifying knee OA severity. Previously, template matching was implemented as a baseline method to localise the knee joints, proposed by Shamir et al. \cite{shamir2009knee,shamir2009early}, and it was shown that the detection accuracy is low ($\sim30\%$) in this method for large datasets like OAI. To improve the localisation, a SVM-based method with Sobel horizontal image gradients as features was proposed in this Section. This method showed a large improvement in detection accuracy (82\%) but still falls short of perfect localisation. The anomalies in localised knee joints can affect the step involving classification of the localised knee joints to quantify knee OA severity. 

Instead of using hand-crafted features, a deep learning-based solution was proposed in this Section to further improve localisation. FCNs were trained to automatically detect and extract the knee joints. All three methods: template matching, SVM-based and FCN-based were evaluated using a common metric: the Jaccard Index. This method achieved almost perfect detection with 100\% accuracy for a Jaccard Index 0.5 and an accuracy of 92\% for a Jaccard index greater than equal to 0.75. %In addition to these results, we will show that the classification  using localized knee joints from this method is on par with manually segmented knee joints.
The author believes this performance is sufficient to localise and extract the knee images for classification. As such further improvements are left as future work. The localisation performance may be improved by including additional pre-processing steps to remove the artefacts and noise in the images, and to normalise the local contrast variations in the images. Using additional data for learning and data augmentation may improve the localisation performance.

%--------------------------------------------------------------------------%

\section{Automatic Assessment of Knee OA Severity}

Previous work on automated assessment of knee OA severity approached it as an image classification problem \cite{thomson2015automated,shamir2009early,subramoniam2013local,subramoniam2015non,deokar2015effective}. Previous methods have tested many hand-crafted features based on pixel statistics, textures, edge and object statistics, and transforms \cite{shamir2009early,shamir2008wndchrm,oka2008fully,park2013practical,subramoniam2013local,subramoniam2015non}. Many classifiers such as the SVM \cite{subramoniam2015non}, the k-nearest neighbour classifier \cite{subramoniam2013local}, the weighted neighbour nearest classifier\cite{shamir2009early,shamir2008wndchrm}, the random forest classifiers \cite{thomson2015automated}, and even artificial neural networks (ANN) \cite{deokar2015effective,yoo2016simple} have been tested for knee image classification. As a baseline (in Section 4.1), the state-of-the-art features successful in computer vision tasks, such as histogram of oriented gradients \cite{dalal2005histograms}, local binary patterns \cite{ojala2002multiresolution}, and Sobel Gradients \cite{sobel1990isotropic} are tested. These features are not included in the previous studies to assess knee OA severity. All the previous approaches based on hand-crafted features give low multi-class classification accuracy when classifying knee images, and in particular classifying fine-grained successive knee OA grades remains a challenge. As a baseline, the state-of-the-art CNNs features (in Section 5.2.2) are also tested for knee images classification on a small baseline data set from OAI and this approach gave promising results. Motivated by this, the use of CNNs are investigated for quantifying knee OA severity.

%The remainder of this Section is structured as follows: Section 5.2 presents the classification and regression of knee OA images using CNNs that are fine-tuned through transfer learning. Section 5.3 introduces the training of CNNs from scratch for classifying knee OA images and analyses the classification results. Section 5.4 elaborates on the joint training of CNNs for simultaneous classification and regression of knee OA images, and shows the results of joint training. Section 5.5 describes the development and training of a CNN for ordinal regression using a custom loss function. Section 5.6 compares and analyses the results from the four approaches to quantify knee OA severity. Section 5.7 summarises this Section and presents the conclusions. 

\subsection{Baseline for Classifying Knee OA Radiographs}

\subsubsection{WNDCHRM Classification}˜\\
WNDCHRM is an open source utility for biological image analysis and medical image classification \cite{shamir2009early,shamir2008wndchrm,shamir2013wnd}. In WNDCHRM, a generic set of image features based on pixel statistics (multi--scale histograms, first four moments), textures (Haralick and Tamura features), factors from polynomial decomposition (Zernike polynomials), and transforms (Radon, Chebyshev statistics, Chebyshev-Fourier statistics) are extracted. For feature selection, every feature is assigned a \textit{Fisher score}\footnote{Fisher score is one of the widely used methods for determining the most relevant features for classification} and 85\% of the features with lowest \textit{Fisher scores} are rejected and the remaining 15\% of the features are used for classification \cite{shamir2009early}.

\subsubsection*{Experiments.}
The dataset used for the initial experiments to classify knee OA images using WNDCHRM are taken from the baseline data sample of 200 progression and incidence cohort. After histogram equalisation and mean normalisation of the X-ray images, the knee joints are extracted manually from the radiographs. The extracted knee joints are split into training (70\%) and test (30\%) sets. The WNDCHRM command line program is used to classify the extracted knee joint images. WNDCHRM uses a variant of k-nearest neighbour classifier.

\subsubsection*{Results and Discussion.}
The baseline dataset is not balanced and there are only 44 samples available in KL grade 4. %Figure \ref{fig:Grades} in Section 3.2 (Page 41) shows the distribution of the entire data set. 
Given the limited number of images in this class, only a small number of images are used for training and testing (35 images for training and 9 images for testing) for multi-class classification. For other classifications 100 images are used for training and 30 images for testing. 

\begin{table}[t]
\caption{Results of WNDCHRM Classification.}
\label{Tab:Wndchrm_Results}
\centering
\begin{tabular}{ l c c }
\toprule
Classification & Grades & Accuracy\tabularnewline
\midrule 
\midrule
\multirow{6}{*}{Binary} & G0 vs G1 & 66.7 \%\tabularnewline
 & G1 vs G2 & 48.3 \%\tabularnewline
 & G2 vs G3 & 60 \%\tabularnewline
 & G3 vs G4 & 55 \%\tabularnewline
 \cmidrule{2-3}
 & G0 vs G2 & 48.3 \%\tabularnewline
 & G0 vs G3 & 70 \%\tabularnewline
\midrule
\multirow{2}{*}{Multi-class} & G0 to G4  & 28.3 \%\tabularnewline
 & G0 to G3 & 35.8 \%\tabularnewline
\bottomrule
\end{tabular}
\end{table}

It is evident from the results (Table \ref{Tab:Wndchrm_Results}) that the multi-class classification accuracy and successive grades classification accuracies are very low. The reason for low classification accuracy is that the features used for classification are not capable of capturing the minute structural and morphological variations in the knee joints between the successive grades. Next, the state-of-the-art hand-crafted features are investigated in an attempt to improve the classification accuracy.

\subsubsection{Classification using Hand-crafted Features}˜\\

Histogram of oriented gradients (HOG), local binary patterns, and Sobel Gradients are tested for classifying knee OA images \cite{shamir2009early,shamir2009knee,orlov2008wnd}. These features are not used in the previous studies. HOG describes the local object shape and appearance within an image by the distribution of intensity gradients or edge directions and the HOG descriptor was successful in human detection \cite{dalal2005histograms}. LBP is powerful for image texture classification.  LBP uses local spatial patterns and grey scale contrast as measures for texture classification \cite{ojala2002multiresolution}. %The Sobel operator calculates the spatial gradients in an image and highlights the regions of high spatial frequency that corresponds to edges \cite{sobel1990isotropic}.

\subsubsection*{Experiments with HOG, LBP and Sobel descriptors.}
Once again the images from the baseline data sample of 200 progression and incidence cohort is used. The HOG, LBP and Sobel descriptors are extracted from the knee joint images and a SVM is used for classification. Table  \ref{Tab:Res_ClsfHC} shows the classification results of successive grades of knee OA images using a SVM and the feature space included the HOG, LBP and Sobel gradients. 

\begin{table}[t]
\caption{Classification results of the proposed methods using hand-crafted features.}
\label{Tab:Res_ClsfHC}
\centering
\begin{tabular}{ l c c c c c}
\toprule
\multirow{2}{*}{Grades} & \multirow{2}{*}{WNDCHRM} & \multicolumn{4}{c}{SVM classification with hand-crafted features}\tabularnewline
\cmidrule{3-6}
 &  & HOG & LBP & Sobel & Combining all\\
\midrule 
\midrule
 G0 vs G1 & \textbf{66.7 \%} & 53.3 \% & 58.3 \% & 58.3 \% & 55 \%\tabularnewline
G1 vs G2 & 48.3 \% & 48.3 \% & 53.3 \% & \textbf{58.3 \%} & 51.6 \%\tabularnewline
G2 vs G3 & 60 \% & 60 \% & 60 \% & 56.7 \% & \textbf{63.3 \%}\tabularnewline
G3 vs G4 & 55 \% & \textbf{65 \%} & \textbf{65 \%} & 50 \% & \textbf{65 \%}\tabularnewline
\bottomrule
\end{tabular}
\end{table}

There is no large improvement in the classification accuracies using HOG, LBP and Sobel gradients features with SVM classification from the previous results with the WNDCHRM classification. To improve the classification, the features space is expanded by including highly effective and top-ranked features from the WNDCHRM classification. 

\subsubsection*{Expanding the feature space.}
The features based on pixel statistics and textures such as Tamura, Haralick, Gabor and Zernike are used for classification. These features are used in the WNDCHRM classification. Tamura texture features represent contrast, coarseness and directionality of an image \cite{tamura1978textural}. Haralick features are the statistics computed on the co-occurrence matrix of an image \cite{haralick1973textural}. Gabor textures are based on Gabor wavelets and the image descriptors are computed using Gabor transform of an image \cite{grigorescu2002comparison}. Zernike features are obtained by the Zernike polynomial approximation of an image \cite{teague1980image}. The feature space for classification is formed by simple concatenation of all the extracted features into a super vector following the early fusion approach. 

\subsubsection*{Results and Discussion.}
First, a SVM is used with the extracted features for classifying knee OA images. Next, a k-nearest neighbour classifier and support vector regression (SVR) are tested for classification. In total, 100 knee joint images are taken for training and 30 for test set in each grade. Table \ref{Tab:Res_Clsf} shows the classification accuracy of the WNDCHRM classifier and the classification using kNN, SVM, and SVR.

\begin{table}[t]
\caption{Classification results of WNDCHRM and the proposed methods using hand-crafted features.}
\label{Tab:Res_Clsf}
\centering
\begin{tabular}{ l c c c c }
\toprule
\multirow{2}{*}{Grades} & \multirow{2}{*}{WNDCHRM} & \multicolumn{3}{c}{Proposed Methods}\tabularnewline
\cmidrule{3-5} 
 &  & kNN & SVM & SVR\tabularnewline
\midrule
\midrule
G0 vs G1 & \textbf{66.7\%} & 55\%  & 60\%  & 60\%\\
G1 vs G2 & 48.3\% & \textbf{61.7 \%} & 46.7\% & 48.3\% \\
G2 vs G3 & \textbf{60\%} & 51.7\% & 55\% & \textbf{60\%} \\
G3 vs G4 & \textbf{55\%} & 35\% & 50\% & 45\% \\
\midrule
G0 vs G2 & 48.3\% & 46.7\% & 55\% & \textbf{56.7\%}\\
G0 vs G3 & \textbf{70\%} & 48.3\% & 58.3\% & 60\% \\
\bottomrule
\end{tabular}
\end{table}

When comparing the classification results of the proposed methods (SVM, kNN, and SVR) to the WNDCHRM classification, for some cases the results are slightly better and promising. Nevertheless, there is a need for a more significant improvement in the classification results. In these experiments, a subset of features from WNDCHRM, such as Tamura \& Haralick texture features, Gabor wavelet features, and Zernike features, were extracted and used for classification. In addition to these features HOG, LBP, and Sobel Gradients were tested. It was found that by further expanding the feature space by including features from WNDCHRM based on transforms such as Radon, Chebyshev, FFT, and Wavelet, and compound image transforms such as Chebyshev-FFT, Chebyshev-Wavelet, and Wavelet-FFT classification can be improved. However, the author believes that learning feature representations can be more effective for fine-grained knee OA classification. In the following section, the state-of-the-art CNN features are investigated for classifying knee OA images. 

\subsection{Automatic Quantification using Convolutional Neural Networks}
First, the use of off-the-shelf CNNs are investigated for quantifying knee OA severity through classification and regression. Two approaches are followed for this: 1) using a pre-trained CNN for fixed feature extraction, and 2) fine-tuning pre-trained CNN following a transfer learning approach. WNDCHRM, an open source utility for medical image classification\cite{shamir2008wndchrm,shamir2009early,shamir2013wnd} is used for benchmarking the classification results obtained from the proposed methods. 

Next, three new methods are investigated to automatically quantify knee OA: 1) training a CNN from scratch for multi-class classification of knee OA images; 2) training a CNN to optimise a weighted ratio of two loss functions categorical cross-entropy for multi-class classification and mean-squared error for regression; and 3) training a CNN for ordinal regression of knee OA images. The results from these methods are compared to the previous methods. The classification results using both manual and automatic localisation of knee joints are also compared.

%The remainder of this Section is structured as follows: Section 5.2 presents the classification and regression of knee OA images using CNNs that are fine-tuned through transfer learning. Section 5.3 introduces the training of CNNs from scratch for classifying knee OA images and analyses the classification results. Section 5.4 elaborates on the joint training of CNNs for simultaneous classification and regression of knee OA images, and shows the results of joint training. Section 5.5 describes the development and training of a CNN for ordinal regression using a custom loss function. Section 5.6 compares and analyses the results from the four approaches to quantify knee OA severity. Section 5.7 summarises this Section and presents the conclusions. 

\subsubsection{Off-the-shelf CNNs}˜\\
The use of well-known off-the-shelf CNNs such as the VGG-16 network \cite{simonyan2014very}, and comparatively simpler networks like VGG-M-128 network \cite{chatfield2014return}, and BVLC reference CaffeNet \cite{jia2014caffe,karayev2013recognizing} (which is very similar to the widely-used \textit{AlexNet} model \cite{krizhevsky2012imagenet}) are investigated to classify knee OA images. These networks are pre-trained for general image classification using a very large dataset: the ImageNet LSVRC dataset \cite{russakovsky2015imagenet} which contains more than 1.2 million images in 1000 classes. Initially, features are extracted from the convolutional, pooling, and fully-connected layers of VGG-16, VGG-M-128, and BVLC CaffeNet, and used to train linear SVMs to classify knee OA images.

The pre-trained networks are fine-tuned for knee OA images classification motivated by the transfer learning approach \cite{yosinski2014}. Transfer learning is adopted as the OAI dataset is small, containing only a few thousand images. In  transfer learning, a base network is first trained on external data, and then the weights of the initial $n$ layers are transferred to a target network \cite{yosinski2014}. The new layers of the target network are randomly initialised following the Xavier weight initialisation procedure \cite{glorot2010understanding}. The random weights initialisations increase the likelihood of the training algorithms during the backpropagation to obtain a global solution through the gradient descent instead of settling to a nearest local solution. 

Intuitively, the lower layers of the networks contain more generic features such as edge or texture detectors useful for multiple tasks, whilst the upper layers progressively focus on more task specific cues \cite{karayev2013recognizing,yosinski2014}. This approach is used for both classification and regression, adding new fully-connected layers, and backpropagation is used to fine-tune the weights for the complete network on the target loss. 

\subsubsection{Classification using CNN Feature Extraction}˜\\
The VGG-16 network \cite{simonyan2014very} is trained with the OAI dataset. Features are extracted from different layers of the VGG net such as fully-connected (fc7), pooling (pool5), and convolutional (conv5\_2) layers to identify the most discriminating set of features. Linear SVMs (LIBLINEAR~\cite{fan2008liblinear}) are trained with the extracted CNN features for classifying knee OA images, where the ground truth are images labelled with KL grades. Next, the use of simple pre-trained CNNs such as VGG-M-128 \cite{chatfield2014return} and the BVLC CaffeNet model \cite{jia2014caffe} are investigated for classifying the knee OA images. These networks have fewer layers and parameters in comparison to the VGG-16 network. The features {are extracted} from the fully-connected, pooling, and convolutional layers, using the VGG-M-128 net and the BVLC reference CaffeNet.

\subsubsection*{Experiments and Results.}
The knee joint images are split into training ($\sim$70\%) and test ($\sim$30\%) set based on the distribution of each KL grade. Features are extracted from fully-connected, pooling, and convolution layers of VGG-16, VGG-M-128, and BVLC CaffeNet. Linear SVMs are trained individually for binary and multi-class classifications on the extracted features. WNDCHRM is used for benchmarking the classification results from the proposed methods in this chapter \cite{shamir2009early,shamir2008wndchrm,shamir2013wnd}. %WNDCHRM uses many hand-crafted features based on pixel statistics, textures, edge and object statistics, polynomial decomposition, transforms and compound transforms of images. WNCHRM uses weighted neighbour classifier \cite{orlov2008wnd}, a variation of k-nearest neighbour classifier. 
WNDCHRM is trained with the same training data so that the classification results from WNDCHRM and CNN features can be compared. The knee OA images are classified in three ways as follows. Classifying healthy knee images (grade 0) with the progressive stages (grade 1, 2, 3, and 4), classifying the images belonging to the successive stages (grade 0 vs 1, grade 1 vs 2, ...) and multi-class classification to classify all the stages of knee OA images. 

\begin{sidewaystable}[htbp]
\centering
\caption{Classification accuracy (\%) achieved by the WNDCHRM and pre-trained CNN features.}
\centering
\begin{tabular}{ c c c c c c c c c c c c}
 \toprule
 \multirow{2}{*}{Category} & \multirow{2}{*} {Classification} & \multirow{2}{*}{WNDCHRM} & \multicolumn{3}{c}{VGG-16 Net} & \multicolumn{3}{c}{VGG-M-128 Net}&  \multicolumn{3}{c}{BVLC ref CaffeNet}\\ 
\cline{4-12}
&  &  & fc7 & pool5 & conv5\_2 & fc6 & pool5 & conv4 & fc7 & pool5 & conv5\\
 \midrule
 \midrule
\multirow{4}{*}{Progressive} & Grade 0 vs Grade 1 & 51.5 &  56.3 &  61.3 & 63.5 & 56.5 & 63.2 & \textbf{64.7} & 62.0  & 64.3 & 63.3\\  
& Grade 0 vs Grade 2 & 62.6 &  68.6 &  74.3 & 76.7 & 67.8 & 75.5 & \textbf{77.6} & 69.6  & 73.6 & 73.9\\ 
& Grade 0 vs Grade 3 & 70.6 &  86.4 &  91.4 & 92.4 & 88.5 & 90.2 & \textbf{92.9} & 87.9  & 92.5 & 91.5\\ 
& Grade 0 vs Grade 4 & 82.8 &  98.1 &  98.6 & 99.3 & 98.8 & 99.3 & 99.2 & 98.5  & \textbf{99.4} & 99.1\\  
 \midrule
 
 \multirow{3}{*}{Successive} & Grade 1 vs Grade 2 & 48.8 &  60.0 &  64.7 & 67.3 & 57.9 & 63.5 & 65.3 & 61.2  & \textbf{65.8} & 62.8 \\
& Grade 2 vs Grade 3 & 54.5 &  69.8 &  76.4 & 77.0 & 73.0 & 77.3 & \textbf{79.0} & 70.3  & 78.1 & 77.1\\ 
& Grade 3 vs Grade 4 & 58.6 &  85.2 &  88.8 & 90.0 & 85.0 & 90.4 & 91.2 & 87.4 & \textbf{91.6} & 91.4\\ 
\midrule

\multirow{3}{*}{Multi-class} & Grade 0 to Grade 2 & 39.9 &  51.1 &  53.4 & 56.9 & 51.1 & 55.0 & \textbf{57.4} & 51.1  & 54.8 & 54.4\\  
& Grade 0 to Grade 3 & 32.0 &  44.6 &  48.7 & 53.9 & 45.4 & 50.2 & \textbf{53.3} & 46.9  & 51.6 & 50.2\\ 
& Grade 0 to Grade 4 & 28.9 &  42.6 &  47.6 & 53.1 & 43.8 & 49.5 & \textbf{53.4} & 44.1  & 50.8 & 50.0\\ 
\bottomrule
\end{tabular}
\label{Tab:Clsf_PT}
\end{sidewaystable}

Table \ref{Tab:Clsf_PT} shows the test set classification accuracies achieved by WNDCHRM and the CNN features. The CNN features consistently outperform WNDCHRM for classifying healthy knee samples against the progressive stages of knee OA. The features from conv4 layer with dimension 512$\times$13$\times$13 and pool5 layer 256$\times$13$\times$13 of VGG-M-128 net, and conv5 layer with dimension 512$\times$6$\times$6 and pool5 layer with dimension 256$\times$6$\times$6 of BVLC reference CaffeNet give higher classification accuracy in comparison to the fully-connected fc6 and fc7 layers of VGG nets and CaffeNet. Intuitively, the lower layers capture more discriminative low-level features such as edge or shape detectors, and the higher layers tend to contain high-level features specific to object classes as per the training data. Features are also extracted from lower layers such as pool4, conv4\_2, pool3, pool2 and train classifiers on top of these features. As the dimension of the bottom layers are high, the training time is increased, however, no improvement in classification accuracy is observed.  

In a fine-grained classification task such as knee OA image classification, the accuracy of classifying successive classes tends to be low, as the variations in the progressive stages of the disease are minimal, and only highly discriminant features can capture these variations. From the experimental results, as shown in Table \ref{Tab:Clsf_PT}, the features extracted from CNNs provide significantly higher classification accuracy in comparison to the WNDCHRM, and these features are effective and promising for classifying the consecutive stages of knee OA.

Multi-class classifications are performed using linear SVMs with the CNN features (Table \ref{Tab:Clsf_PT}, multi-class). Again, the CNN features outperform WNDCHRM. The classification accuracies obtained using convolutional (conv4, conv5) and pooling (pool5) layers are slightly higher in comparison to fully-connected layer features. There are minimal variations in classification accuracy obtained with the features extracted from VGG-M-128 net and BVLC reference CaffeNet in comparison to VGG-16.

\subsubsection{Transfer Learning}˜\\
As a next approach \cite{antony2016}, the BVLC CaffeNet~\cite{jia2014caffe} and VGG-M-128 \cite{chatfield2014return} networks are fine-tuned using transfer learning to classify knee images. These two smaller networks are chosen because they contain fewer layers and parameters ($\sim$62M), over the much deeper VGG-16, which has $\sim$138M parameters. The top fully-connected layer of both networks is replaced and  the model is retrained on the OAI dataset using backpropagation. The lower-level features in the bottom layers are also updated during fine-tuning. Standard softmax loss is used as the objective for classification, and accuracy layers are added to monitor the training progress. A Euclidean loss layer (mean squared error) is used for the regression experiments.

\subsubsection*{Experiments and Results.}

\begin{figure}[t]
  \centering
  \subfloat{\includegraphics[width=0.48\textwidth, height=0.4\textwidth]{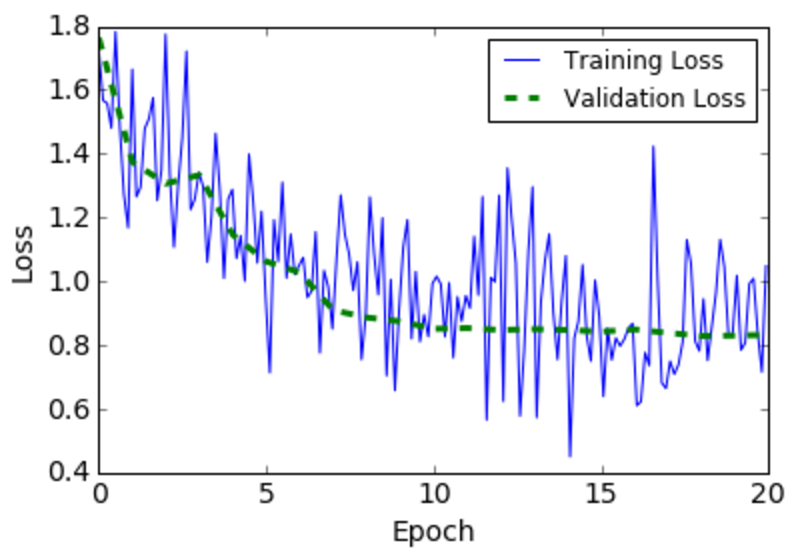}}
  \hspace{1em}
  \subfloat{\includegraphics[width=0.48\textwidth, height=0.4\textwidth]{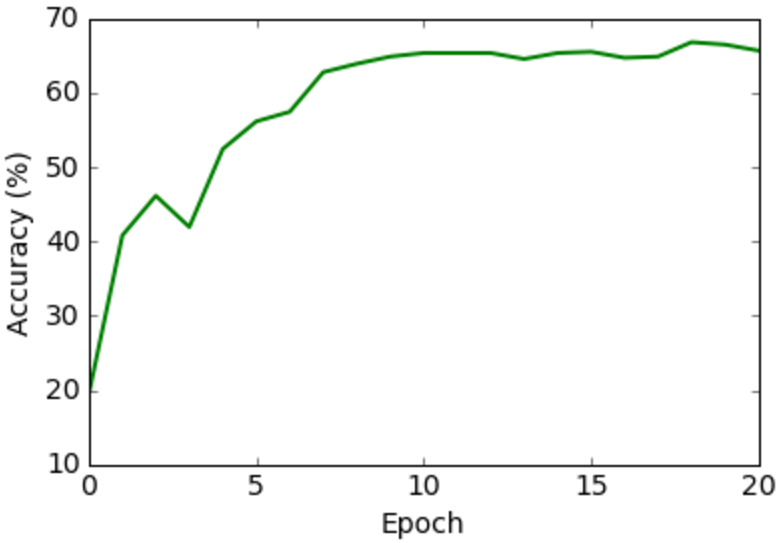}}
  \caption{Learning curves:training and validation losses (left), and validation accuracy (right) during fine-tuning.}
  \label{fig:LossAcc}
\end{figure}

Table \ref{Tab:Clsf_FT} shows the multi-class classification results for the fine-tuned BVLC CaffeNet. The VGG-16 network is omitted in these experiment since the variation in accuracy among the pre-trained CNNs is small, and fine-tuning VGG-16 is more computationally expensive.

The dataset is split into training (60\%), validation (10\%) and test (30\%) sets for fine-tuning. The right-left flipped knee joint images are included in the training set to increase the number of training samples. The networks are fine-tuned for 20 epochs using a learning rate of 0.001 for the transferred layers, and 0.01 for the newly introduced layers. The performance of fine-tuned BVLC CaffeNet is slightly better than VGG-M-128. Hence, the results of fine-tuning BVLC CaffeNet is only shown here. Figure \ref{fig:LossAcc} shows the learning curves for training and validation loss, and validation accuracy. The decrease in loss and increase in accuracy shows that the fine-tuning is effective and makes the CNN features more discriminative, which improves classification accuracy (Table~\ref{Tab:Clsf_PT}). The features extracted from the fully connected (fc7) layer provide slightly better classification in comparison to pooling (pool5) and convolution (conv5) layers. 

\begin{table}[t]
\caption{Classification accuracy (\%) achieved with the features extracted from fine-tuned BVLC Net.}
\centering
\begin{tabular}{c c c c c c c}
 \toprule
\multirow{2}{*}{Classification} &  \multicolumn{3}{c}{Before Fine-Tuning} & \multicolumn{3}{c}{After Fine-Tuning}\\ 
\cline{2-7}
 & fc7 & pool5 & conv5 & fc7 & pool5 & conv5\\
 \midrule
 \midrule
grade 0 vs grade 1 & 62.0 &  64.3 &  63.3 & 63.3 & \textbf{64.3} & 61.9\\  
grade 0 vs grade 2 & 69.6 &  73.6 &  73.9 & 76.3 & \textbf{77.2} & 74.1\\ 
grade 0 vs grade 3 & 87.9 &  92.5 &  91.5 & \textbf{96.7} & 96.0 & 96.3\\ 
grade 0 vs grade 4 & 98.5 &  99.4 &  99.1 & \textbf{99.8} & 99.7 & 99.7\\  
 \midrule
grade 1 vs grade 2 & 61.2 &  65.8 &  62.8 & 63.3 & \textbf{66.7} & 62.7\\ 
grade 2 vs grade 3 & 70.3 &  78.1 &  77.1 & \textbf{85.8} & 83.9 & 83.3\\ 
grade 3 vs grade 4 & 87.4 &  91.6 &  91.4 & \textbf{94.4} & 93.6 & 92.6\\  
 \midrule
grade 0 to grade 2 & 51.1 &  54.8 &  54.4 & \textbf{57.4} & 57.0 & 52.0\\  
grade 0 to grade 3 & 46.9 &  51.6 &  50.2 & \textbf{57.2} & 56.5 & 51.8\\ 
grade 0 to grade 4 & 44.1 &  50.8 &  50.0 & \textbf{57.6} & 56.2 & 51.8\\ 
\bottomrule
\end{tabular}
\label{Tab:Clsf_FT}
\end{table}

\subsubsection*{Regression using Fine-tuned CNNs.}
Existing work on automatic assessment of knee OA severity treats it as an image classification problem, assigning each KL grade to a distinct category \cite{shamir2009knee,thomson2015automated,shamir2009early,orlov2008wnd}. To date, evaluation of automatic KL grading algorithms has been based on binary and multi-class classification accuracy with respect to these discrete KL grades \cite{oka2008fully,shamir2009early,orlov2008wnd}. Nevertheless, KL grades are not categorical, but rather represent an ordinal scale of increasing severity. Treating them as categorical during evaluation means that the penalty for incorrectly predicting that a subject with grade 0 OA has grade 4 is the same as the penalty for predicting that the same subject has grade 1 OA. Clearly the former represents a more serious error, yet this is not captured by evaluation measures that treat grades as categorical variables \cite{antony2016}. In this set up, permuting the ordering of the grades has no effect on classification performance. Moreover, the quantisation of the KL grades to discrete integer levels is essentially an artefact of convenience; the true progression of the disease in nature is continuous, not discrete.

The author proposes that it is more appropriate to measure the performance of an automatic knee OA severity assessment system using a continuous evaluation metric like mean squared error. Such a metric appropriately penalises errors in proportion to their distance from the ground truth, rather than treating all errors equally. Directly optimising mean squared error on a training set also naturally leads to the formulation of knee OA assessment as a standard regression problem. Treating it as such provides the model with more information on the structure and relationship between training examples with successive KL grades. It is demonstrated that the use of regression reduces both the mean squared error and improves the multi-class classification accuracy of the model \cite{antony2016}.

The pre-trained BVLC CaffeNet model is fine-tuned using both classification loss (cross entropy on softmax outputs) and regression loss (mean squared error) to compare their performance in assessing knee OA severity. In both cases, the fully connected layer fc7 is replaced with a randomly initialised layer and fine-tuned for 20 epochs, selecting the model with the highest validation performance. The classification network uses a 5D fully connected layer and softmax following the fc7 layer, and the regression network uses a 1D fully connected node with a linear activation. 

The models are compared using both mean squared error (MSE) and standard multi-class classification metrics. The mean squared error is calculated using the standard formula:

$$ MSE = \frac{1}{n} \sum_{i=1}^{n}(y_{i} - \hat{y_{i}})^{2},$$ where $n$ is the number of test samples, $y_{i}$ is the true (integer) label and $\hat{y_{i}}$ is the predicted label. For the classification network the predicted labels $y_{i}$ are integers and for the regression network they are real numbers. A configuration is tested, where the real outputs are rounded from the regression network to produce integer labels. Table~\ref{Tab:MSE} shows the MSE for classification using the WNDCHRM and the CNN trained with classification loss (CNN-Clsf), regression loss (CNN-Reg), and regression loss with rounding (CNN-Reg*). Regression loss clearly achieves significantly lower mean squared error than both the CNN classification network and the WNDCHRM features.

\begin{table}[t]
\caption{MSE for classification and regression.}
\label{Tab:MSE}
\centering
\begin{tabular}{c c c c c}
\toprule
Classes & WNDCHRM & CNN-Clsf & CNN-Reg & CNN-Reg*\\
\midrule
\midrule
 %grade 0 to grade 2  & 0.835 & \textbf{0.516} & 0.596  \\
 %grade 0 to grade 3  & 0.841 & \textbf{0.509} & 0.585  \\
 grade 0 to 4  & 2.459  & 0.836 & \textbf{0.504} & 0.576  \\
\bottomrule
\end{tabular}
\end{table}

To demonstrate that the regression loss also produces better classification accuracy, the classification accuracy from the network trained with classification loss and the network trained with regression loss and rounded labels are compared. Rounding in this case is necessary to allow the use of standard classification metrics. Table~\ref{Tab:Clsf_stats} compares the resulting precision, recall, and $F_{1}$ scores. The multi-class (grade 0--4) classification accuracy of the network fine-tuned with regression loss is 59.6\%. The network trained using regression loss clearly gives superior classification performance. The author suspects this is due to the fact that using regression loss gives the network more information about the ordinal relationship between the KL grades, allowing it to converge on parameters that better generalise to unseen data.

\begin{table}[t]
\caption{Comparison of classification performance using classification (left) and regression (right) losses.}
\label{Tab:Clsf_stats}
\centering
\begin{tabular}{c c c c c c c}
\toprule
\multirow{2}{*}{Classification} &  \multicolumn{3}{c}{Classification Loss} & \multicolumn{3}{c}{Regression Loss}\\ 
\cline{2-7}
 & Precision & Recall & $F_{1}$ & Precision & Recall & $F_{1}$\\
\midrule
\midrule
 0 & 0.53 & 0.64 & 0.58   & 0.57 & 0.92 & 0.71 \\
 1 & 0.25 & 0.19 & 0.22   & 0.32 & 0.14 & 0.20 \\
 2 & 0.44 & 0.32 & 0.37   & 0.71 & 0.46 & 0.56 \\
 3 & 0.37 & 0.47 & 0.41   & 0.78 & 0.73 & 0.76 \\
 4 & 0.56 & 0.54 & 0.55   & 0.89 & 0.73 & 0.80 \\
 \midrule
 Mean & 0.43 & 0.44 & 0.43 & 0.61 & 0.62 & 0.59\\
 \bottomrule
\end{tabular}
\end{table}

\subsubsection*{Discussion.}
The initial approach to quantify knee OA severity used features extracted from pre-trained CNNs. Three pre-trained networks are investigated and it is found that the BVLC reference CaffeNet and VGG-M-128 networks perform best. A linear SVM trained on features from these networks achieved significantly higher classification accuracy (53.4\%) in comparison to the previous state-of-the-art (28.9\%). The features from pooling and convolutional layers were found to be more accurate than the fully connected layers. Fine-tuning the networks by replacing the top fully connected layer gave further improvements in multi-class classification accuracy.

Previous studies have assessed their algorithms using binary and multi-class classification metrics. The author proposes that it is more suitable to treat KL grades as a continuous variable and assess accuracy using mean squared error. This approach allows the model to be trained using regression loss so that errors are penalised in proportion to their severity, producing more accurate predictions. This approach also has the nice property that the predictions can fall between grades, which aligns with continuous disease progression.

In summary, this section presented two approaches based on the existing pre-trained CNNs for quantifying knee OA severity: first, the CNNs were used for fixed feature extraction and next, the CNNs were fine-tuned using transfer learning. Both the approaches outperformed the previous state-of-the-art, the WNDCHRM classifier, giving promising results. As a next logical step, CNNs are trained from scratch to investigate if this leads to further improvement in quantifying knee OA severity.

\subsubsection{Training CNNs from Scratch}˜\\
Training a CNN from scratch (or full training) is challenging and complicated, because it requires a large amount of annotated training data. The learning curves during training should ensure proper convergence to generalise well avoiding overfitting \cite{tajbakhsh2016conv}. An alternative to full training is transfer learning, fine-tuning CNNs pre-trained in other domain (for instance ImageNet dataset with natural images) to a target domain, for instance medical domain. However, the knowledge transfer may be limited by the substantial differences between the source and the target domains, which may mitigate the performance of the fine tuned CNNs. Nevertheless, with sufficient labelled training data and carefully selected hyper-parameters, fully trained CNNs can outperform fine-tuned CNNs and hand-crafted alternatives \cite{tajbakhsh2016conv,litjens2017survey}. 

Fully trained CNNs have been found to be highly successful in many medical applications \cite{tajbakhsh2016conv,litjens2017survey}. Some of the applications that use fully trained CNNs for musculo-skeletal (including knee) image analysis are knee cartilage segmentation using multi-stream CNNs \cite{prasoon2013deep}, total knee arthoplasty kinematics by real-time 2D/3D registration using CNN regressors \cite{miao2016real}, automated skeletal bone age assessment in X-ray images using deep learning \cite{spampinato2017deep}, and posterior-element fractures detection on spine CT using deep convolutional networks \cite{roth2016deep}. Motivated by these approaches, CNNs are trained from scratch to quantify knee OA severity using both classification and regression.

\subsubsection*{Dataset and Preprocessing.}
The data used for the initial experiments are taken from the baseline OAI dataset. There are 4,446 X-ray images with the KL grade annotations in this dataset. The MOST dataset is included for later experiments and this dataset consists of 2,920 X-ray images with KL grade annotations. Two set of knee joint images are used separately for the experiments: 1) extracted after automatic localisation and 2) extracted after manual annotation of the ROI. This is to compare the quantification performance of the CNNs trained with knee joints from automatic localisation and manual annotation. As a preprocessing step, all the knee joint images are subjected to histogram equalisation for intensity level normalisation. The images were resized to 256$\times$256 pixels for the initial experiments. Later, the input image size is changed to 200$\times$300. This size is chosen to approximately preserve the aspect ratio  based on the mean aspect ratio (1.6) of all the extracted knee joints.  Right-left flip of the knee joint images are used to generate more training data. 

\subsubsection*{Initial Configuration.}
A CNN is configured with a lightweight architecture with 4 layers of learned weights: 3 convolutional layers and 1 fully connected layer. As the training data set is relatively small,  a lightweight architecture is considered with minimal (4.5 million) parameters in comparison to the existing CNNs. Table \ref{Tab:cnn1} shows the CNN configuration in detail. Each convolutional layer is followed by batch normalisation and a ReLU activation layer. A max pooling layer is included after each convolution stage. The final pooling layer is followed by a fully connected layer (fc4), and a softmax dense layer (fc5) with an output shape 5 for the multi-class classification of (0--4) ordinal KL grades. A drop out layer with a drop out ratio of 0.5 is included after the fully connected layer (fc5) to avoid overfitting. The input images are of size 256$\times$256 pixels and fed to the network after sub-sampling by a factor of 2. So, the input size is 128$\times$128 pixels.

\begin{table}[t]
\caption{Initial CNN configuration.}
\label{Tab:cnn1}
\centering
\begin{tabular}{l c c c c }
\toprule 
Layer & Kernels & Kernel Size & Strides & Output shape\tabularnewline
\midrule
\midrule
conv1 & 32 & 11 $\times$11 & 2 & 32$\times$128$\times$128 \tabularnewline
maxPool1 & -- & 3$\times$3 & 3 & 32$\times$42$\times$42 \tabularnewline
conv2 & 96 & 7$\times$7 & 1 & 96$\times$42$\times$42 \tabularnewline
maxPool2 & -- & 3$\times$3 & 3 & 96$\times$14$\times$14 \tabularnewline
conv3 & 128 & 3$\times$3 & 1 & 128$\times$14$\times$14  \tabularnewline
maxPool3 & -- & 3$\times$3 & 2 & 128$\times$4$\times$4 \tabularnewline
fc4 & -- & -- & -- & 2048 \tabularnewline
fc5 & -- & -- & -- & 5 \tabularnewline
\bottomrule
\end{tabular}
\end{table}

\subsubsection*{Training Process and Initial Results.}
The network parameters are trained from scratch with the knee joint images as training samples and the KL grades (0, 1, 2, 3 or 4) as labels. To start, the knee joint images extracted manually from the radiographs of the OAI dataset are used. The dataset is split into training (70\%) and test (30\%) sets. The validation (10\%) data is taken from the training set. The network is trained to minimise categorical cross entropy for multi-class classification. \textit{Stochastic gradient descent} (SGD) is used with default parameters: decay = $1\mathrm{e}^{-6}$, momentum = $0.9$, and nesterov = True and the initial learning rate is set to 0.0001. The networks are trained with fixed learning rate in the initial experiments. The Adam optimiser with default parameters: initial learning rate $(\alpha) = 0.001$, $\beta_{1} = 0.9$, $\beta_{2} = 0.999$, $\epsilon = 1\mathrm{e}^{-8}$ is tested, instead of SGD for the later experiments. The benefits of the Adam optimiser are that it uses adaptive learning rates and provides faster convergence.  

This network achieves a multi-class classification accuracy of 44.7\% on the test data. The mean-squared error is 1.75. Table \ref{Tab:cnn1_res} shows the classification results: precision, recall, and $F_1$ score of the initial configuration. The results show that the classification performance is low and the mean-squared error is high. These are initial results and the hyper-parameters of this network are tuned to improve the classification performance. Further, the number of convolutional layers, convolutional-pooling stages, the number of convolutional kernels, kernel sizes and other parameters are experimented. 

% Remove this paragraph, if not required.
The terms `parameters' and `hyper-parameters' in machine learning are often used interchangeably, but there is a difference between them. Parameters are learned by a classifier or a machine learning model from the training data, for instance weights or coefficients of the independent variables. Hyper-parameters are the settings used to optimise the performance of a classifier or a model and they are not fit based on the training data. The hyper-parameters for a CNN include the number and size of the hidden layers, learning rate and its decay, drop out regularisation, gradient clipping threshold and other settings. 

\begin{table}[t]
\caption{Classification results of the initial CNN configuration.}
\label{Tab:cnn1_res}
\centering
\begin{tabular}{c c c c}
\toprule
grade & Precision & Recall & $F_{1}$ Score \\
\midrule
\midrule
 0 & 0.45 & 0.92 & 0.60 \\
 1 & 0.24 & 0.07 & 0.11 \\
 2 & 0.49 & 0.18 & 0.26 \\
 3 & 0.50 & 0.39 & 0.44 \\
 4 & 1.00 & 0.01 & 0.02 \\
 \midrule
 Mean & 0.45 & 0.45 & 0.37 \\
 \bottomrule
\end{tabular}
\end{table}

\subsubsection*{Tuning Hyper-parameters.}
After the initial CNN configuration giving low classification accuracy (44.7\%), as a first step the depth of the network is increased. A convolutional layer and a pooling layer are included. This increases the number of layers with learned weights to 5 layers: 4 convolutional layers and 1 fully connected layer. SGD with default parameters: decay = $1\mathrm{e}^{-6}$, momentum = $0.9$, and nesterov = True, is used for training this network. Learning rates from 0.0001 to 0.01 with an incremental increase by a factor 10 are tested, and  the learning rate 0.001 is found to be the best. After experimenting with the convolutional kernel size, the number of kernels in the convolutional layer, the number of outputs of the fully connected layer and other parameters, the final architecture in this configuration is obtained. Table \ref{Tab:cnn2} shows the CNN architecture in detail. 

\begin{table}[t]
\caption{CNN architecture (CNN-1) after tuning hyper-parameters.}
\label{Tab:cnn2}
\centering
\begin{tabular}{l c c c c}
\toprule 
Layer & Kernels & Kernel Size & Strides & Output shape\tabularnewline
\midrule
\midrule
conv1 & 32 & 11$\times$11 & 2 & 32$\times$128$\times$128 \tabularnewline
maxPool1 & -- & 3$\times$3 & 2 & 32$\times$63$\times$63 \tabularnewline
conv2 & 96 & 5$\times$5 & 1 & 64$\times$63$\times$63 \tabularnewline
maxPool2 & -- & 3$\times$3 & 2 & 64$\times$31$\times$31 \tabularnewline
conv3 & 128 & 3$\times$3 & 1 & 128$\times$31$\times$31  \tabularnewline
maxPool3 & -- & 3$\times$3 & 2 & 128$\times$15$\times$15 \tabularnewline
conv4 & 256 & 3$\times$3 & 1 & 256$\times$15$\times$15  \tabularnewline
maxPool4 & -- & 3$\times$3 & 2 & 256$\times$7$\times$7 \tabularnewline
fc5 & -- & -- & -- & 1024 \tabularnewline
fc6 & -- & -- & -- & 5 \tabularnewline
\bottomrule
\end{tabular}
\end{table}

\begin{figure}[t]
\centering
\includegraphics[scale=0.6]{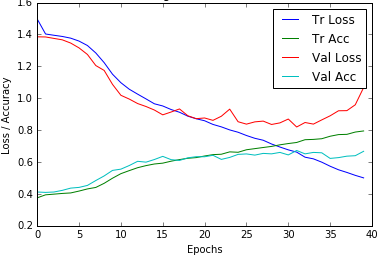}
\caption{Learning curves: training and validation losses, and accuracies of the fully trained CNN} 
\label{fig:LC_cnn2}
\end{figure}

After 20 epochs of training, this network gave a multi-class classification accuracy of 55.2\% with a mean-squared error 0.803 on the validation data. After 35 epochs the network achieves the best results for this configuration with a classification accuracy of 60.4\% and mean-squared error 0.838. Table \ref{Tab:cnn2_res} shows the classification results: precision, recall, and $F_1$ score of this network. There is an improvement in the overall classification results in comparison to the previous results (Table \ref{Tab:cnn1_res}). Figure \ref{fig:LC_cnn2} shows the learning curves with increase in the training and validation accuracies, and decrease in the training and validation losses whilst training this network. It can be observed from the learning curves (Figure \ref{fig:LC_cnn2}), after 32 epochs there is an increase in validation loss with decrease in training loss and also there is no further increase in validation accuracy whilst training accuracy increases: the network is starting to overfit. A drop out regularisation by a ratio of 0.5 is included after the fully connected layer (fc5) to mitigate overfitting. Also, data augmentation is used to increase the training samples by including the right-left flip of the knee joints and this doubles the number of training samples. Drop out regularisation after convolutional layers and fully connected layers, and l2-norm weight regularisations are used to further mitigate overfitting in the next set of experiments. 

\begin{table}[t]
\caption{Classification results after tuning hyper-parameters.}
\label{Tab:cnn2_res}
\centering
\begin{tabular}{c c c c}
\toprule
grade & Precision & Recall & $F_{1}$ Score \\
\midrule
\midrule
 0 & 0.57 & 0.90 & 0.70 \\
 1 & 0.31 & 0.11 & 0.16 \\
 2 & 0.64 & 0.45 & 0.53 \\
 3 & 0.74 & 0.77 & 0.76 \\
 4 & 0.86 & 0.72 & 0.78 \\
 \midrule
 Mean & 0.58 & 0.60 & 0.57 \\
 \bottomrule
\end{tabular}
\end{table}

Next, the depth of the network is further increased by increasing the number of layers with learned weights, continuing the experimentation with the other associated hyper-parameters. Up to 5 convolutional-pooling stages followed by two fully connected layers are tested. The classification accuracy with 4 convolutional-pooling stages is 60.8\% and with 5 convolutional-pooling stages is 61\%.

\begin{table}[t]
\caption{CNN architecture (CNN-2) after tuning hyper-parameters.}
\label{Tab:cnn3}
\centering
\begin{tabular}{l c c c c}
\toprule 
Layer & Kernels & Kernel Size & Strides & Output shape\tabularnewline
\midrule
\midrule
conv1 & 32 & 11$\times$11 & 2 & 32$\times$128$\times$128 \tabularnewline
maxPool1 & -- & 3$\times$3 & 2 & 32$\times$63$\times$63 \tabularnewline

conv2 & 64 & 5$\times$5 & 1 & 64$\times$63$\times$63 \tabularnewline
maxPool2 & -- & 3$\times$3 & 2 & 64$\times$31$\times$31 \tabularnewline

conv3-1 & 64 & 3$\times$3 & 1 & 64$\times$31$\times$31  \tabularnewline
conv3-2 & 64 & 3$\times$3 & 1 & 64$\times$31$\times$31  \tabularnewline
maxPool3 & -- & 3$\times$3 & 2 & 64$\times$15$\times$15 \tabularnewline

conv4-1 & 96 & 3$\times$3 & 1 & 96$\times$15$\times$15  \tabularnewline
conv4-2 & 96 & 3$\times$3 & 1 & 96$\times$15$\times$15  \tabularnewline
maxPool4 & -- & 3$\times$3 & 2 & 96$\times$7$\times$7 \tabularnewline

fc5 & -- & -- & -- & 1024 \tabularnewline
fc6 & -- & -- & -- & 5 \tabularnewline
\bottomrule
\end{tabular}
\end{table}

Previous networks use a single convolutional layer followed by a pooling layer. Next, cascaded convolutional layers are used in a convolution-pooling stage like VGG-16 model. Each convolutional layer is followed by a ReLU activation. Figure \ref{Tab:cnn3} shows the CNN architecture that gives the best results in this approach. This network gives a classification accuracy of 60.1\% with a mean-squared error 0.838. 

Inspired by the success of VGG networks \cite{simonyan2014very}, a network with cascaded convolutional layers of uniform (3$\times$3) kernel size and (2$\times$2) max pooling with stride 2 is trained, and the hyper-parameters are tuned. This network gives a classification accuracy of 57.5\% with a mean-squared error 0.961. There is no further improvement in the classification results in comparison to the previous results. 

\subsubsection*{Training Off-the-shelf CNNs from Scratch.}
Earlier, the widely used off-the-shelf CNNs such as BVLC reference CaffeNet \cite{jia2014caffe,karayev2013recognizing} (which is very similar to the AlexNet model \cite{krizhevsky2012imagenet}), VGG-M-128 network \cite{chatfield2014return}, and VGG-16 network \cite{simonyan2014very} were fine-tuned for knee images classification. The pre-trained VGG-16 network has $\sim$138 million free parameters, and the other networks, Alexnet with $\sim$62 million and the VGG-M-128 with $\sim$26 million parameters, are relatively simple. Training these networks, in particular VGG-16, from scratch is computationally very expensive due to the depth and the number of free parameters. Previously trained CNNS have relatively fewer parameters ($\sim$4 to 6 million) to suit the relatively small dataset with a few thousand of training examples. 

Next, CNNs are fully trained using the AlexNet and the VGG-M-128 architectures. This is to compare the classification performance of these networks to the previously trained networks from scratch. Table \ref{Tab:AlexNet} shows the AlexNet architecture in detail. The convolutional layers conv1 and conv2 in this network are followed by Relu and batch normalisation layers. The two fully connected layers (fc6) and (fc7) are followed by a drop out regularisation by a ratio 0.5. This network was pre-trained for 1,000 classes in the ImageNet \cite{imagenet_cvpr09} dataset. The output of the last fully connected layer (fc8) is replaced with a 5 output dense layer for multi-class knee OA image classification. This network is trained using SGD with default parameters. Learning rates from 0.00001 to 0.01 with an incremental increase by a factor 10 are tested. The learning rate set at 0.001 gives the best results. 

% AlexNet

\begin{table}[t]
\caption{AlexNet architecture.}
\label{Tab:AlexNet}
\centering
\begin{tabular}{l c c c c}
\toprule 
Layer & Kernels & Kernel Size & Strides & Output shape\tabularnewline
\midrule
\midrule
conv1 & 96 & 11$\times$11 & 4 & 96$\times$64$\times$64 \tabularnewline
maxPool1 & -- & 3$\times$3 & 2 & 96$\times$31$\times$31 \tabularnewline

conv2 & 256 & 5$\times$5 & 1 & 256$\times$31$\times$31 \tabularnewline
maxPool2 & -- & 3$\times$3 & 2 & 256$\times$15$\times$15 \tabularnewline

conv3 & 384 & 3$\times$3 & 1 & 384$\times$15$\times$15  \tabularnewline
conv4 & 384 & 3$\times$3 & 1 & 384$\times$15$\times$15  \tabularnewline
conv5 & 256 & 3$\times$3 & 1 & 256$\times$15$\times$15  \tabularnewline
maxPool5 & -- & 3$\times$3 & 2 & 256$\times$7$\times$7 \tabularnewline

fc6 & -- & -- & -- & 4096 \tabularnewline
fc7 & -- & -- & -- & 4096 \tabularnewline
fc8 & -- & -- & -- & 5 \tabularnewline
\bottomrule
\end{tabular}
\end{table}

\begin{figure}[t]
\centering
\includegraphics[width=0.75\textwidth]{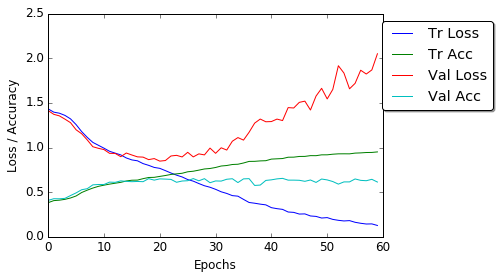}
\caption{Learning curves: training and validation losses, and accuracies of the fully trained AlexNet.} 
\label{fig:Lc_AlexNet}
\end{figure}

The fully trained AlexNet gives a classification accuracy of 57.2\% with a mean-squared error 0.741. Table \ref{Tab:AlexNet_res} shows the classification results; precision, recall, and $F_{1}$ score of the fully trained AlexNet model. These results show that the classification accuracy achieved by the fully trained AlexNet is low (57.2\%) in comparison to the accuracy (60.8\%) achieved by previous networks. Moreover, this network is overfitting. This is evident from the learning curves (Figure \ref{fig:Lc_AlexNet}) obtained whilst training this network. After 30 epochs, the learning curves show an increase in validation loss whilst the training loss is decreasing and there is no improvement in the validation accuracy whilst the training accuracy keeps increasing. The reason for overfitting is the number of training samples in the dataset ($\sim$10,000) is very low in comparison to the number of free parameters ($\sim$62 million) in AlexNet. This model was originally developed and trained on datasets like ImageNet \cite{imagenet_cvpr09} that consists of more than $\sim$1.2 million images. There are two fully connected layers with 4,096 outputs in the AlexNet and these layers contribute to more than 95\% of the total free parameters in this network. Next, a relatively simple architecture (VGG-M-128) is investigated for the knee OA images classification.

\begin{table}[t]
\caption{Classification results of the fully trained AlexNet.}
\label{Tab:AlexNet_res}
\centering
\begin{tabular}{c c c c}
\toprule
grade & Precision & Recall & $F_{1}$ Score \\
\midrule
\midrule
 0 & 0.65 & 0.61 & 0.63 \\
 1 & 0.29 & 0.36 & 0.32 \\
 2 & 0.59 & 0.55 & 0.57 \\
 3 & 0.75 & 0.73 & 0.74 \\
 4 & 0.77 & 0.79 & 0.78 \\
 \midrule
 Mean & 0.59 & 0.57 & 0.58 \\
 \bottomrule
\end{tabular}
\end{table}

% VGG-M-128 Network
The VGG-M-128 network is a simplified model of the AlexNet \cite{simonyan2014very}. The last fully connected layer (fc7) of AlexNet has 4,096 outputs. The number of fc7 outputs is reduced to 128 in VGG-M-128. This reduces the number of free parameters and this network contains ($\sim$26 million) parameters in total. The AlexNet configuration is retained in the VGG-M-128 network with a few changes in the architecture. The kernel size of the first convolutional layer is reduced to (7$\times$7) and the stride is reduced to 2. The number of filters is fixed to 512 in the conv3, conv4, and conv5 layers. Table \ref{Tab:VggM128} shows the architecture details. This network parameters are trained from scratch using SGD with default parameters: decay = $1\mathrm{e}^{-6}$, momentum = $0.9$, and nesterov = True. The learning rate is fixed to 0.001 after testing different rates like before.

\begin{table}[t]
\caption{VGG-M-128 architecture.}
\label{Tab:VggM128}
\centering
\begin{tabular}{l c c c c}
\toprule 
Layer & Kernels & Kernel Size & Strides & Output shape\tabularnewline
\midrule
\midrule
conv1 & 96 & 7$\times$7 & 2 & 96$\times$128$\times$128 \tabularnewline
maxPool1 & -- & 3$\times$3 & 2 & 96$\times$63$\times$63 \tabularnewline

conv2 & 256 & 5$\times$5 & 1 & 256$\times$32$\times$32 \tabularnewline
maxPool2 & -- & 3$\times$3 & 2 & 256$\times$15$\times$15 \tabularnewline

conv3 & 512 & 3$\times$3 & 1 & 512$\times$15$\times$15  \tabularnewline
conv4 & 512 & 3$\times$3 & 1 & 512$\times$15$\times$15  \tabularnewline
conv5 & 512 & 3$\times$3 & 2 & 512$\times$8$\times$8  \tabularnewline
maxPool5 & -- & 3$\times$3 & 2 & 512$\times$3$\times$3 \tabularnewline

fc6 & -- & -- & -- & 4096 \tabularnewline
fc7 & -- & -- & -- & 128 \tabularnewline
fc8 & -- & -- & -- & 5 \tabularnewline
\bottomrule
\end{tabular}
\end{table}

\begin{figure}[t]
\centering
\includegraphics[width=0.75\textwidth]{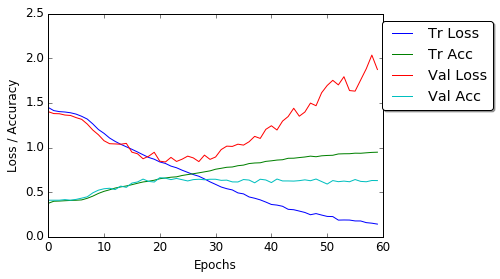}
\caption{Learning curves: training and validation losses, and accuracies of the fully trained VGG-M-128 network.} 
\label{fig:Lc_VggM128}
\end{figure}

This network gives a classification accuracy of 56.3\% and the mean-squared error is 0.685. Table \ref{Tab:VggM128_res} shows the classification results of this network. The results show a slightly lower classification accuracy (56.3\%) in comparison to the previous results. There is no significant difference in the precision, recall, and $F_{1}$ score of this network in comparison to the AlexNet classification results (Table \ref{Tab:AlexNet_res}). This network is also overfitting like the AlexNet. This is evident from the learning curves (Figure \ref{fig:Lc_VggM128}) of this network. The learning curves show increase in the validation loss after 30 epochs and the validation accuracy remains almost the same. The drop out regularisations after the fully connected layers fc6 and fc7 are not able to fully mitigate overfitting. The reason for overfitting remains the same as for AlexNet. The number of training samples is very low even for the number of free parameters in this network ($\sim$26 million). 

\begin{table}[t]
\caption{Classification results of the fully trained VGG-M-128.}
\label{Tab:VggM128_res}
\centering
\begin{tabular}{c c c c}
\toprule
grade & Precision & Recall & $F_{1}$ Score \\
\midrule
\midrule
 0 & 0.66 & 0.65 & 0.66 \\
 1 & 0.27 & 0.42 & 0.33 \\
 2 & 0.62 & 0.46 & 0.53 \\
 3 & 0.77 & 0.69 & 0.72 \\
 4 & 0.87 & 0.73 & 0.79 \\
 \midrule
 Mean & 0.60 & 0.56 & 0.58 \\
 \bottomrule
\end{tabular}
\end{table}

\subsubsection*{Best Performing CNN for Classification.}
After experimenting with different configurations, the network in Table \ref{Tab:bestClsf} is found to be the best for classifying knee images. This network is similar to the previous configuration (Table \ref{Tab:cnn3}), but with slight variations. The network contains five layers of learned weights: four convolutional layers and a fully connected layer. The total number of free parameters in the network is $\sim$5.4 million. Each convolutional layer in the network is followed by batch normalisation and a ReLU activation layer. After each convolutional stage there is a max pooling layer. The final pooling layer (maxPool4) is followed by a fully connected layer (fc5) and a softmax dense (fc6) layer. To avoid overfitting, a drop out layer with a drop out ratio of 0.25 is included after the last convolutional (conv4) layer and a drop out layer with a drop out ratio of 0.5 after the fully connected layer (fc5). Also, a L2-norm weight regularisation penalty of 0.01 is applied in the last two convolutional layers (conv3 and conv4) and the fully connected layer (fc5). Applying a regularisation penalty to other layers increases the training time whilst not introducing significant variation in the learning curves. The network is trained to minimise categorical cross-entropy loss using the Adam optimiser with default parameters: initial learning rate $(\alpha) = 0.001$, $\beta_{1} = 0.9$, $\beta_{2} = 0.999$, $\epsilon = 1\mathrm{e}^{-8}$. The inputs to the network are knee images of size 200$\times$300. This size is selected to approximately preserve the aspect ratio  based on the mean aspect ratio (1.6) of all the extracted knee joints.

\begin{table}[t]
\caption{Best performing CNN for classifying the knee images.}
\label{Tab:bestClsf}
\centering
\begin{tabular}{l c c c c}
\toprule 
Layer & Kernels & Kernel Size & Strides & Output shape\tabularnewline
\midrule
\midrule
conv1 & 32 & 11$\times$11 & 2 & 32$\times$100$\times$150 \tabularnewline
maxPool1 & -- & 3$\times$3 & 2 & 32$\times$49$\times$74 \tabularnewline
conv2 & 64 & 5$\times$5 & 1 & 64$\times$49$\times$74 \tabularnewline
maxPool2 & -- & 3$\times$3 & 2 & 64$\times$24$\times$36 \tabularnewline
conv3 & 96 & 3$\times$3 & 1 & 96$\times$24$\times$36  \tabularnewline
maxPool3 & -- & 3$\times$3 & 2 & 96$\times$11$\times$17 \tabularnewline
conv4 & 128 & 3$\times$3 & 1 & 128$\times$11$\times$17  \tabularnewline
maxPool4 & -- & 3$\times$3 & 2 & 128$\times$5$\times$8 \tabularnewline
fc5 & -- & -- & -- & 1024 \tabularnewline
fc6 & -- & -- & -- & 5 \tabularnewline
\bottomrule
\end{tabular}
\end{table}

\begin{figure}[t]
\centering
\includegraphics[width=0.75\textwidth]{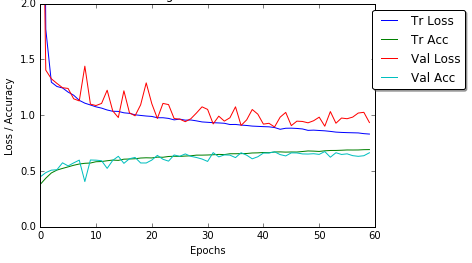}
\caption{Learning curves: training and validation losses, and accuracies of the best performing fully trained CNN.} 
\label{fig:Lc_bestClsf}
\end{figure}

First, this network is trained using the OAI dataset like the previous network trainings. This network achieves a classification accuracy of 61\% with a mean-squared error 0.861. Next, training samples are included from the MOST dataset. This network achieves a classification accuracy of 61.8\% with a mean-squared error 0.735 for the combined OAI-MOST dataset. There is a slight increase in the classification accuracy (0.8\%) and decrease in the mean-squared error (0.126). Table \ref{Tab:bestClsf_res} shows the classification results: precision, recall, and $F_{1}$ score of this network for the combined OAI-MOST dataset. Figure \ref{fig:Lc_bestClsf} shows the learning curves whilst training this network. The learning curves show proper convergence of the training and validation losses with consistent increase in the training and validation accuracies till they reach constant values. 

\begin{table}[t]
\caption{Classification results of the best performing fully trained CNN.}
\label{Tab:bestClsf_res}
\centering
\begin{tabular}{c c c c}
\toprule
grade & Precision & Recall & $F_{1}$ Score \\
\midrule
\midrule
 0 & 0.65 & 0.83 & 0.73 \\
 1 & 0.30 & 0.10 & 0.14 \\
 2 & 0.51 & 0.60 & 0.55 \\
 3 & 0.77 & 0.69 & 0.73 \\
 4 & 0.87 & 0.70 & 0.78 \\
 \midrule
 Mean & 0.59 & 0.62 & 0.59 \\
 \bottomrule
\end{tabular}
\end{table}

To sum up, a high classification accuracy (61\%) is achieved with the CNN (Table \ref{Tab:bestClsf}) trained from scratch and outperform the VGG-M-128 and the AlexNet trained from scratch. The fully trained AlexNet gives a classification accuracy of 57.2\% and VGG-M-128 gives an accuracy of 56.3\%. The classification results of the methods proposed in this section and the previous state-of-the-art are compared in the next section. 

\subsubsection*{Classification Results.}
The classification results of the fully trained network is compared to WNDCHARM, the multi purpose medical image classifier \cite{orlov2008wnd,shamir2008wndchrm,shamir2013wnd} that gave the previous best results for automatically classifying knee OA X-ray images, and to previous results (Table \ref{Tab:Clsf_FT}) on fine-tuning BVLC reference CaffeNet for this task (Section 5.2.3). WNDCHARM is trained with the data taken from the OAI and MOST datasets.

\begin{table}[t]
\caption{Classification results of the proposed methods and the existing methods.}
\label{Tab:ClsfRes}
\centering
\begin{tabular}{l c c c}
\toprule
Method & Test Data & Accuracy & Mean-Squared Error\\
\midrule
\midrule
 Wndchrm & OAI {\&} MOST & 34.8\% & 2.112 \\
 Fine-Tuned BVLC CaffeNet & OAI & 57.6\% & 0.836 \\
 Fully trained CNN & OAI & 61\% & 0.861 \\
 Fully trained CNN & OAI {\&} MOST & \textbf{61.8\%} & \textbf{0.735} \\
\bottomrule
\end{tabular}
\end{table}

Table \ref{Tab:ClsfRes} shows the multi-class classification accuracy and mean-squared error of the fine-tuned BVLC CaffeNet, the network trained from scratch and WND-CHARM for the OAI and MOST datasets. The results show that the network trained from scratch for classifying knee OA images clearly outperforms WNDCHARM. This shows learning feature representations using CNNs for fine-grained knee OA images classification is highly effective and a better approach in comparison to using a combination of hand-crafted features in WNDCHARM. The other reason for low classification accuracy of WNDCHARM is that it uses only a balanced dataset for training. Both the OAI and MOST datasets are very unbalanced and in particular the number of knee images available in KL grade 4 is very small, $\sim$5\% in total.

Moreover, these results show an improvement over previous methods that used fine-tuned off-the-shelf networks such as VGG-M-128 and the BVLC Reference CaffeNet for classifying knee OA X-ray images through transfer learning. These improvements are due to the lightweight architecture of the network trained from scratch with less ($\sim$5.4 million) free parameters in comparison to 62 million free parameters of BVLC CaffeNet for the small amount of training data available. The off-the-shelf networks are trained using a large dataset like ImageNet containing  millions of images, whereas the dataset used in this experiment contains much fewer ($\sim$10,000) training samples. Furthermore, the results show an increase in classification accuracy from 61\% to 61.8\% when the MOST dataset is included in the training set. This result is promising and it shows that with more training data the CNN performance maybe further improved. Next, the use of regression by fully trained CNNs is investigated to improve the quantification performance. 

\subsubsection*{Training CNNs for Regression.}
CNNs are trained from scratch to classify knee images in the previous approach. The outcomes are ordinal KL grades (0, 1, 2, 3 or 4) that quantify knee OA severity. CNNs are trained for regression in the next approach. This is to assess knee OA severity in a continuous scale (0--4). The author argued earlier (Section 5.2.2) that it is more appropriate to assess knee OA in a continuous scale as knee OA is progressive in nature, not discrete \cite{antony2016}. The existing CNNs are fine-tuned to quantify knee OA severity using regression. 

\subsubsection*{Initial Configuration.}
A CNN is trained for regression using almost the same architecture (Table \ref{Tab:bestClsf}) that gave the highest multi-class classification accuracy previously. The last fully connected layer (fc6) with softmax activation and an output shape 5 for multi-class classification is replaced with a linear activation with an output shape of 1 for regression. The CNN is trained to minimise mean-squared error using the Adam optimiser with default parameters: initial learning rate $(\alpha) = 0.001$, $\beta_{1} = 0.9$, $\beta_{2} = 0.999$, $\epsilon = 1\mathrm{e}^{-8}$. Like before, the inputs to the network are images of size 200$\times$300. The data for training is taken from both the OAI and the MOST datasets. Both these datasets contain discrete KL grade (0, 1, 2, 3 or 4) annotations for the knee joints. These labels are used in the previous approach to train classifiers. However, there is no ground truth for KL grades on a continuous scale for either of these datasets to train a network directly for regression output. Hence, the discrete KL grades are used as labels to train CNNs for regression. 

\subsubsection*{Initial Results.}
This CNN gives a mean-squared error of 0.654 on the test data after training. In comparison to the mean-squared error achieved by the classifier (0.898) with almost the same architecture, there is definitely an improvement in the quantification using regression. The performance metrics: accuracy, precision, recall, and $F_{1}$ score are computed  for the regression results by rounding the predicted continuous grade to the next integer value. Rounding, in this case, is necessary to allow the use of standard classification metrics and to  compare the performances of classification and regression. Table \ref{Tab:reg1_res} shows the precision, recall, and $F_{1}$ score for regression. In comparing these results to the previous classification results (Table \ref{Tab:bestClsf_res}), there is a decrease in precision, recall, and $F_{1}$ score. The classification accuracy achieved by regression is 36.9\% with a mean-squared error 0.75. From these results it is evident that the regression performance is low in this initial configuration. Next, the hyper parameters of this network are tuned to improve the regression performance.  

\begin{table}[t]
\caption{Results of the initial network trained for regression after rounding the predicted continuous grades.}
\label{Tab:reg1_res}
\centering
\begin{tabular}{c c c c c c c}
\toprule
Grade & Precision & Recall & $F_{1}$ Score \\
\midrule
\midrule
 0 & 0.78 & 0.18 & 0.29 \\
 1 & 0.24 & 0.83 & 0.37 \\
 2 & 0.49 & 0.32 & 0.39 \\
 3 & 0.63 & 0.42 & 0.50 \\
 4 & 0.57 & 0.20 & 0.30 \\
 \midrule
 Mean & 0.57 & 0.37 & 0.36 \\
 \bottomrule
\end{tabular}
\end{table}

\subsubsection*{Tuning the Hyper-parameters.}
\begin{figure}[t]
\centering
\includegraphics[width=0.6\textwidth]{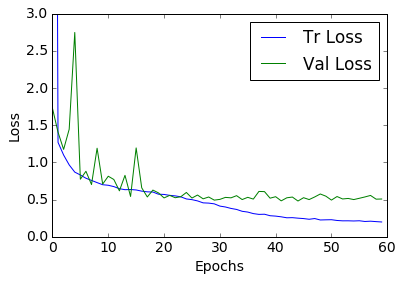}
\caption{Learning curves: training and validation losses for the best performing CNN for regression.}
\label{fig:Lc_bestReg}
\end{figure}

The experiment is continued by varying the number of layers with learned weights in the architecture, the number of convolutional-pooling stages, the number of kernels and kernel sizes in the convolutional layers and the regularisations to avoid overfitting. The architecture in Table \ref{Tab:bestReg} is found to be the best for quantifying knee OA severity using regression. This network contains seven layers of learned weights: six convolutional layers and a fully connected layer. This network has $\sim$5.6 million free parameters in total. Each convolutional layer is followed by batch normalisation and a ReLU layer. The last pooling layer (maxPool4) is followed by two dense layers: fc5 with ReLU and fc6 with linear activations. A drop out layer with a drop out ratio 0.5 is added after fc6. The network is trained to minimise the mean-squared error using the Adam optimiser with default parameters: initial learning rate $(\alpha) = 0.001$, $\beta_{1} = 0.9$, $\beta_{2} = 0.999$, $\epsilon = 1\mathrm{e}^{-8}$. The network is trained with knee images taken from the OAI and the MOST datasets. Figure \ref{fig:Lc_bestReg} shows the learning curves: training and validation losses whilst training this network. The learning curves show convergence in the losses.

\begin{table}[t]
\caption{Best performing CNN for regression of the knee images.}
\label{Tab:bestReg}
\centering
\begin{tabular}{l c c c c }
\toprule 
Layer & Kernels & Kernel Size & Strides & Output shape\tabularnewline
\midrule
\midrule
conv1 & 32 & 11$\times$11 & 2 & 32$\times$100$\times$158 \tabularnewline
maxPool1 & -- & 3$\times$3 & 2 & 32$\times$49$\times$74 \tabularnewline

conv2 & 64 & 5$\times$5 & 1 & 64$\times$49$\times$74 \tabularnewline
maxPool2 & -- & 3$\times$3 & 2 & 64$\times$24$\times$36 \tabularnewline

conv3-1 & 64 & 3$\times$3 & 1 & 64$\times$24$\times$36  \tabularnewline
conv3-2 & 64 & 3$\times$3 & 1 & 64$\times$24$\times$36  \tabularnewline
maxPool3 & -- & 3$\times$3 & 2 & 64$\times$11$\times$17 \tabularnewline

conv4-1 & 128 & 3$\times$3 & 1 & 96$\times$11$\times$17  \tabularnewline
conv4-2 & 128 & 3$\times$3 & 1 & 96$\times$11$\times$17  \tabularnewline
maxPool4 & -- & 3$\times$3 & 2 & 96$\times$5$\times$8 \tabularnewline

fc5 & -- & -- & -- & 1024 \tabularnewline
fc6 & -- & -- & -- & 1 \tabularnewline
\bottomrule
\end{tabular}
\end{table}

\subsubsection*{Comparison of Classification and Regression Results.}
The best performing CNN for regression gives a mean-squared error of 0.574. After rounding the continuous grade predictions, this network achieves a multi-class classification accuracy of 54.7\% and the mean-squared error is 0.661. Table \ref{Tab:Comp_ClsfReg} shows the accuracy and mean-squared error for the fully trained CNN for classification and regression. The results show that the multi-class classification accuracy calculated after rounding the output is low for CNN-regression. The main reason for this likely is training the regression network with ordinal labels instead of continuous labels. There is also a decrease in accuracy due to the rounding of regression output and the rounding is necessary to compute standard classification metrics. On the other hand, the mean-squared error of the fully trained CNN for regression is low in both the cases before rounding (0.574) and after rounding (0.661) in comparison to the fully trained CNN for regression. Table \ref{Tab:comp_metrics} shows the precision, recall, and $F_{1}$ score of the rounded regression output and the classification output. These results show that the network trained with classification loss outperforms the regression loss. The reason for this is again likely the lack of continuous KL grade ground truth to train a CNN directly for regression output. 

\begin{table}[t]
\caption{Comparison of classification and regression results.}
\label{Tab:Comp_ClsfReg}
\centering
\begin{tabular}{l c c c}
\toprule
Method & Accuracy & MSE (before rounding) & MSE (after rounding)\\
\midrule
\midrule
CNN-Classification & \textbf{61.8\%} & 0.735 & -- \\
CNN-Regression & 54.7\% & \textbf{0.574} & 0.661  \\
\bottomrule
\end{tabular}
\end{table}

\begin{table}[t]
\caption{Comparison of the regression and classification performances.}
\label{Tab:comp_metrics}
\centering
\begin{tabular}{c c c c c c c c c c}
\toprule
	\multirow{2}{*}{Grade}
      & \multicolumn{3}{c}{Regression} 
      & \multicolumn{3}{c}{Classification} \\
      \cmidrule{2-4} \cmidrule{5-7}
 & Precision & Recall & $F_{1}$ & Precision & Recall & $F_{1}$\\
\midrule
\midrule
 0 & 0.70 & 0.71 & 0.70   & 0.65 & 0.83 & 0.73 \\
 1 & 0.29 & 0.42 & 0.34   & 0.30 & 0.10 & 0.14 \\
 2 & 0.52 & 0.39 & 0.45   & 0.51 & 0.60 & 0.55 \\
 3 & 0.67 & 0.51 & 0.58   & 0.77 & 0.69 & 0.73 \\
 4 & 0.58 & 0.55 & 0.57   & 0.87 & 0.70 & 0.78  \\
 \midrule
 Mean & 0.57 & 0.55 & 0.55 & 0.59 & 0.62 & 0.59\\
 \bottomrule
\end{tabular}
\end{table}

To sum up, training a CNN from scratch for regression output gives low mean-squared error. The lack of ground truth affects the performance of the regression. To overcome this drawback, in the next approach multi-objective convolutional learning is investigated to quantify knee OA severity.

\subsubsection{Multi-objective Convolutional Learning}˜\\
In general, assessing knee OA severity is based on the multi-class classification of knee images and assigning KL grade to each distinct category \cite{shamir2009early,shamir2008wndchrm,orlov2008wnd,oka2008fully}. The author previously argued that assigning a continuous grade (0--4) to knee images through regression is a better approach for quantifying knee OA severity as the disease is progressive in nature. However, there is no ground truth i.e. KL grades on a continuous scale to train a network directly for regression output. Therefore, the networks are trained using multi-objective convolutional learning \cite{liu2015multi,tompson2014joint,ranftl2014deep,ning2005toward,rudd2016moon} to optimise a weighted-ratio of two loss functions: categorical cross-entropy and mean-squared error. Mean squared error gives the network information about the ordering of grades, and cross entropy gives information about the quantisation of grades. Intuitively, optimising a network with two loss functions provides a stronger error signal and it is a step to improve the overall quantification, considering both classification and regression results. 

\subsubsection*{Initial Configuration.}
\begin{figure}[t]
\centering
\includegraphics[width=0.8\textwidth]{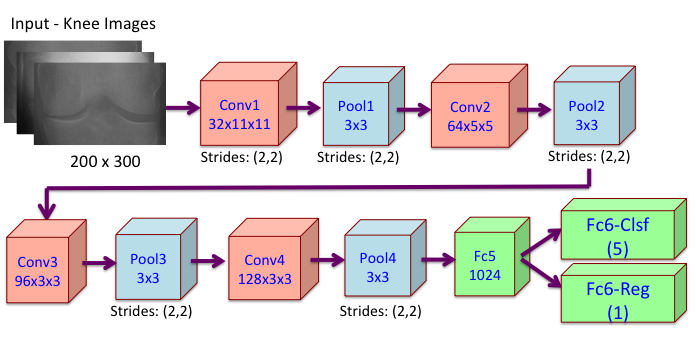}
\caption{Initial configuration to jointly train a CNN for classification and regression outputs.}
\label{fig:JL1}
\end{figure}

The same architecture of the best performing CNN is used for classification (Table \ref{Tab:bestClsf}) as an initial configuration to jointly train a CNN for classification and regression outputs. Table \ref{Tab:JL1} and Figure \ref{fig:JL1} shows the configuration details of the initial configuration. The network has five layers with learned weights: four convolutional layers and a fully connected layer. The total free parameters in the network are $\sim$5.4 million. The last fully connected layer (fc5) is followed by two dense layers with softmax and linear activations for simultaneous multi-class classification and regression outputs. Drop out layers with a drop out ratio 0.25 are included after the conv4 layer and a drop out ratio 0.5 after the fc6 layer to avoid overfitting. In addition to this, a L2-norm weight regularisation penalty of 0.01 is applied in conv3, conv4 and fc6 layers to avoid overfitting. Applying a regularisation penalty to other layers did not introduce significant variations in the learning curves. Unlike the previous approaches, this network is trained to minimise a weighted ratio of two loss functions: categorical cross-entropy and mean-squared error. After testing different values from 0.2 to 0.6 for the weight of regression loss, a ratio of 0.5 is fixed, as this ratio gives optimal results.

\begin{table}[t]
\caption{Initial configuration to jointly train a CNN for classification and regression outputs.}
\label{Tab:JL1}
\centering
\begin{tabular}{l c c c c }
\toprule 
Layer & Kernels & Kernel Size & Strides & Output shape\tabularnewline
\midrule
\midrule
conv1 & 32 & 11$\times$11 & 2 & 32$\times$100$\times$150 \tabularnewline
maxPool1 & -- & 3$\times$3 & 2 & 32$\times$49$\times$74 \tabularnewline
conv2 & 64 & 5$\times$5 & 1 & 64$\times$49$\times$74 \tabularnewline
maxPool2 & -- & 3$\times$3 & 2 & 64$\times$24$\times$36 \tabularnewline
conv3 & 96 & 3$\times$3 & 1 & 128$\times$24$\times$36  \tabularnewline
maxPool3 & -- & 3$\times$3 & 2 & 128$\times$11$\times$17 \tabularnewline
conv4 & 128 & 3$\times$3 & 1 & 256$\times$11$\times$17  \tabularnewline
maxPool4 & -- & 3$\times$3 & 2 & 256$\times$5$\times$8 \tabularnewline
fc5 & -- & -- & -- & 1024 \tabularnewline
fc6-Clsf & -- & -- & -- & 5 \tabularnewline
fc6-Reg & -- & -- & -- & 1 \tabularnewline
\bottomrule
\end{tabular}
\end{table}

The input to the network are knee images of size 200$\times$300. The knee images taken from the combined OAI-MOST dataset is used for training this network. The same train (70\%) and test (30\%) split are maintained from the previous experiments to make valid comparisons of the quantification results from the different methods. The right-left flip of the knee images is included to increase the training data. A validation split of 20\% from the training data is used. This network is trained using the Adam optimise with default parameters: initial learning rate $(\alpha) = 0.001$, $\beta_{1} = 0.9$, $\beta_{2} = 0.999$, $\epsilon = 1\mathrm{e}^{-8}$, as it gives faster convergence in comparison to the standard SGD. 

\subsubsection*{Initial Results.}
\begin{figure}[t]
\centering
	\subfloat[Classification]{\includegraphics[width=0.8\textwidth]{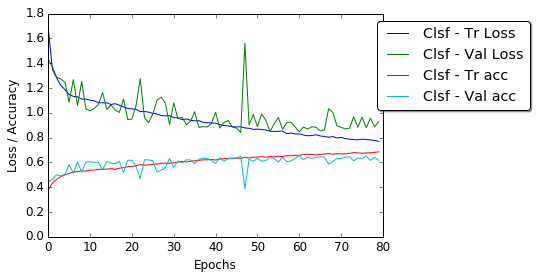}}
    \hfill
    \subfloat[Regression]{\includegraphics[width=0.8\textwidth]{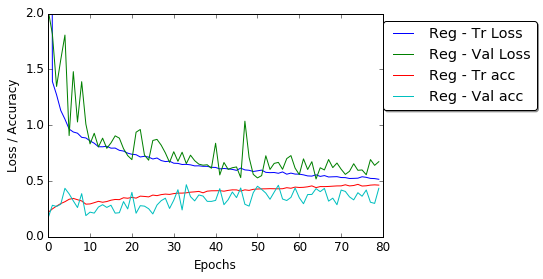}}
\caption{Learning curves for (a) classification and (b) regression in jointly trained CNN.}
\label{fig:Lc_JL1}
\end{figure}

Figure \ref{fig:Lc_JL1} shows the learning curves obtained whilst jointly training the CNN for classification and regression outputs. The learning curves show convergence in the validation and training losses with improvement in validation and classification accuracies. The jointly trained CNN with the initial configuration gives a classification accuracy of 60.8\% with mean-squared error 0.795 for the classification outputs and 0.652 for the regression outputs. These results do not show improvement from the previous results. Previously, the network with the same configuration gave a classification accuracy of 61.8\% and a mean-squared error 0.735 (Table \ref{Tab:bestClsf_res}) when trained to minimise only the classification loss. The same configuration after training to minimise only with the regression loss gave a mean squared error of 0.654 (Table \ref{Tab:reg1_res}). This configuration is optimal to minimise classification loss as it gave the highest classification accuracy (61.8\%). However, this configuration is not optimal for regression as it gives a high mean-squared error (0.654). Next, the number of layers with learned weights and other hyper-parameters in this configuration are varied to find a good architecture that will give improved results for both classification and regression outputs. 

\subsubsection*{Tuning Hyper-parameters.}
Cascaded convolutional stages are included in the next configuration in an attempt to improve the regression outputs. The CNNs with cascaded convolutional stages gave best results for regression (Table \ref{Tab:bestReg}) in the previous approach. Table \ref{Tab:JL2} shows the network details. This network contains six layers of learned weights: five convolutional layers and a fully connected layer. The total free parameters in this network are $\sim$7.8 million. The other settings remain the same from the previous network and the same training procedure is followed. 

\begin{table}[t]
\caption{Jointly trained network for classification and regression outputs.}
\label{Tab:JL2}
\centering
\begin{tabular}{l c c c c }
\toprule 
Layer & Kernels & Kernel Size & Strides & Output shape\tabularnewline
\midrule
\midrule
conv1 & 32 & 11$\times$11 & 2 & 32$\times$100$\times$150 \tabularnewline
maxPool1 & -- & 3$\times$3 & 3 & 32$\times$33$\times$50 \tabularnewline

conv2-1 & 64 & 3$\times$3 & 1 & 64$\times$33$\times$50 \tabularnewline
conv2-2 & 64 & 3$\times$3 & 1 & 64$\times$33$\times$50 \tabularnewline
maxPool2 & -- & 3$\times$3 & 2 & 64$\times$16$\times$24 \tabularnewline

conv3-1 & 96 & 3$\times$3 & 1 & 96$\times$16$\times$24  \tabularnewline
conv3-2 & 96 & 3$\times$3 & 1 & 96$\times$16$\times$24  \tabularnewline
maxPool3 & -- & 3$\times$3 & 2 & 96$\times$7$\times$11 \tabularnewline

fc4 & -- & -- & -- & 1024 \tabularnewline
fc5-Clsf & -- & -- & -- & 5 \tabularnewline
fc5-Reg & -- & -- & -- & 1 \tabularnewline
\bottomrule
\end{tabular}
\end{table}

This network gives a multi-class classification accuracy of 62.9\% and the mean-squared error is 0.754 for the classification output and 0.583 for the regression output.  These results show improvement in the quantification performance in comparison to the previous results. Tuning the hyper parameters improves both the classification and the regression outcomes. Next, the depth of the architecture is increased and other related hyper parameters are tuned to investigate further improvement in the classification and regression outputs. 

\subsubsection*{Best Performing Jointly Trained CNN.}
The best configuration (Table \ref{Tab:bestJL}) is obtained after experimenting with different settings for jointly training a CNN for classification and regression outputs. This network has eight layers with learned weights: seven convolutional layers and a fully connected layer. This network has $\sim$2.9 million free parameters in total. This is a lightweight architecture with minimal parameters in comparison to the previous networks and the existing off-the-shelf CNNs. Each convolutional layer is followed by batch normalisation and a ReLU activation layer. The fc5 layer is followed by two dense layers with softmax and linear activations for multi-class classification and regression outputs. To avoid overfitting,  drop out with ratio 0.3 is included after the last fully connected (fc5) layer. Also, a L2 weight regularisation penalty of 0.01 is applied to all the convolutional and fully connected layers except the first two convolutional layers. This network is trained to minimise a weighted ratio of two loss functions: categorical cross-entropy and mean-squared error. This network is trained using the Adam optimiser with default parameters: initial learning rate $(\alpha) = 0.001$, $\beta_{1} = 0.9$, $\beta_{2} = 0.999$, $\epsilon = 1\mathrm{e}^{-8}$. 

\begin{table}[t]
\caption{Jointly trained network for classification and regression outputs.}
\label{Tab:bestJL}
\centering
\begin{tabular}{l c c c c }
\toprule 
Layer & Kernels & Kernel Size & Strides & Output shape\tabularnewline
\midrule
\midrule
conv1 & 32 & 11$\times$11 & 2 & 32$\times$100$\times$150 \tabularnewline
maxPool1 & -- & 3$\times$3 & 2 & 32$\times$49$\times$74 \tabularnewline

conv2-1 & 64 & 3$\times$3 & 1 & 64$\times$49$\times$74 \tabularnewline
conv2-2 & 64 & 3$\times$3 & 1 & 64$\times$49$\times$74 \tabularnewline
maxPool2 & -- & 3$\times$3 & 2 & 64$\times$24$\times$36 \tabularnewline

conv3-1 & 96 & 3$\times$3 & 1 & 96$\times$24$\times$36  \tabularnewline
conv3-2 & 96 & 3$\times$3 & 1 & 96$\times$24$\times$36  \tabularnewline
maxPool3 & -- & 3$\times$3 & 2 & 96$\times$11$\times$17 \tabularnewline

conv4-1 & 128 & 3$\times$3 & 1 & 128$\times$11$\times$17 \tabularnewline
conv4-2 & 128 & 3$\times$3 & 1 & 128$\times$11$\times$17 \tabularnewline
maxPool4 & -- & 3$\times$3 & 2 & 128$\times$5$\times$8 \tabularnewline

fc5 & -- & -- & -- & 512 \tabularnewline
fc6-Clsf & -- & -- & -- & 5 \tabularnewline
fc6-Reg & -- & -- & -- & 1 \tabularnewline
\bottomrule
\end{tabular}
\end{table}

\subsubsection*{Jointly Trained CNN Results.}
\begin{figure}[t]
\centering
	\subfloat[Classification]{\includegraphics[width=0.6\textwidth]{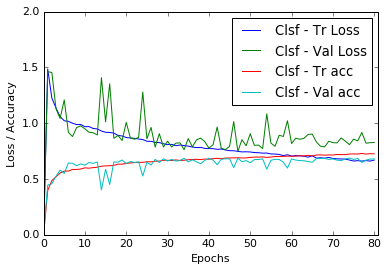}}
    \hfill
    \subfloat[Regression]{\includegraphics[width=0.6\textwidth]{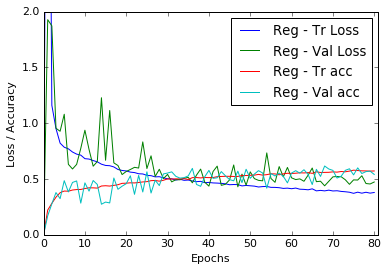}}
\caption{Learning curves for (a) classification and (b) regression in jointly trained CNN.}
\label{fig:Lc_bestJL}
\end{figure}

Figure \ref{fig:Lc_bestJL} shows the learning curves obtained whilst jointly training the CNN for classification and regression outputs. The learning curves show convergence to the minimum of the validation and training losses with improvement in validation and classification accuracies. The combined OAI-MOST dataset is used to compute these results. The same train-test split is maintained from the previous experiments. This jointly trained network gives a multi-class classification accuracy of 64.6\% with a mean-squared error 0.685 for the classification outputs and 0.507 for the regression outputs. Table \ref{Tab:bestJL_res} shows the precision, recall, and $F_{1}$ score of this network. There is an improvement in the results: the classification accuracy increases to 64.6\% from the initial configuration (60.8\%), the mean-squared error for regression decreases to 0.507 from the initial configuration (0.652). Increasing the depth of the architecture by including more layers with learned weights to the initial configuration and tuning the other hyper-parameters improves both the classification and regression results. Intuitively, providing a stronger error signal using both the classification and regression loss to the network allow to fit more parameters. 

\begin{table}[t]
\caption{Results of the best performing jointly trained CNN for classification and regression outputs.}
\label{Tab:bestJL_res}
\centering
\begin{tabular}{c c c c c c c}
\toprule
Grade & Precision & Recall & $F_{1}$ Score \\
\midrule
\midrule
 0 & 0.68 & 0.85 & 0.75 \\
 1 & 0.34 & 0.07 & 0.12 \\
 2 & 0.53 & 0.63 & 0.57 \\
 3 & 0.74 & 0.77 & 0.75 \\
 4 & 0.86 & 0.81 & 0.84 \\
 \midrule
 Mean & 0.62 & 0.65 & 0.60 \\
 \bottomrule
\end{tabular}
\end{table}

\subsubsection*{Results Comparison.}
The results of the jointly trained CNN are compared to the previous CNNs trained separately for classification and regression outputs. Table \ref{Tab:Comp_all} shows the multi-class classification accuracy and mean-squared error of the jointly trained CNN and the separately trained CNNs for classification and regression outputs. There is an improvement in the classification accuracy and also the mean-squared error decreases for the joint training. These results show that the network jointly trained for classification and regression learns a better representation in comparison to the previous network trained separately for classification and regression outputs. 

\begin{table}[t]
\caption{Comparison of results from jointly trained CNN and individually trained CNNs for classification and regression results.}
\label{Tab:Comp_all}
\centering
\begin{tabular}{l c c c}
\toprule
Method & Clsf-Accuracy & Clsf-MSE & Reg-MSE \\
\midrule
\midrule
CNN-Classification & 61.8\% & 0.735 & --- \\
CNN-Regression & 54.7\% & --- & 0.574  \\
\midrule
Jointly trained CNN & \textbf{64.6\%} & \textbf{0.685} & \textbf{0.507} \\
\bottomrule
\end{tabular}
\end{table}

In summary, CNNs are trained from scratch to quantify knee OA severity using three approaches: classification, regression and jointly training for simultaneous classification and regression. From the results it is evident that the joint training outperforms both the individual training for classification and regression outputs. This supports the hypothesis that training a CNN for optimising a weighted ratio of two loss functions can improve the overall quantification of knee OA severity. 

\subsubsection*{Error analysis.}
\begin{figure}[t]
\centering
\includegraphics[width=0.45\textwidth]{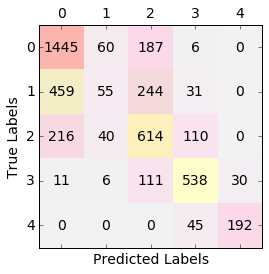}
\caption{Confusion matrix for the multi-class classification using the jointly trained CNN.}
\label{fig:ConfMat_JL}
\end{figure}

\begin{figure}[t]
\centering
\includegraphics[width=0.75\textwidth]{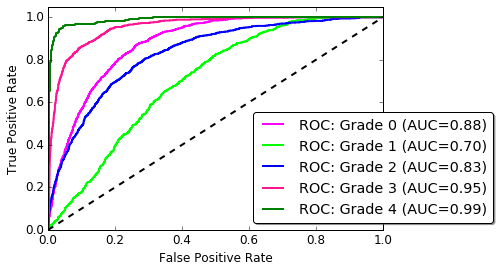}
\caption{ROC for the multi-class classification using the jointly trained CNN. }
\label{fig:Roc_JL}
\end{figure}

A confusion matrix and the area under curve (AUC) after plotting the receiver operating characteristics are computed to perform an error analysis on the classification of the knee images by the jointly trained CNN. From the classification metrics (Table \ref{Tab:bestJL_res}), the confusion matrix (Figure \ref{fig:ConfMat_JL}), and the receiver operating characteristic (ROC)  curves (Figure \ref{fig:Roc_JL}), it is evident that classification of successive grades is challenging, and in particular classification metrics for grade 1 have low values in comparison to the other grades.

\begin{figure}[t]
\centering
\includegraphics[width=0.9\textwidth]{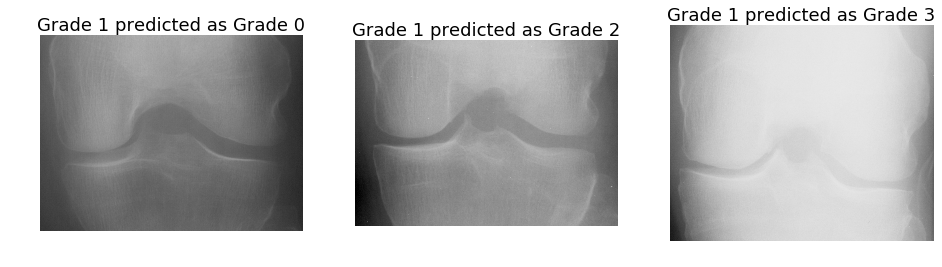}
\caption{Mis-classifications: grade 1 joints predicted as grade 0, 2, and 3.}
\label{fig:Err_gr1as023}
\end{figure}

\begin{figure}[t]
\centering
\includegraphics[width=0.9\textwidth]{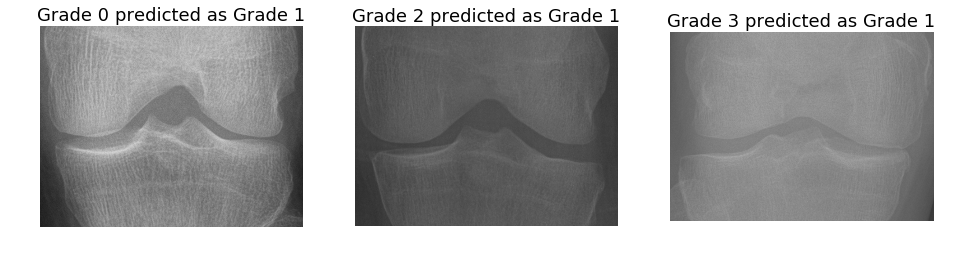}
\caption{Misclassification: other grade knee joints predicted as grade 1.}
\label{fig:Err_gr023as1}
\end{figure}

\begin{figure}[t]
\centering
\includegraphics[width=0.7\textwidth]{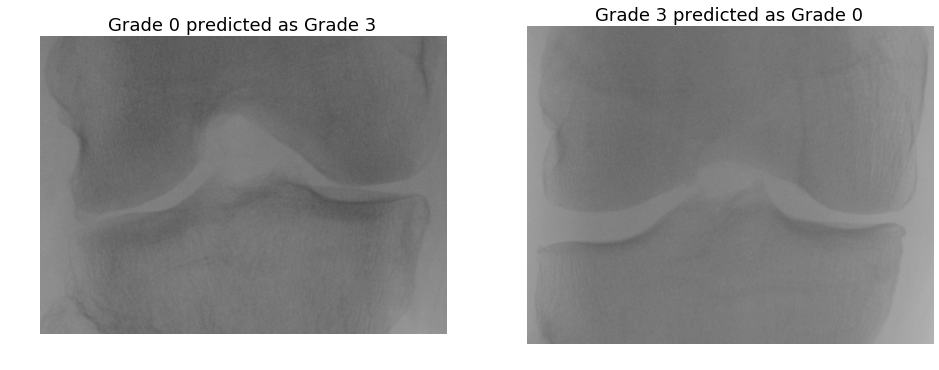}
\caption{An instance of more severe misclassification: grade 0 and grade 3.}
\label{fig:Err_gr03}
\end{figure}

Figure \ref{fig:Err_gr1as023} shows some examples of misclassification: grade 1 knee joints predicted as grade 0, 2, and 3. Figure \ref{fig:Err_gr023as1} shows the misclassification of knee joints categorised as grade 0, 2 and 3 predicted as grade 1. These images show minimal variations in terms of joint space width and osteophytes formation, making them challenging to distinguish. Even  the more serious misclassification in Figure \ref{fig:Err_gr03}, for instance grade 0 predicted as grade 3 and vice versa, do not show very distinguishable variations. Furthermore, when the knee X-ray images belonging to grade 0 and grade 1 severity are examined, it can be seen that there are very subtle variations in terms of the joint space width and osteophytes formation. Even better representations are needed to capture these fine-grained variations and to distinguish coarse grades: grade 0 and grade 1 images.

\subsubsection*{Discussion.}
Jointly training a CNN from scratch using the multi-objective convolutional approach improves the multi-class classification accuracy and minimises the mean-squared error. However, successive grade classification still remains a challenge. Even though the KL grades are widely used for assessing knee OA severity in clinical settings, there has been continued investigation and criticism over the use of KL grades as the individual categories are not equidistant from each other \cite{felson1997defining,schiphof2008differences,hart2003kellgren,emrani2008joint,shamir2010assessment}. This could be a reason for the low multi-class classification accuracy in the automatic quantification. Using OARSI readings instead of KL grades could possibly provide better results for automatic quantification as the knee OA features such as joint space narrowing, osteophytes formation, and sclerosis are separately graded. Moreover, when the knee X-ray images belonging to grade 0 and grade 1 severity are visually examined, it can be seen that there are very subtle variations in terms of the joint space width and osteophytes formation. To capture these variations and distinguish these coarse grades, for instance grade 0 versus grade 1, even better representations are required. Indeed, it should also be recognised that even medical experts do not always agree upon a particular KL grade e.g. either 0 or 1 attributed to the initial stage of knee OA \cite{felson1997defining,schiphof2008differences,hart2003kellgren}. 

\subsubsection{Ordinal Regression}˜\\
Ordinal regression\footnote{\small{\url{https://statistics.laerd.com/spss-tutorials/ordinal-regression-using \ -spss-statistics.php}}} is an intermediate task between multi-class classification and regression, sharing the properties of both. The outcomes or predictions in multi-class classification are discrete values and there is a meaningful order in the classes in regression. Ordinal regression is useful to classify patterns using a categorical scale which shows a natural order between the labels \cite{gutierrez2016ordinal,pedregosa2017consistency}. The misclassification from a normal classifier are treated the same, that is no misclassification are worse than others \cite{beckham2016simple}. However, some mis-classifications in ordinal regression, for instance the mis-classification on the extreme grades: grade 0 to grade 4 is treated worse than the others. This implies that the distances between the classes need to be taken into account when training a classifier. When quantifying the stages of a physical disease, it is preferable to predict the stage as `mild' or `doubtful' than `absent' when the true label is `severe'. Ordinal regression models formalise this notion of order by ensuring that
predictions farther from the true label incur a greater penalty than those closer to the true label \cite{pedregosa2017consistency}. The author believes that the KL grades prediction based on ordinal regression can further improve classification performance by reducing the margin of error (mean-squared error), considering the progressive nature of knee OA and the ground truth or labels for training a CNN i.e. the KL grades in an ordinal scale (0--4).

\subsubsection*{CNN Configuration for Ordinal Regression.} 
\begin{figure}[t]
\centering
\includegraphics[width=0.9\textwidth]{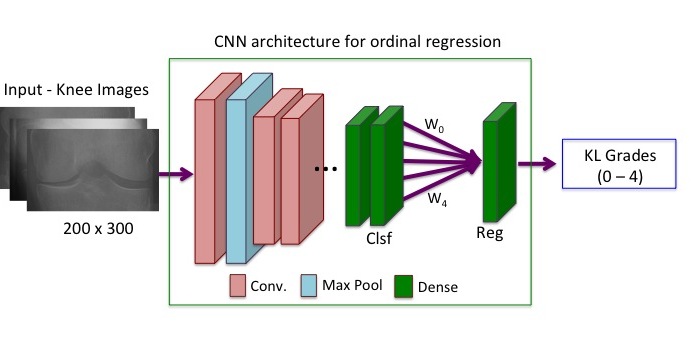}
\caption{The CNN configuration for ordinal regression. }
\label{fig:ordReg}
\end{figure}

For ordinal regression output, the last stage of the CNN (Table \ref{Tab:bestJL}) that gave best results on the joint training for multi-class classification and regression is modified. The previous approach on the joint training used two dense layers with softmax and linear activations in parallel (Figure \ref{fig:JL1}) for simultaneous multi-class classification and regression outputs. To train the CNN for ordinal regression, fixed weights ($[w_{0}, w_{1}, w_{2}, w_{3}, w_{4}] = [0, 1, 2, 3, 4]$) are applied to the outputs (probabilities) from the dense layer (Clsf) with softmax activations and back-propagate through a dense layer (Reg) with linear activations, optimising the mean-squared error loss function. The dense layer with softmax activations is treated as a hidden layer in this configuration. This is similar to the approach proposed by Beckham et al. \cite{beckham2016simple} for ordinal classification. Figure \ref{fig:ordReg} shows the CNN configuration for ordinal regression. 

\subsubsection*{CNN Training.}
The CNN for ordinal regression (Table \ref{Tab:ordReg}) is based on a lightweight architecture with $\sim$2.9 million free parameters in total and it contains eight layers with learned weights: seven convolutional layers and a fully connected layer. Each convolutional layer is followed by batch normalisation and a ReLU activation layer. To avoid overfitting, drop out with ratio 0.3 is applied after the last fully connected (fc5) layer. In addition to this, a L2 weight regularisation penalty of 0.01 is applied to all the convolutional and fully connected layers except the first two convolutional layers. The fc5 layer is followed by two dense layers with softmax (fc6-Clsf) and linear activations (fc7-Reg). The output of the softmax (fc6-Clsf) layer is multiplied (dot product) with fixed weights ([0,1,2,3,4]) and given as input to the last dense layer (fc7-Reg). This network is trained to minimise two loss functions: categorical cross-entropy and mean-squared error with equal weights. The dense layers (fc7-Reg) and (fc6-Clsf) provides the ordinal regression and multi-class classification outputs. The network is trained for 80 epochs with a batch size 32, using the Adam optimiser with default parameters: initial learning rate $(\alpha) = 0.001$, $\beta_{1} = 0.9$, $\beta_{2} = 0.999$, $\epsilon = 1\mathrm{e}^{-8}$. The same training, validation, and test data are used from the joint training to make valid comparison of the results.

The CNN configuration in Table \ref{Tab:ordReg} gives the best results for ordinal regression and this configuration is similar to the network for joint training (Table \ref{Tab:bestJL}) except the arrangement of the last two dense layers. Tuning the hyper-parameters of this CNN by increasing the number of layers with learned weights does not improve the quantification performance. Therefore, this CNN configuration is selected as the final network for ordinal regression.

\begin{table}[t]
\caption{CNN architecture for ordinal regression}
\label{Tab:ordReg}
\centering
\begin{tabular}{l c c c c }
\toprule 
Layer & Kernels & Kernel Size & Strides & Output shape\tabularnewline
\midrule
\midrule
input & -- & -- & -- & 1$\times$200$\times$300 \tabularnewline
conv1 & 32 & 11$\times$11 & 2 & 32$\times$100$\times$150 \tabularnewline
maxPool1 & -- & 3$\times$3 & 2 & 32$\times$49$\times$74 \tabularnewline

conv2-1 & 64 & 3$\times$3 & 1 & 64$\times$49$\times$74 \tabularnewline
conv2-2 & 64 & 3$\times$3 & 1 & 64$\times$49$\times$74 \tabularnewline
maxPool2 & -- & 3$\times$3 & 2 & 64$\times$24$\times$36 \tabularnewline

conv3-1 & 96 & 3$\times$3 & 1 & 96$\times$24$\times$36  \tabularnewline
conv3-2 & 96 & 3$\times$3 & 1 & 96$\times$24$\times$36  \tabularnewline
maxPool3 & -- & 3$\times$3 & 2 & 96$\times$11$\times$17 \tabularnewline

conv4-1 & 128 & 3$\times$3 & 1 & 128$\times$11$\times$17 \tabularnewline
conv4-2 & 128 & 3$\times$3 & 1 & 128$\times$11$\times$17 \tabularnewline
maxPool4 & -- & 3$\times$3 & 2 & 128$\times$5$\times$8 \tabularnewline

fc5 & -- & -- & -- & 512 \tabularnewline
fc6-Clsf & -- & -- & -- & 5 \tabularnewline

input-weights & -- & -- & -- & 5 \tabularnewline
merge-product & -- & -- & -- & 1 \tabularnewline
fc7-Reg & -- & -- & -- & 1 \tabularnewline

\bottomrule
\end{tabular}
\end{table}

\subsubsection*{Results.}

\begin{figure}[ht]
\centering
	\subfloat[Classification]{\includegraphics[width=0.6\textwidth]{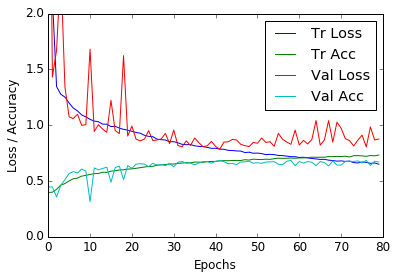}}
    \hfill
    \subfloat[Ordinal regression]{\includegraphics[width=0.6\textwidth]{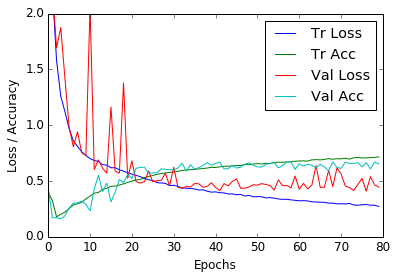}}
\caption{Learning curves for (a) classification and (b) ordinal regression .}
\label{fig:Lc_OR}
\end{figure}

\begin{figure}[ht]
\centering
\includegraphics[width=0.6\textwidth]{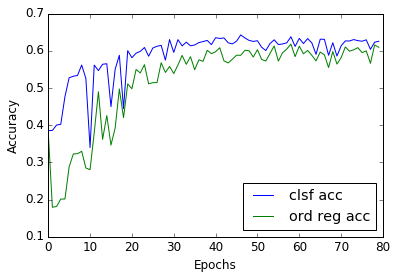}
\caption{The CNN configuration for ordinal regression.}
\label{fig:OR_test}
\end{figure}

The learning curves (Figure \ref{fig:Lc_OR}) obtained whilst training the CNN for ordinal regression shows convergence of the validation and training losses with improvement in validation and classification accuracies. Figure \ref{fig:OR_test} shows the classification accuracy of the trained CNN model after every epoch on the test data for the ordinal regression and classification output. After 40 epochs of training, there is no significant improvement in the classification accuracies. There is a slight decrease in the accuracy of the ordinal regression in comparison to the classification. This is likely due to the rounding of the output.

After training, the CNN gives a classification accuracy of 64.3\% on the test data. In the previous method on jointly training a CNN for classification and regression (Section 5.4.5), a multi-class classification accuracy of 64.6\% (Table \ref{Tab:bestJL_res}) is achieved. As the same CNN configuration  is used except the last stage (Figure \ref{fig:ordReg}) and other settings are retained, the classification performance remains almost the same for the CNN trained for ordinal regression in comparison to the jointly trained CNN. 

The classification metrics for the ordinal regression output are computed by rounding the predictions to integer values (0, 1, 2, 3, or 4). After rounding, the classification accuracy for the ordinal regression output is 61.8\% with mean-squared error 0.504 on the test data. The classification metrics for the regression output from the jointly trained CNN gives a classification accuracy 53.3\% with mean squared error 0.595 on the test data. There is an improvement in the classification accuracy and mean-squared error for ordinal regression in comparison to the previous regression results (see Table 40). Table \ref{Tab:OR_comp} shows the precision, recall, and $F_{1}$ score for regression and ordinal regression. From the results it is evident that ordinal regression is out performing regression for quantifying knee OA images in a continuous scale. 

\begin{table}[ht]
\caption{Comparison of classification metrics from regression and ordinal regression.}
\label{Tab:OR_comp}
\centering
\begin{tabular}{c c c c c c c}
\toprule
\multirow{2}{*}{Grades} &  \multicolumn{3}{c}{Regression} & \multicolumn{3}{c}{Ordinal Regression}\\ 
\cline{2-7}
 & Precision & Recall & $F_{1}$ & Precision & Recall & $F_{1}$\\
\midrule
\midrule
 0 & 0.78 & 0.58 & 0.66   & 0.71 & 0.79 & 0.75 \\
 1 & 0.29 & 0.55 & 0.38   & 0.31 & 0.33 & 0.32 \\
 2 & 0.51 & 0.50 & 0.50   & 0.59 & 0.44 & 0.50 \\
 3 & 0.65 & 0.53 & 0.58   & 0.74 & 0.73 & 0.73 \\
 4 & 0.63 & 0.33 & 0.43   & 0.81 & 0.76 & 0.78 \\
 \midrule
 Mean & 0.60 & 0.53 & 0.55 & 0.62 & 0.62 & 0.61\\
 \bottomrule
\end{tabular}
\end{table}

\subsubsection*{Results Comparison.}
Four approaches are investigated to automatically quantify knee OA severity. Table \ref{Tab:Comp_allRes} shows the multi-class classification accuracy and mean-squared error for the four approaches: 1) fine-tuning off-the-shelf CNN (BVLC CaffeNet), 2) training CNNs from scratch individually for classification and regression, 3) jointly training a CNN based on multi-objective convolutional learning for classification and regression, and 4) training a CNN for ordinal regression. These results are compared to WNDCHRM, that gave the previous best results for automatically classifying knee OA radiographs. The results show that jointly trained CNN gives the best results for multi-class classification. The ordinal regression outperforms all the other methods for quantifying knee OA images in a continuous scale with low mean-squared error. 

\begin{table}[t]
\caption{Comparison of classification, regression and ordinal regression results.}
\label{Tab:Comp_allRes}
\centering
\begin{tabular}{l c c}
\toprule
Method & Classification Accuracy & Mean-Squared Error \\
\midrule
\midrule
WNDCHRM & 34.8\% & 2.112 \\
Fine-Tuned BVLC CaffeNet & 57.6\% & 0.836 \\
CNN-Classification & 61.8\% & 0.735 \\
CNN-Regression & 54.7\% & 0.574  \\
Jointly trained CNN & \textbf{64.6\%} & 0.507 \\
Ordinal Regression & 64.3\% & \textbf{0.480} \\
\bottomrule
\end{tabular}
\end{table}

\subsection{An Automatic Knee OA Diagnostic System}
\begin{figure}[t]
  \centering
  \includegraphics[scale=0.45]{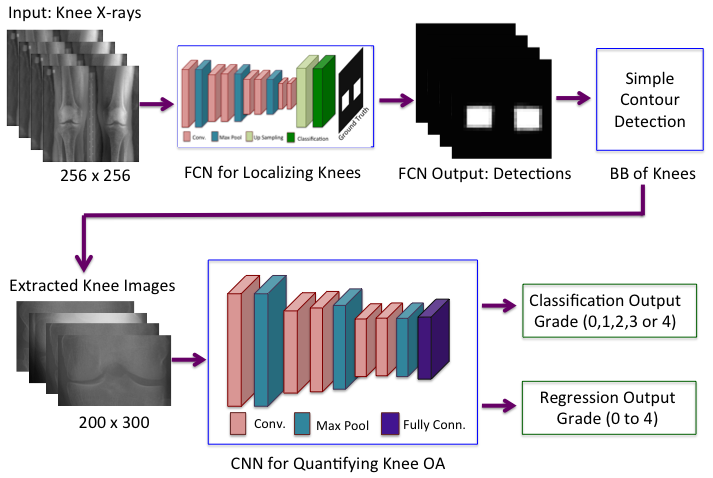}
  \caption{The proposed pipeline for quantifying knee OA severity.}
  \label{fig:Pipeline}
\end{figure}

An automatic knee OA diagnostic system is developed combining the automatic localisation pipeline and the quantification pipeline developed in the previous sections. Figure \ref{fig:Pipeline} shows the proposed end-to-end diagnostic pipeline to automatically quantify knee OA severity based on KL grades. The input X-ray images are subjected to histogram equalisation, mean normalisation and resized to a fixed size 256$\times$256. A fully convolutional network (FCN) is used to automatically detect the ROI, the knee joint regions. The bounding box coordinates of the ROI are calculated using simple contour detection. The knee joint regions are extracted from the knee radiographs using the bounding box coordinates. The localised and extracted knee images are resized to 200$\times$300 to preserve the mean aspect ratio ($\sim$1.6) and fed to the jointly trained CNN. This system provides quantification of knee OA severity in both discrete grades (classification) and continuous grades (regression) simultaneously.

The major pathological features that indicate the onset of knee OA include: reduction in joint space width due to loss of knee cartilage, and the formation of bone spurs (osteophytes) or bony projections along the joint margins. The author believes that quantifying these features along with the KL grades can provide deeper insights to assess knee OA severity and to study the progression of knee OA. Therefore, a deep learning-based automatic knee OA diagnostic system that can provide simultaneous predictions of KL grades, JSN, and osteophytes is developed. 

\subsection{Summary and Discussion}
Four approaches are presented in this section to automatically assess knee OA severity using CNNs. First, the existing pre-trained CNNs are investigated for classifying knee images based on KL grades. Two methods are used in this approach: using the pre-trained CNNs for fixed feature extraction, and fine-tuning the pre-trained CNNs using the transfer learning approach. The predictions or outputs from these methods are ordinal KL grades (0,1,2,3 or 4). Furthermore, the author argued that quantifying knee OA severity in a continuous scale (0--4) is more appropriate as the OA degradation is progressive in nature, not discrete. Regression is used to quantify the knee OA severity on  a continuous scale. The classification and regression results from the proposed methods in this chapter outperform the previous best results achieved by WNDCHRM, which uses many hand-crafted features with a variation of k-nearest neighbour classifier for classifying knee OA radiographs.

Second, CNNs are trained from scratch for classification and regression. The objective was to further improve the quantification results. As the training data is relatively scarce, a lightweight architectures with fewer ($\sim$4 to $\sim$5 million) free parameters are used in the CNNs. The fully trained CNN for classification achieved high classification accuracy in comparison to the pre-trained CNNs. However, the fully trained CNN for regression did not achieve high-performing results as no ground truth of KL grades was available on a continuous scale. Therefore, the discrete KL grades are used to train the CNNs for regression.

Third, CNNs are fully trained using multi-objective learning for simultaneous classification and regression. The intuition behind this is optimising a CNN with two loss functions provide a stronger error signal and it is a step to improve the overall quantification, considering both classification and regression results. The jointly trained CNN achieved better quantification results with a high classification accuracy in comparison to the previous methods.

As a last approach, CNNs are fully trained for ordinal regression using a softmax dense layer as the hidden layer. This approach achieved low mean-squared error and outperformed other methods to quantify knee OA severity in a continuous scale. The added benefit of this method is to provide simultaneous multi-class classification output. 

In summary, a progressive improvement is achieved in the quantification performance with an increase in classification accuracy and other performance metrics in the four approaches to automatically quantify knee OA severity. To conclude this Section, an error analysis was presented that discusses the possible reasons for the misclassification from the jointly trained CNN. The variations in the X-ray imaging protocols and discrepancies in the KL grades scoring need to be taken into account when analysing the mis-classification. 
%--------------------------------------------------------------------------%

\section{Conclusion}

The main goal of this research is to advance the state-of-the-art in computer aided diagnostics of the severity of knee OA by developing deep learning based automatic methods. According to the literature, automatic assessment of knee OA severity has been previously approached as an image classification problem and existing approaches report low accuracy for multi-class and classification of successive grades. The state-of-the-art machine learning based methods are investigated for image classification, and we developed new methods using convolutional neural networks (CNNs) to automatically classify knee OA images. A significant outcome of this research is a new automatic knee OA diagnostic system that achieves high accuracy, on par with radiologic reliability readings, which are considered the gold standard for knee OA assessment. 

In recent years, deep learning-based approaches, in particular CNNs, have become highly successful in many computer vision tasks and medical applications. Our research has mainly focused on developing a deep learning-based computer aided diagnostic system. The proposed approaches in this chapter are related to two main medical applications: localising or automatically detecting and extracting a region of interest (ROI) from a radiograph, and classifying the ROI to automatically assess disease severity. The FCN-based localisation approach could be extended to other medical applications such as localising a substructure or a ROI in MRI and CT scan images, object or lesion detection, locating anatomical landmarks or identifying imaging markers to study the disease progression. For instance, we followed a similar FCN-based approach to automatically detect and quantify ischemic injury (brain lesions) on diffusion-weighted MRI of infants, and we improved the state-of-the-art by achieving promising results. 

In the author's opinion the most interesting research findings in this research are as follows. First, fine-tuning off-the-shelf CNNs pre-trained on very large datasets such as ImageNet (with $\sim$1M images) to classify knee images with relatively small datasets (with $\sim$10,000 images) is promising for medical image classification. The main challenge in medical image classification is a lack of sufficient annotated data for training deep networks from scratch. Fine-tuning existing CNNs that have been trained using a large annotated dataset from a different application is possibly the best alternative to full training for medical applications. A second extremely interesting result is that training CNNs, optimising a weighted ratio of two loss functions for simultaneous classification and regression, provides a better error signal to the network and improves the overall classification performance. Many diseases are progressive by nature such as Alzheimer's disease, cancer, emphysema, tumours, lesions, and muscular dystrophy. Automatic quantification of such diseases using jointly trained CNNs may improve the quantification performance and provide insights to study the progression of the disease. Finally, it is very interesting that using multi-objective convolutional learning to jointly train CNNs based on different diagnostic features of a disease as the ground truth, can produce an overall improvement in the quantification performance achieving, results on par with human accuracy. Multi-objective learning and joint prediction of multiple regression and classification variables may be useful to assess diseases involving multiple diagnostic features like Alzheimer's, multiple sclerosis, and multiple myeloma (cancer). 

There are several potential directions for future work and further development of the research in this chapter. Some of the interesting extensions and prospects are outlined as follows.

\textbf{Training an end-to-end deep learning model:} The knee OA diagnostic pipeline consists of two steps: 1) localising the knee joints in radiographs and 2) assessing the knee OA severity from the localised knee joints. A FCN was trained for automatic localisation and a CNN was jointly trained for classification and regression of knee OA images. It would be interesting to train a single deep learning model integrating the FCN for localisation and the CNN for classification and/or regression, as this could further improve the automatic assessment of knee OA. In a recent work, Gorriz et al. \cite{gorriz2019assessing} presented a CNN attention-based end-to-end architecture to automatically assess knee OA severity. %Recently, end-to-end trained CNNs have become highly successful in saliency prediction \cite{pan2015end}, object detection \cite{wan2015end}, video classification \cite{fernando2016learning}, text recognition \cite{wang2012end}, and speech recognition \cite{zhang2017very}. 

\textbf{Using semantic segmentations to measure joint space width:} Among the knee OA diagnostic features, joint space narrowing is highly sensitive to changes due to disease progression. The proposed approach to automatically localise the knee joints using fully convolutional network can be extended for semantic segmentation of the knee joints and can be used to automatically measure the joint space width (JSW) between the femur and tibia. However, pixel level knee joint annotations in radiographs are needed to measure the JSW. 

\textbf{Assessing the progression of knee OA severity:} The automatic quantification methods developed in this study can be extended to assess the progression and early detection of knee OA severity. The baseline datasets are used from the OAI and the MOST dataset. Datasets are available for annual follow-up visits up to 9 years. These datasets could be used to detect the features predictive of radiographic knee OA progression. Shamir et al. reported a similar approach using WNDCHRM that predicted whether a knee would change from KL grade 0 to grade 3 with 72\% accuracy using 20 years of data \cite{shamir2009early}. The findings of our current work has indicated the ability to improve upon the existing methods. 

\textbf{Relating the automatic quantification results to knee pain:} The primary clinical features to assess knee OA are pain and radiographic evidence of deformity \cite{williams2004knee}. It would be interesting to study the relationship between the automatic assessments of the proposed methods (KL grades) to WOMAC scores for knee pain. WOMAC is one among the most widely used assessments in knee OA. 

\textbf{Relating the automatic quantification results to physiological variables:} There are several pathological and physiological variables available in the OAI and the MOST datasets. These variables include potential predictors of knee pain status. It would be interesting to study the relationship between the outcomes of the automatic methods and the predictions from the pathological and physiological variables. Abedin et al. \cite{abedin2019predicting} presented a comparative analysis to predict knee OA severity based on statistical models developed on physiological variables and a CNN model developed on features from X-ray images. 

\textbf{Investigating human level accuracy:} The radiologic reliability readings from the OAI used 150 participants (300 knees) to evaluate the test-retest reliability of semi-quantitative readings. This is considered the current gold standard for knee OA assessment. Simple kappa coefficients and weighted kappa coefficients were used to evaluate the inter-rater agreement. Investigating the human level accuracy for a large sample and comparing it with the automatic quantification results would provide insight to help reduce the error involved in automatic assessments.

%--------------------------------------------------------------------------%

\subsubsection*{Acknowledgments.}~\\
This publication has emanated from research conducted with the financial support of Science Foundation Ireland (SFI) under grant numbers SFI/12/RC/2289 and 15/SIRG/3283.

The OAI is a public-private partnership comprised of five contracts (N01-AR-2-2258; N01-AR-2-2259; N01-AR-2- 2260; N01-AR-2-2261; N01-AR-2-2262) funded by the National Institutes of Health, a branch of the Department of Health and Human Services, and conducted by the OAI Study Investigators. Private funding partners include Merck Research Laboratories; Novartis Pharmaceuticals Corporation, GlaxoSmithKline; and Pfizer, Inc. Private sector funding for the OAI is managed by the Foundation for the National Institutes of Health.

MOST is comprised of four cooperative grants (Felson
-- AG18820; Torner -- AG18832; Lewis -- AG18947; and Nevitt -- AG19069) funded by the National Institutes of Health, a branch of the Department of Health and Human Services, and conducted by MOST study investigators. This manuscript was prepared using MOST data and does not necessarily reflect the opinions or views of MOST investigators.

%--------------------------------------------------------------------------%

\bibliography{Biblio}{}
\bibliographystyle{splncs}

\end{document}